\documentclass[12pt]{report}   

\usepackage{setspace} 
 \usepackage{times}  
 \usepackage[explicit]{titlesec}  
 \usepackage[titles]{tocloft}  

\usepackage{appendix} 
 \usepackage{rotating}  
 \usepackage[normalem]{ulem}  
 \usepackage{textcomp} 
 \usepackage{indentfirst}
 \usepackage{booktabs,array,arydshln}
 \usepackage{amsmath} 
 \usepackage[T1]{fontenc} 
 \usepackage[utf8]{inputenc} 
 \usepackage{newtxmath} 

\usepackage[letterpaper, left=1in, right=1in, top=1in, bottom=1in]{geometry}

\usepackage{subfloat}

\usepackage{amsfonts}
\usepackage{bbm}

\usepackage{tgtermes}
\usepackage{scalefnt,letltxmacro}
\LetLtxMacro{\oldtextsc}{\textsc}
\renewcommand{\textsc}[1]{\oldtextsc{\scalefont{1.10}#1}}
\usepackage[scaled=0.92]{PTSans}

\usepackage[usenames,dvipsnames]{xcolor}
\usepackage{graphicx}
\usepackage{natbib}   
\usepackage[parfill]{parskip}
\usepackage{afterpage}
\usepackage{framed}
\usepackage{nicefrac}
\usepackage{bm}
\usepackage{fixmath}

\DeclareMathOperator*{\kl}{KL}

\newcommand{\xvec}{\mathbf{x}}
\newcommand{\zvec}{\mathbf{z}}

\usepackage[colorinlistoftodos,
           prependcaption,
           textsize=small,
           backgroundcolor=yellow,
           linecolor=lightgray,
           bordercolor=lightgray]{todonotes}

\usepackage{lineno}

\usepackage{ragged2e}

\DeclareRobustCommand{\parhead}[1]{\noindent\textbf{#1}~}

\usepackage{paralist}

\usepackage[labelfont=it,labelsep=period]{caption}
\usepackage{subfig}

\usepackage{multirow}
\usepackage{booktabs}
\usepackage{arydshln} 

\usepackage[ruled, lined, longend, linesnumbered]{algorithm2e}
\usepackage{algorithmic}
\usepackage{listings}
\usepackage{fancyvrb}
\fvset{fontsize=\normalsize}

\usepackage[colorlinks,linktoc=all]{hyperref}
\usepackage[all]{hypcap}
\hypersetup{citecolor=Violet}
 \hypersetup{linkcolor=black}
 \hypersetup{urlcolor=MidnightBlue}
\usepackage{url}

\usepackage[acronym,smallcaps,nowarn]{glossaries}

\newcommand{\red}[1]{\textcolor{BrickRed}{#1}}

\newcommand{\green}[1]{\textcolor{OliveGreen}{#1}}

\usepackage{listings}
\lstdefinestyle{alp_style}{
    commentstyle=\color{OliveGreen},
    numberstyle=\tiny\color{black!60},
    stringstyle=\color{BrickRed},
    basicstyle=\ttfamily\scriptsize,
    breakatwhitespace=false,
    breaklines=true,
    captionpos=b,
    keepspaces=true,
    numbers=none,
    numbersep=5pt,
    showspaces=false,
    showstringspaces=false,
    showtabs=false,
    tabsize=2
}
\usepackage{lipsum} \newtheorem{thm}{Theorem} \newtheorem{defn}[thm]{Definition}

\DeclareMathOperator*{\argmax}{arg\,max}
\DeclareMathOperator*{\argmin}{arg\,min}

\DeclareRobustCommand{\E}[2]{\mathbb{E}_{#1}\left[#2\right]}

\DeclareRobustCommand{\diag}[1]{\textrm{diag}\left(#1\right)}

\newcommand{\g}{\, | \,}
\newcommand{\prm}{\, ; \,}

\newcommand{\Lcal}{\mathcal{L}}
\newcommand{\Ncal}{\mathcal{N}}

\newcommand{\bzero}{\mathbf{0}}

\newcommand{\bz}{\mathbf{z}}
\newcommand{\bx}{\mathbf{x}}
\newcommand{\bw}{\mathbf{w}}
\newcommand{\bh}{\mathbf{h}}

\newcommand{\bc}{\mathbf{c}}
\newcommand{\bs}{\mathbf{s}}
\newcommand{\bp}{\mathbf{p}}

\newcommand{\bl}{\mathbf{l}}

\newcommand{\bv}{\mathbf{v}}
\newcommand{\bV}{\mathbf{V}}
\newcommand{\bI}{\mathbold{I}}

\newcommand{\beps}{\mathbold{\epsilon}}
\newcommand{\blambda}{\mathbold{\lambda}}
\newcommand{\bmu}{\mathbold{\mu}}
\newcommand{\btheta}{\mathbold{\theta}}
\newcommand{\bbeta}{\mathbold{\beta}}
\newcommand{\bdelta}{\mathbold{\delta}}
\newcommand{\bepsilon}{\mathbold{\epsilon}}

\newcommand{\bsigma}{\mathbold{\sigma}}
\newcommand{\balpha}{\mathbold{\alpha}}
\newcommand{\bSigma}{\mathbold{\Sigma}}
\newcommand{\cL}{\mathcal{L}}

\newcommand{\dd}{\, \mathrm{d}} \newacronym{ADVI}{advi}{automatic differentiation variational inference}
\newacronym{AR}{a{\small\&}r}{augment and reduce}
\newacronym{AI}{ai}{artificial intelligence}

\newacronym{BBVI}{bbvi}{black-box variational inference}

\newacronym{CBOW}{cbow}{continuous bag-of-words}
\newacronym{CDF}{cdf}{cumulative distribution function}
\newacronym{CS-EFE}{cs-efe}{context selection for exponential family embeddings}
\newacronym{CTM}{ctm}{correlated topic model}

\newacronym[\glslongpluralkey={deep exponential families}]{DEF}{def}{deep exponential family}
\newacronym{DMIS}{dmis}{deterministic multiple importance sampling}
\newacronym{DL}{dl}{deep learning}
\newacronym{DPGM}{dpgm}{\textit{deep probabilistic graphical modeling}}
\newacronym{DLDA}{d-lda}{dynamic latent {D}irichlet allocation}

\newacronym{EFE}{efe}{exponential family embeddings}
\newacronym{ELBO}{elbo}{evidence lower bound}
\newacronym{EM}{em}{expectation maximization}
\newacronym{ETM}{etm}{\textit{embedded topic model}}
\newacronym{DETM}{detm}{\textit{deep embedded topic model}}

\newacronym{GNTS}{gn-ts}{gamma-normal time series model}
\newacronym{G-REP}{g-rep}{generalized reparameterization}

\newacronym{HMC}{hmc}{{H}amiltonian {M}onte {C}arlo}

\newacronym{KL}{kl}{{K}ullback-{L}eibler}

\newacronym{LDA}{lda}{latent {D}irichlet allocation}

\newacronym{MAP}{map}{\textit{maximum a posteriori}}
\newacronym{MCMC}{mcmc}{{M}arkov chain {M}onte {C}arlo}
\newacronym{MF}{mf}{matrix factorization}
\newacronym{MIS}{mis}{multiple importance sampling}
\newacronym{ML}{ml}{machine learning}

\newacronym{NVDM}{nvdm}{neural variational document model}

\newacronym{OBBVI}{o-bbvi}{overdispersed black-box variational inference}
\newacronym{OVE}{ove}{one-vs-each}

\newacronym{PGM}{pgm}{probabilistic graphical modeling}

\newacronym{SIVI}{sivi}{semi-implicit variational inference}
\newacronym{SVI}{svi}{stochastic variational inference}

\newacronym{TMES}{tmes}{topic model in embedding space}

\newacronym{USIVI}{uivi}{unbiased implicit variational inference}

\newacronym{VAE}{vae}{variational autoencoder}
\newacronym{VEM}{vem}{variational expectation maximization}
\newacronym{VI}{vi}{variational inference}

\newacronym{ALI}{ali}{adversarially learned inference}
\newacronym{BIGAN}{bigan}{bidirectional generative adversarial network}
\newacronym{RNN}{rnn}{recurrent neural network}
\newacronym{MLP}{mlp}{feed forward neural network}
\newacronym{LSTM}{lstm}{long-short term memory}
\newacronym{GRU}{gru}{gated recurrent unit}
\newacronym{VRNN}{vrnn}{variational recurrent neural network}
\newacronym{SRNN}{srnn}{stochastic recurrent neural network}
\newacronym{ERNN}{ernn}{Elman recurrent neural network}
\newacronym{DCVAE}{dcvae}{deep-convolutional variational auto-encoder}
\newacronym{UIVI}{uivi}{unbiased implicit variational inference}
\newacronym{DGM}{dgm}{deep generative model}
\newacronym{IWAE}{iwae}{importance weighted auto-encoder}
\newacronym{WS}{ws}{wake-sleep}
\newacronym{RWS}{rws}{reweighted wake-sleep}
\newacronym{REM}{rem}{\emph{reweighted expectation maximization}}

\newacronym{GAN}{gan}{generative adversarial network}
\newacronym{DCGAN}{dcgan}{deep convolutional generative adversarial network}
\newacronym{PresGAN}{presgan}{prescribed generative adversarial network}
\newacronym{PGAN}{pgan}{prescribed generative adversarial network}
\newacronym{VEEGAN}{veegan}{veegan}
\newacronym{PACGAN}{pacgan}{packed {GAN}}
\newacronym{STYLEGAN}{stylegan}{Style {GAN}}
\newacronym{FID}{fid}{{F}r\'{e}chet {I}nception distance}

\newacronym{klVI}{klvi}{$\operatorname{KL}(q || p)$ variational inference}
\newacronym{AVA}{ava}{autoritative variational autoencoder}
\newacronym{SAVAE}{sa-vae}{semi-amortized variational autoencoder}
\newacronym{SkipVAE}{skip-vae}{Skip Variational Autoencoder}
\newacronym{SkipSAVAE}{skip-sa-vae}{Skip Variational Autoencoder}
\newacronym{LM}{language model}{Language Model}
\newacronym{GSM}{gsm}{generative skip model}
\newacronym{DGSM}{dgsm}{deep generative skip model}

\newacronym{NEA}{nea}{neural embedding allocation}

\newacronym{PRODLDA}{prodlda}{PRODLDA}
\newacronym{PML}{pml}{probabilistic machine learning}

\newacronym{WELDA}{welda}{WELDA}

\newacronym{UN}{un}{{U}nited {N}ations}
\newacronym{AVI}{avi}{amortized variational inference}
\newacronym{PCA}{pca}{principal component analysis}
\newacronym{EF-PCA}{ef-pca}{exponential family principal component analysis}
\newacronym{CNN}{cnn}{convolutional neural network}
\newacronym{RNNs}{rnns}{recurrent neural networks}
\newacronym{RNNLM}{rnnlm}{Recurrent Neural Network-based language model}
\newacronym{TopicRNN}{TopicRNN}{TopicRNN}
\newacronym{TopicLSTM}{TopicLSTM}{TopicLSTM}
\newacronym{TopicGRU}{TopicGRU}{TopicGRU}
\newacronym{IMDB}{IMDB}{IMDB}
\newacronym{HMM}{hmm}{hidden {M}arkov model}
\newacronym{AE}{ae}{autoencoder}
 
\iffalse
\AtEveryBibitem{\clearfield{issn}}
\AtEveryBibitem{\clearlist{issn}}

\AtEveryBibitem{\clearfield{language}}
\AtEveryBibitem{\clearlist{language}}

\AtEveryBibitem{\clearfield{doi}}
\AtEveryBibitem{\clearlist{doi}}

\AtEveryBibitem{\clearfield{url}}
\AtEveryBibitem{\clearlist{url}}

\AtEveryBibitem{  \ifentrytype{online}
    {}
    {\clearfield{urlyear}\clearfield{urlmonth}\clearfield{urlday}}}
\fi

\begin{document}
\doublespacing

\begin{titlepage}
\begin{center}

\begin{singlespacing}
\vspace*{6\baselineskip}
\textbf{\Large{Deep Probabilistic Graphical Modeling}}  \\
\vspace{3\baselineskip}
Adji Bousso Dieng\\
\vspace{18\baselineskip}
Submitted in partial fulfillment of the\\
requirements for the degree of\\
Doctor of Philosophy\\
under the Executive Committee\\
of the Graduate School of Arts and Sciences\\
\vspace{3\baselineskip}
COLUMBIA UNIVERSITY\\
\vspace{3\baselineskip}
2020
\vfill

\end{singlespacing}

\end{center}
\end{titlepage}

 \currentpdfbookmark{Title Page}{titlePage}  

\begin{titlepage}
\begin{singlespacing}
\begin{center}

\vspace*{35\baselineskip}

\textcopyright  \, 2020 \\
\vspace{\baselineskip}	
Adji Bousso Dieng\\
\vspace{\baselineskip}	
All Rights Reserved
\end{center}
\vfill

\end{singlespacing}
\end{titlepage}
 
\pagenumbering{gobble}

\begin{titlepage}
\begin{center}

\vspace*{5\baselineskip}
\textbf{\large Abstract}

Deep Probabilistic Graphical Modeling

Adji Bousso Dieng
\end{center}

\hspace{10mm} \Gls{PGM} provides a framework for formulating an interpretable generative process of data and expressing uncertainty about unknowns. This makes \gls{PGM} very useful for understanding the phenomena underlying data and for decision making. \gls{PGM} has been  successfully used in domains where interpretable inferences are key, e.g. marketing, medicine, neuroscience, and the social sciences. However \gls{PGM} tends to lack flexibility. This lack of flexibility makes \gls{PGM} often inadequate for modeling large-scale high-dimensional complex data and performing tasks that do require flexibility, e.g. vision and language applications. 

\Gls{DL} is an alternative framework for modeling and learning from data that has seen great empirical success in recent years. 
 \gls{DL} is very powerful and offers great flexibility, but it lacks the interpretability and calibration of \gls{PGM}. 

This thesis develops \gls{DPGM}. \gls{DPGM} consists in leveraging \gls{DL} to make \gls{PGM} more flexible. 
\gls{DPGM} brings about new methods for learning from data that exhibit the advantages of both \gls{PGM} and \gls{DL}. 

We use \gls{DL} within \gls{PGM} to build flexible models endowed with an interpretable latent structure. 
One family of models we develop extends \gls{EF-PCA} using neural networks to improve predictive performance 
while enforcing the interpretability of the latent factors. Another model class we introduce enables 
accounting for long-term dependencies when modeling sequential data, which is a 
challenge when using purely \gls{DL} or \gls{PGM} approaches. This model class 
for sequential data was successfully applied to language modeling, 
unsupervised document representation learning for sentiment analysis, 
conversation modeling, and patient representation learning for hospital readmission prediction. 
Finally, \gls{DPGM} successfully solves several outstanding problems of probabilistic topic models, a widely used family of probabilistic graphical models. 

Leveraging \gls{DL} within \gls{PGM} also brings about new algorithms for learning with complex data. 
We develop \gls{REM}, an algorithm that unifies several existing maximum likelihood-based algorithms 
for learning models parameterized by deep neural networks. This unifying view is made possible using expectation maximization, 
a canonical inference algorithm in \gls{PGM}. 
We also develop \textit{entropy-regularized adversarial learning}, a learning paradigm that deviates from the traditional maximum likelihood 
approach used in \gls{PGM}. From the \gls{DL} perspective, entropy-regularized adversarial learning 
provides a solution to the long-standing mode collapse problem of \glspl{GAN}, a widely used \gls{DL} approach. 

\vspace*{\fill}
\end{titlepage}

\pagenumbering{roman}
\setcounter{page}{1} 
\renewcommand{\cftchapdotsep}{\cftdotsep}  \renewcommand{\cftchapfont}{\normalfont}  \renewcommand{\cftchappagefont}{}  \renewcommand{\cftchappresnum}{Chapter }
\renewcommand{\cftchapaftersnum}{:}
\renewcommand{\cftchapnumwidth}{5em}
\renewcommand{\cftchapafterpnum}{\vskip\baselineskip} \renewcommand{\cftsecafterpnum}{\vskip\baselineskip}  \renewcommand{\cftsubsecafterpnum}{\vskip\baselineskip} \renewcommand{\cftsubsubsecafterpnum}{\vskip\baselineskip} 
\titleformat{\chapter}[display]
{\normalfont\bfseries\filcenter}{\chaptertitlename\ \thechapter}{0pt}{\large{#1}}

\renewcommand\contentsname{Table of Contents}

\begin{singlespace}
\tableofcontents
\end{singlespace}

\currentpdfbookmark{Table of Contents}{TOC}

\clearpage

\addcontentsline{toc}{chapter}{List of Tables}
\begin{singlespace}
	\setlength\cftbeforetabskip{\baselineskip}  	\listoftables
\end{singlespace}

\clearpage

\addcontentsline{toc}{chapter}{List of Figures}
\begin{singlespace}
\setlength\cftbeforefigskip{\baselineskip}  \listoffigures
\end{singlespace}

\clearpage

\addcontentsline{toc}{chapter}{Acknowledgments}

\clearpage
\begin{center}

\vspace*{5\baselineskip}
\textbf{\large Acknowledgements}
\end{center}

\hspace{10mm} I take this opportunity to thank people and organizations without whom this dissertation would not be possible. 

I thank my advisors David Blei and John Paisley for their unwavering support and for offering me the flexibility to pursue my own research interests. 
I was lucky to have them as advisors. I also thank Tian Zheng, Kyunghyun Cho, and John Cunningham for taking time to be part of my thesis committee.  

I thank Columbia and Google for awarding me fellowships to support my PhD. I thank Microsoft for granting me 
Azure cloud credits to facilitate my research. 

I am very fortunate to have had the opportunity to collaborate with many wonderful people: 
Francisco Ruiz, Michalis Titsias, Chong Wang, Jianfeng Gao, Dustin Tran, Rajesh Ranganath, 
Jaan Altosaar, Yoon Kim, and Sasha Rush. 

I am grateful to my mentors Yann LeCun and Kyunghyun Cho. I also thank all the people who hosted and mentored me during internships at Microsoft Research, Facebook AI Research, and DeepMind: Chong Wang, Jianfeng Gao, Yann LeCun, Kyunghyun Cho, Lei Yu, and Chris Dyer. I thank Alp Kucukelbir for helping me find my first PhD internship. 
I thank Philip Protter, Peter Orbanz, Kyle Cranmer, Hugo Larochelle, Danilo Rezende, Jasper Snoek, Scott Linderman, and Wesley Tansey 
for their support and/or advice. 

I am thankful to several people I met in the course of my PhD whom I am fortunate to have as friends: Makhtar Ba, Yassin Choye, 
Ibrahima Niang, Saliou Diallo, Nadia Raynes, Maimouna Diagne, Kashif Yusuf, Jing Wu, Emma Zhang, Peter Lee, 
Laurent Dinh, Zack Lipton, Cathy Seya, Yoon Kim, Siva Reddy, Maja Rudolph, 
Diana Cai, and Rajarshi Das. 

I would like to thank people I met before starting my PhD and without whom I wouldn't be able to pursue a PhD in the U.S. 
I thank Dr. Cheick Modibo Diarra for awarding me a scholarship to study abroad through his Pathfinder Foundation. 
I thank Mary Levy-Bruhl, Remi Barbet-Massin, and Jean Pierre Foulon from Lycee Henri IV. I thank Eric Moulines and Francois Roueff 
whom I had the fortune to learn Statistics and Probability from while at Telecom ParisTech. I thank Martin Wells and David Lifka who 
supported me and offered me the opportunity to work on solving concrete problems using data and computing during the time I spent at Cornell. 

Finally, I thank Patrick Guelah, Ousmane Kane, Mada Niang, Tening Diouf, Sarah Eugene, Arthur Bauer, Laetitia Gerin, and Cherif Gassama for their long-lasting friendship. 
I thank Jarra Jagne and Tonton Goumbala for their support. I owe a great deal to my family for providing me with plenty of love and moral support. 

\clearpage

\addcontentsline{toc}{chapter}{Dedication}

\begin{center}

\vspace*{5\baselineskip}
\textbf{\large Dedication}
\end{center}

\begin{flushright}
\hspace{10mm} 
To my late father. \\
To my mother, who gave me the gift of education. 
\end{flushright}

\clearpage
\pagenumbering{arabic}
\setcounter{page}{1} 

\addcontentsline{toc}{chapter}{Introduction}

\glsresetall

\begin{center}

\vspace*{5\baselineskip}
\textbf{\large Introduction}
\end{center}

\hspace{10mm} \Gls{PML} turns data into knowledge about the world.
This involves collecting data, specifying a model, fitting this model to the data, and performing evaluation on some 
criterion of interest, e.g. predictive performance.

\Gls{PGM} is an approach to \gls{PML} that specifies a model by specifying an interpretable generative process of data. 
This generative process often involves sampling a set of \textit{latent variables} from some prior distribution and 
then conditioning on these latent variables to generate data. The latent variables carry meaning; they represent 
the hidden structure underlying the data. Learning with \gls{PGM} involves estimating any parameters involved in specifying the model 
and discovering the hidden structure by performing posterior inference, i.e. learning the conditional distribution of the latent variables 
given the data. Often the posterior is intractable and we resort to variational inference, which uses optimization to find a tractable 
proxy for the true posterior. \gls{PGM} has been widely applied, for example to discover themes underlying 
a corpus of documents~\citep{blei2003latent}, to model speech~\citep{rabiner1989tutorial}, to understand user 
preferences for recommendation~\citep{wang2011collaborative}, to learn interaction patterns 
between different countries~\citep{schein2015bayesian}, etc. 

Despite its wide application, \gls{PGM} may lack flexibility. This has prevented its use in applications that do require flexibility, 
for example in vision and language applications. This thesis develops \gls{DPGM}, which consists in 
leveraging \gls{DL} to bring flexibility to \gls{PGM}. Leveraging \gls{DL} often means using neural networks to parameterize conditional 
distributions within a latent-variable model. \gls{DPGM} is agnostic to the choice of the architecture of these underlying neural networks.  
Therefore, the methodologies we develop in this thesis are amenable to more recent neural network 
architectures developed by the \gls{DL} community and any future innovations in the development of neural network architectures.  
The promise of \gls{DPGM} is to birth methodologies that enjoy the interpretability and calibration of \gls{PGM}, and the flexibility of \gls{DL}. 
Interpretability, by means of composing latent variables and parameters using inductive biases from domain knowledge, offers the ability 
to control the behavior of \gls{AI} systems. This controllability is key to a safe application of \gls{AI} to critical domains such as 
healthcare, autonomous and automated vision and language systems, and science.

The rest of the thesis is organized as follows. 

In Chapter\nobreakspace \ref {chap:foundations} we review the foundations for \gls{DPGM}. 
We first review \gls{PGM}, with a focus on the latent variable approach to \gls{PGM}. We describe exponential families as 
a unifying framework for representing distributions over random variables, observed or latent. We then describe several examples of \gls{PGM}s:  
\gls{EF-PCA}, \gls{LDA}, and \gls{DLDA}. 
We then discuss variational inference, a framework for approximating posterior distributions over latent variables. 
The second part of this chapter is a review of \gls{DL}. 
We first describe several neural network architectures and then review two \gls{DL} methodologies that are key to \gls{DPGM}, 
auto-encoding for dimensionality reduction and word embeddings. 
The final section of this chapter is a discussion of a line of work that combines neural networks with latent variables. 
In particular, we will describe \glspl{VAE} and discuss \textit{latent variable collapse}, a phenomenon that arises when 
parameterizing conditional distributions of a latent variable model with deep neural networks. 

In Chapter\nobreakspace \ref {chap:dpgm} we first describe three desiderata for \gls{DPGM} and then introduce several instances 
of \gls{DPGM}. One model class we introduce, called \textit{deep generative skip models}, extends \gls{EF-PCA}
using neural networks. Deep generative skip models achieve superior predictive performance 
and learn interpretable latent factors~\citep{dieng2019avoiding}. 
A second model class we introduce, called TopicRNN, marries latent variables and neural networks 
to model sequential data, addressing the long-term dependency issue encountered by 
purely \gls{DL} and \gls{PGM} approaches such as \glspl{RNN} and \glspl{HMM}. 
The model class defined by TopicRNN encodes inductive biases that have been shown useful for language 
modeling~\citep{dieng2016topicrnn}, conversation modeling~\citep{wen2018latent}, 
unsupervised document representation learning~\citep{dieng2016topicrnn}, 
and patient representation learning for hospital readmission prediction~\citep{xiao2018readmission}. 
Finally, we describe how to leverage word embeddings, a successful \gls{DL} approach that consists in representing words 
as continuous low-dimensional vectors, to solve several problems 
that pertain to probabilistic topic models, one of the most important \gls{PGM} class of models in 
terms of domain application~\citep{dieng2019topic, dieng2019dynamic}.  

The models introduced in Chapter\nobreakspace \ref {chap:dpgm} have intractable likelihoods. They are fit by maximizing a lower bound of 
the log marginal likelihood of the data, called the \gls{ELBO}, using \gls{VI}. 
This is the approach of \gls{VAE}s~\citep{Gershman2014, kingma2014autoencoding, rezende2014stochastic}. 
Since the lower bound is intractable, \gls{VAE}s use a Monte Carlo approximation of it for learning. 
\gls{VAE}s are prone to two main problems. First, the \gls{ELBO} they optimize may be a lose lower 
bound to the log-marginal likelihood of the data, which may hurt generalization performance. Second,  
\gls{VAE}s often suffer from \emph{latent variable collapse}, a phenomenon in which the learned 
latent variables do not represent good summaries of the data. 
In Chapter\nobreakspace \ref {chap:rem} we propose an alternative approach for learning \gls{DPGM}s called \gls{REM}. 
\gls{REM} optimizes a better approximation of the log marginal likelihood 
of the data~\citep{dieng2019reweighted}. It uses self-normalized importance sampling with moment matching to maximize the log marginal likelihood. 
\gls{REM} generalizes several existing algorithms that are based on maximum likelihood, such as 
the \gls{IWAE}~\citep{Burda2015} and \gls{RWS}~\citep{bornschein2014reweighted}. 
\gls{REM} leads to better generalization performance and yields more interpretable latent variables. 

Leveraging \gls{DL} for \gls{PGM} offers the opportunity to take advantage of algorithmic innovations in \gls{DL} 
to learn \glspl{PGM}. In Chapter\nobreakspace \ref {chap:presgan} we build on \glspl{GAN} and develop \textit{entropy-regularized adversarial learning}. 
Entropy-regularized adversarial learning provides an alternative to maximum likelihood for fitting \glspl{DPGM}. 
From the perspective of \gls{DL}, entropy-regularized adversarial learning constitutes a solution to the \textit{mode collapse} 
problem of \glspl{GAN}~\citep{dieng2019prescribed}. Addressing this mode collapse problem is important because 
under mode collapse, \gls{GAN} outputs lack diversity. This lack of diversity in outputs negatively affects the use  
of \glspl{GAN} for data augmentation, but also its application in healthcare and branches of \gls{ML} such as \textit{Fairness}. 

We conclude with a discussion of the contributions of this thesis and possibilities for future work. 
 
\titleformat{\chapter}[display]
{\normalfont\bfseries\filcenter}{}{0pt}{\large\chaptertitlename\ \large\thechapter : \large\bfseries\filcenter{#1}}  
\titlespacing*{\chapter}
  {0pt}{0pt}{30pt}	  
\titleformat{\section}{\normalfont\bfseries}{\thesection}{1em}{#1}

\titleformat{\subsection}{\normalfont}{\thesubsection}{0em}{\hspace{1em}#1}

\glsresetall

\chapter{Foundations}
\label{chap:foundations}

In this chapter we lay the foundations for \gls{DPGM} by reviewing \gls{PGM} and \gls{DL}. 
We end the chapter with a discussion of probabilistic conditioning with neural networks 
and the latent variable collapse issue that might arise from it.

\section{Probabilistic Graphical Modeling} 
\Gls{PGM} provides a useful framework for extracting knowledge from data. For example, a \gls{PGM} 
fit on a corpus of documents can tell us about the thematic structure underlying the documents. 
The \gls{PGM} approach to learning from data is to mimic the true process that generated the data. 
When specifying a generative process for data, to approximate the true data generating process, 
\gls{PGM} offers the ability to incorporate our prior knowledge about the phenomenon under study. 
For example, when studying a corpus of documents, \gls{PGM} allows us to integrate the knowledge 
that there is a set of topics discussed by all the documents in the corpus and that a given 
document expresses these topics at different lengths. 

\subsection{Latent Variables \& Interpretability} 
Consider observed a set of $N$ i.i.d data points. Denote them by $\bx_1, \dots, \bx_N$. 
The true phenomenon that generated these data is unknown and we want to learn about it. 
This will allow us to understand and make discoveries about the phenomenon underlying the data, 
perform prediction, and simulate new data. 
The \gls{PGM} approach is to posit the existence of a set of \textit{latent variables}, 
unobserved random variables that represent the hidden structure underlying the observed data. 
These latent variables are composed with the observations to form an interpretable generative process 
of data that approximates the true underlying data generating process. Often there are two 
sets of latent variables: \textit{global} latent variables and \textit{local} latent variables. 
Global latent variables capture the stable aspects of the underlying data generating process; 
they are shared across all the observations. Local latent variables express the singularities 
of each observation. For example consider a dataset of images of human faces. 
Global latent variables may represent the features of a face, e.g. eyes, lips, nose, cheeks, hair. 
Local latent variables will capture instantiations of these features; for example one image might 
depict brown eyes and dark hair whereas another image may depict green eyes and red hair. 

Denote by $\bbeta$ the global latent variables and by $\bz_1, \dots, \bz_N$ the local latent variables 
in a \gls{PGM}. The generative process specified by the \gls{PGM} implies 
a joint distribution over data and latent variables,
\begin{align}\label{eq:general_pgm}
	p(\bx_{1:N}, \bz_{1:N}, \bbeta) &= p(\bbeta) \cdot \prod_{i=1}^{N} p(\bz_i \vert \bbeta) \cdot p(\bx_i \vert \bz_i, \bbeta).
\end{align}
The distributions $p(\bbeta)$ and $p(\bz_i \vert \bbeta)$ are the priors over the global and $i^{th}$ local latent variable respectively. 
Their distributional forms can be chosen depending on the problem under study. The distribution $p(\bx_i \vert \bz_i, \bbeta)$ 
describes how to generate the $i^{th}$ observation $\bx_i$ by conditioning on $\bbeta$ and $\bz_i$. Our knowledge about 
the phenomenon under study is also expressed in terms of conditional independencies between the different variables, observed and latent. 
For example, the conditional distribution of $\bx_i$ may only depend on $\bz_i$. 

\subsection{Exponential Families} The exponential family provides a unifying framework for 
specifying probability distributions over random variables.  
The distributions mentioned above can be chosen to be in the exponential family.  
Almost all of the distributions used in practice are members of the exponential family, for example 
Gaussian, Gamma, Poisson, Bernoulli, Categorical, and Dirichlet.
Below is the formal definition of an exponential family.
\begin{defn}
A family of probability density functions $\mathcal{P} = \{p_{\theta}: \theta \in \Theta\}$ 
on a measure space $(\mathcal{X}, \mathcal{B}, \nu)$ is said to form an exponential family if 
\begin{align*}
	p_{\theta}(\bx) &= \exp \Big(\eta(\theta)^Tt(\bx) - A(\eta(\theta)) \Big) \\
	A(\eta(\theta)) &= \log \int_{}^{} \exp \Big(\eta(\theta)^Tt(\bx) \Big) \nu(d\bx)
\end{align*}
where $A(\eta(\theta))$ is called the log partition function (or log normalizer), $\eta(\theta)$ 
is called the natural parameter, and $t(\bx)$ denotes the vector of sufficient statistics.
\end{defn}

\begin{table*}[t]
\centering
\caption{Expressions of the sufficient statistic, the natural parameter, and the log normalizer for different members of the exponential family.}
\begin{tabular}{l|c|c|c|c}
\toprule
Distribution & Parameter $\theta$ & $t(\bx)$ & $\eta(\theta)$ & $A(\eta(\theta))$  \\
\midrule
Bernoulli & $p$  & $\bx$ & $\log \frac{p}{1 - p}$ &  $\log (1 + \exp(\eta(\theta)))$  \\
Gaussian & $(\bmu, \bsigma^2)$  & $\bx, \bx^2$ & $(\frac{\bmu}{\bsigma^2}, -\frac{1}{2\bsigma^2})$ & $-\frac{\eta(\theta)_1^2}{4\eta(\theta)_2} - \frac{1}{2}\log(-2\eta(\theta)_2)$ \\
Poisson & $\lambda$ & $\bx$ & $\log(\lambda)$ & $\exp(\eta(\theta))$ \\
Categorical & $p_{1:K}$ & $(\mathbb{I}(\bx = 1), \dots, \mathbb{I}(\bx = K))$ & $\log (p_{1:K})$ & $0$  \\
Dirichlet & $\alpha_{1:K}$ & $\log(\bx_{1:K})$  & $\alpha_{1:K}$ & $\sum_{k=1}^{K} \log \frac{\Gamma(\eta_k(\theta))}{\Gamma(\sum_{k=1}^{K} \eta_k(\theta))^{\frac{1}{K}}}$ \\
Gamma & $(\alpha, \beta)$ & $(\log(\bx), \bx)$ & $(\alpha -1, -\beta)$ & $\log\frac{\Gamma(\eta_1(\theta) + 1))}{-\eta_2(\theta)^{\eta_1(\theta) + 1}}$ \\
\bottomrule
\end{tabular}\label{tab:expfam}
\end{table*}

Table\nobreakspace \ref {tab:expfam} provides the expressions of the sufficient statistics, the natural parameter, 
and the log normalizer for several members of the exponential family. 

\subsection{Example: Exponential Family PCA} \label{subsec:efpca}
One canonical example of a \gls{PGM} is \gls{EF-PCA}~\citep{tipping1999probabilistic, collins2002generalization}. 
Assume observed $N$ i.i.d data points $\bx_1, \dots, \bx_N$ where $\bx_i \in \mathbb{R}^D$. \gls{EF-PCA} 
posits the following data generative process:
\begin{enumerate}
	\item Draw global latents $\bbeta \sim p(\bbeta)$
	\item For each data point $i = 1 \dots N$:
	 \begin{enumerate}
	 	\item Draw local latent variable $\bz_i \sim p(\bz)$
		\item Draw data point $\bx_i \sim \text{EF}(\eta_i = f(\bbeta^\top\bz_i))$ 
	 \end{enumerate}
\end{enumerate}
Here $\text{EF}(\eta)$ stands for an exponential family distribution with natural parameter $\eta$ 
and $f(\cdot)$ is a deterministic function that maps the dot product $\bbeta^\top\bz_i$ to the right space for the natural parameter. 
Note the local latent variables and the global latent variables interact linearly. 
This is a simplifying assumption that we will relax in Chapter\nobreakspace \ref {chap:dpgm}. 
When fit to data, \gls{EF-PCA} learns interpretable low-dimensional representations of data $\bz_{1:N}$.

\subsection{Example: Latent Dirichlet Allocation}
Another canonical \gls{PGM} is \gls{LDA}~\citep{blei2003latent}. 
\gls{LDA} is a probabilistic generative model of documents. It posits $K$
topics $\beta_{1:K}$, each of which is a distribution over a 
vocabulary (a predefined set of words).  \gls{LDA} assumes each document comes from a mixture of
topics, where the topics are shared across the corpus (they are global latent variables) and the mixture
proportions are unique to each document (they are local latent variables).  The generative process for
each document is the following:
\begin{compactenum}
\item Draw topic proportion $\theta_d \sim \textrm{Dirichlet}(\alpha_{\theta})$.
\item For each word $n$ in the document:
  \begin{compactenum}
    \setlength{\itemindent}{-0.3cm}
  \item Draw topic assignment $z_{dn} \sim \text{Cat}(\theta_d)$.
  \item Draw word $w_{dn} \sim \text{Cat}(\beta_{z_{dn}})$.
  \end{compactenum}
\end{compactenum}
Here, Cat$(\cdot)$ denotes the categorical distribution. \gls{LDA}
places a Dirichlet prior on the topics,
\begin{align*}
	\beta_k\sim \textrm{Dirichlet}(\alpha_{\beta}) \text{ for } k=1,\ldots,K.
\end{align*}
The concentration parameters $\alpha_{\beta}$ and $\alpha_{\theta}$ of the
Dirichlet distributions are fixed model hyperparameters often chosen to achieve 
a certain level of sparsity. Note all the distributions in \gls{LDA} are members of the exponential family. 

\gls{LDA} is a powerful model for document corpora. It has been extended in many ways 
and applied to many fields, such as marketing, sociology, political science, and the 
digital humanities. \citet{boydgraber2017applications} provide a review.

\subsection{Example: Dynamic Latent Dirichlet Allocation} \Gls{DLDA} is an extension of \gls{LDA} 
that allows topics to vary over time in order to analyze time-series corpora \citep{blei2006dynamic}. 
The generative model of \gls{DLDA} differs from \gls{LDA} in that the topics are
time-specific, i.e., they are $\beta_{1:K}^{(t)}$, where
$t\in\{1,\ldots,T\}$ indexes time steps. Moreover, the prior over the
topic proportions $\theta_d$ depends on the time stamp of document
$d$, denoted $t_d\in\{1,\ldots,T\}$. The generative process for each
document is:
\begin{compactenum}
\item Draw topic proportions $\theta_d \sim \mathcal{LN}(\eta_{t_d}, a^2 I)$.
\item For each word $n$ in the document:
  \begin{compactenum}
    \setlength{\itemindent}{-0.3cm}
  \item Draw topic assignment $z_{dn} \sim \text{Cat}(\theta_d)$.
  \item Draw word $w_{dn} \sim \text{Cat}(\beta_{z_{dn}}^{(t_d)})$.
  \end{compactenum}
\end{compactenum}
Here, $a$ is a model hyperparameter and $\eta_{t}$ is a latent variable that controls
the prior mean over the topic proportions at time $t$. To encourage smoothness over
the topics and topic proportions, \gls{DLDA} places random walk priors over
$\beta_{1:K}^{(t)}$ and $\eta_t$,
\begin{align*}
 	 \widetilde{\beta}_k^{(t)}\g\widetilde{\beta}_k^{(t-1)} &\sim\mathcal{N}(\widetilde{\beta}_k^{(t-1)}, \sigma^2 I) \text{ and } 
	 \beta_k^{(t)}=\textrm{softmax}(\widetilde{\beta}_k^{(t)})\\
	\eta_t\g \eta_{t-1} &\sim\mathcal{N}(\eta_{t-1}, \delta^2 I).
\end{align*}
The variables $\widetilde{\beta}_k^{(t)}\in\mathbb{R}^V$ are the transformed topics;
the topics $\beta_k^{(t)}$ are obtained after mapping $\widetilde{\beta}_k^{(t)}$
to the simplex, via the $\text{softmax}(\cdot)$ function. The hyperparameters $\sigma$ and $\delta$ control the smoothness of
the Markov chains.

\subsection{Posterior Inference} 
To discover the structure specified by all the models described above, we 
need to revert the generative process of data and compute the conditional distribution of the latent 
variables given the data. This conditional distribution is called the \textit{posterior distribution} of the latent variables. 
Consider the canonical \gls{EF-PCA} described earlier. The posterior distribution is 
\begin{align}
	p(\bbeta, \bz_{1:N} \vert \bx_{1:N}) 
	&= \frac{p(\bbeta) \cdot \prod_{i=1}^{N} p(\bz_i \vert \bbeta) \cdot p(\bx_i \vert \bz_i, \bbeta)}{\int_{}^{} p(\bbeta) \cdot \left[ \prod_{i=1}^{N} p(\bz_i \vert \bbeta) \cdot p(\bx_i \vert \bz_i, \bbeta) d\bz_i \right] d \bbeta }
\end{align} 

For simple models, the posterior has an analytical form. For many models, this is not the case and we must find a way to approximate the posterior.

\subsection{Variational Inference}
\Gls{VI} approximates the posterior using optimization. The idea is to posit a family of
approximating distributions and then to find the member of the family that is closest to the posterior.
Typically, closeness is defined by the \gls{KL} divergence between the approximating distribution and the true posterior. 

Concretely, consider the canonical running example in Eq.\nobreakspace \ref {eq:general_pgm}. 
Denote by $q(\bbeta, \bz_{1:N}; \blambda)$ the approximating family, also called 
the \textit{variational family}; it is indexed by $\blambda$, the \textit{variational parameters}. \gls{VI} solves the following optimization 
procedure:
\begin{align}\label{eq:klvi}
	\blambda^* 
	&= \arg\min_{\blambda} \gls{KL}\left(q(\bbeta, \bz_{1:N}; \blambda) \vert\vert p(\bbeta, \bz_{1:N} \vert \bx_{1:N})\right). 
\end{align}
The \gls{KL} above is intractable because the posterior is intractable. However, we can write the \gls{KL} as follows,
\begin{align}\label{eq:kl-elbo}
	\gls{KL}\left(q(\bbeta, \bz_{1:N}; \blambda) \vert\vert p(\bbeta, \bz_{1:N} \vert \bx_{1:N})\right)
	&= \log p(\bx_{1:N}) - \mathbb{E}_{q(\bbeta, \bz_{1:N}; \blambda)}\left[ 
		\log \frac{p(\bx_{1:N}, \bz_{1:N}, \bbeta)}{q(\bbeta, \bz_{1:N}; \blambda)} 
	\right]
\end{align}
This expression of the \gls{KL} reveals two things. First, because $\log p(\bx_{1:N})$ does not depend 
on the parameters $\blambda$ we want to optimize over, minimizing the \gls{KL} is equivalent 
to maximizing the second term on the right hand side of Eq.\nobreakspace \ref {eq:kl-elbo}. 
Second, because \gls{KL} is nonnegative, the second term on the right hand side of Eq.\nobreakspace \ref {eq:kl-elbo}, called the \gls{ELBO}, 
is a lower bound of the log marginal likelihood of the data $\log p(\bx_{1:N})$. 
The \gls{ELBO} is a function of the data and the variational parameters $\blambda$,
\begin{align}
	\gls{ELBO}(\bx_{1:N}, \blambda)
	&= \mathbb{E}_{q(\bbeta, \bz_{1:N}; \blambda)}\left[ 
		\log \frac{p(\bx_{1:N}, \bz_{1:N}, \bbeta)}{q(\bbeta, \bz_{1:N}; \blambda)} 
	\right] \leq \log p(\bx_{1:N})
\end{align}
The \gls{ELBO} is tractable, or can be tractably approximated, if we specify a tractable density for $q(\bbeta, \bz_{1:N}; \blambda)$.
There are many ways to specify the family $q(\bbeta, \bz_{1:N}; \blambda)$. One way to specify $q(\bbeta, \bz_{1:N}; \blambda)$ is to use the \textit{mean field} assumption. 

\parhead{Mean-field \gls{VI}.} The mean field assumption decomposes the variational distribution $q(\bbeta, \bz_{1:N}; \blambda)$ 
into a product of factors,
\begin{align}
	q(\bbeta, \bz_{1:N}; \blambda)
	&= q(\bbeta; \blambda_{\beta}) \cdot \prod_{i=1}^{N} q(\bz_i; \blambda_i)
\end{align}
where $\blambda = (\blambda_{\beta}, \blambda_1, \dots, \blambda_N)$. 
Using this decomposition, mean-field \gls{VI} then maximizes the \gls{ELBO}, 
\begin{align}
	\gls{ELBO}(\bx_{1:N}, \blambda)
	&= \mathbb{E}_{q(\bbeta; \blambda_{\beta})  \prod_{i=1}^{N} q(\bz_i; \blambda_i)}\left\{
		\log \frac{p(\bbeta)}{q(\bbeta)} + \sum_{i=1}^{N} \log \frac{p(\bz_i \vert \bbeta)}{q(\bz_i; \blambda_i)} 
		+ \sum_{i=1}^{N} \log p(\bx_i \vert \bz_i, \bbeta)
	\right\}
\end{align}
For certain classes of models, e.g. conditionally conjugate models~\citep{ghahramani2001propagation}, the \gls{ELBO} can be optimized using coordinate ascent or stochastic optimization. 
\citet{Blei2017} provide a review. 

\parhead{\Gls{BBVI}.} For a general class of models, the \gls{ELBO} can be maximized 
using \gls{BBVI}~\citep{Paisley2012, Ranganath2014}. 
For simplicity, let's lump all latent variables into $\bz$ and all the observations into $\bx$. The \gls{ELBO} is,
\begin{align}\label{eq:simple_elbo}
	\gls{ELBO} &= \mathbb{E}_{q(\bz; \blambda)}\left[\log p(\bx, \bz) - \log q(\bz; \blambda) \right]
\end{align}
\gls{BBVI} optimizes the \gls{ELBO} with respect to $\blambda$ using a Monte Carlo approximation of its gradients. 

\textit{Score gradients.} We can compute the gradient of the \gls{ELBO} with respect to $\blambda$ as follows.
\begin{align}
	\nabla_{\blambda}\gls{ELBO} 
	&= \nabla_{\blambda} \int_{}^{} \left[q(\bz; \blambda) \log p(\bx, \bz) - q(\bz; \blambda)\log q(\bz; \blambda)\right] d\bz\\
	&=\int_{}^{}  \left[\nabla_{\blambda}q(\bz; \blambda) \log p(\bx, \bz) - \nabla_{\blambda} \left(q(\bz; \blambda)\log q(\bz; \blambda) \right) \right] d\bz\\
	&= \int_{}^{}  \left[ \log p(\bx, \bz) \nabla_{\blambda}q(\bz; \blambda) -  \log q(\bz; \blambda) \nabla_{\blambda} q(\bz; \blambda)  - q(\bz; \blambda) \nabla_{\blambda} \log q(\bz; \blambda)\right]d\bz\\
	&=  \int_{}^{}  \left[ \log p(\bx, \bz) -  \log q(\bz; \blambda)\right] \nabla_{\blambda}q(\bz; \blambda) d\bz  -  \int_{}^{} \nabla_{\blambda} q(\bz; \blambda)d\bz\\
	&=  \int_{}^{} q(\bz; \blambda)  \left[ \log p(\bx, \bz) -  \log q(\bz; \blambda)\right] \nabla_{\blambda} \log q(\bz; \blambda) d\bz - \nabla_{\blambda} \int_{}^{} q(\bz; \blambda)d\bz\\
	&= \mathbb{E}_{q(\bz; \blambda)}\left[ \left(\log p(\bx, \bz) -  \log q(\bz; \blambda)\right) \nabla_{\blambda} \log q(\bz; \blambda) \right] \label{eq:score_grad}
\end{align}
where we used the identities $ \int_{}^{} q(\bz; \blambda)d\bz = 1$ and $\nabla_{\blambda} \log q(\bz; \blambda) = \frac{\nabla_{\blambda} q(\bz; \blambda)}{q(\bz; \blambda)}$. 
The expectation in Eq.\nobreakspace \ref {eq:score_grad} can be approximated using Monte Carlo, by averaging the quantity inside the expectation evaluated at different samples 
$\bz^{(1)} \dots \bz^{(S)}$ from $q(\bz; \blambda)$,
\begin{align}\label{eq:noisy_score}
	\nabla_{\blambda}\gls{ELBO} 
	&\approx \frac{1}{S} \sum_{s=1}^{S}\left(\log p(\bx, \bz^{(s)}) -  \log q(\bz^{(s)}; \blambda)\right) \nabla_{\blambda} \log q(\bz^{(s)}; \blambda)
\end{align}
The estimator in Eq.\nobreakspace \ref {eq:noisy_score} is an unbiased and consistent estimator of the true score gradient in Eq.\nobreakspace \ref {eq:score_grad}. However, 
it has high variance, especially when the dimensionality of the latents is large enough. 
Researchers have developed other gradient approximation methods, for example using 
Rao-Blackwellization~\citep{casella1996rao,Ranganath2014} or control variates~\citep{givens2012computational, Paisley2012}.

Throughout this dissertation, we will rely on \textit{reparameterization}~\citep{givens2012computational, kingma2014autoencoding}, a 
simple method to approximate gradients of Monte Carlo objectives, which we discuss next. 

\textit{Reparameterization gradients.} An alternative way to compute gradients of the \gls{ELBO} in Eq.\nobreakspace \ref {eq:simple_elbo} is to introduce 
variables $\bepsilon$ whose distribution $q(\bepsilon)$ is free from the variational parameters $\blambda$ and such that 
\begin{align}
	\bz \sim q(\bz; \blambda) \iff \bepsilon \sim q(\bepsilon) \text{ and } \bz = g(\bepsilon; \blambda)
\end{align}
where $g(\cdot)$ is a function that composes $\bepsilon$ and $\blambda$ into samples from the variational distribution.
Under this reparameterization procedure, the \gls{ELBO} takes the form
\begin{align}\label{eq:simple_elbo}
	\gls{ELBO} &= \mathbb{E}_{q(\bepsilon)}\left[\log p(\bx, g(\bepsilon; \blambda)) - \log q(g(\bepsilon; \blambda); \blambda) \right]
\end{align}
The gradient of the \gls{ELBO} is therefore,
\begin{align}\label{eq:simple_elbo}
	\nabla_{\blambda}\gls{ELBO} &= \mathbb{E}_{q(\bepsilon)} \nabla_{\blambda}\left[\log p(\bx, g(\bepsilon; \blambda)) - \log q(g(\bepsilon; \blambda); \blambda) \right] 
\end{align}
and can be simply approximated using Monte Carlo,
\begin{align}\label{eq:simple_elbo}
	\nabla_{\blambda}\gls{ELBO} &\approx \frac{1}{S} \sum_{s=1}^{S} \nabla_{\blambda}\left[\log p(\bx, g(\bepsilon^{(s)}; \blambda)) - \log q(g(\bepsilon^{(s)}; \blambda); \blambda) \right] 
\end{align}
where $\bepsilon^{(1)} \dots \bepsilon^{(S)} \sim q(\bepsilon)$. In all our experiments, we set $S=1$, which has been shown enough for learning~\citep{kingma2014autoencoding}.

\section{Deep Learning}\label{sec:dl}
\gls{DL} is a framework for learning from complex high-dimensional large-scale data. 
It has been very successful in the domain of supervised learning, where data are labeled. 
In particular, \gls{DL} is very successful for vision and language applications, 
e.g. object detection, image captioning, machine translation, document classification, etc. 
While \gls{PGM} specifies the structure underlying the data through a set of latent variables~\citep{koller2009probabilistic}, 
\gls{DL} uses neural networks to capture the structure in data. 

\subsection{Neural Networks \& Flexibility} 
Neural networks are a hierarchy of nonlinear deterministic functions~\citep{lecun2015deep, goodfellow2016deep}. 
Consider $N$ i.i.d pairs $(\bx_i, y_i) \text{ for } i=1 \dots N$. 
A neural network with $L$ layers maps a given input $\bx_i$ to its output $y_i$ following a chain of transformations,
\begin{align}
	\bh_0 &= \bx_i \\
	\bh_l &= f_l(\bh_{l-1}; W_l)\\
	\bh_L &= f_L(\bh_{L-1}; W_L)\\
	y_i &\sim \text{EF}(\eta_i = g(V^\top \bh_L))
\end{align}
Here $\bh_{1:L}$ are called \textit{hidden states}. The $l^{th}$ hidden state is computed by composing 
an \textit{activation function} $f_l(\cdot)$ with some transformation of the output of the previous layer 
that uses the weights $W_l$. Different choices for the activation functions and the transformations 
yield different neural network architectures, we review some later. 
The weights represent the model parameters which we aim to learn. 
$\text{EF}(\eta)$ denotes an exponential family with natural parameter $\eta$. The function $g(\cdot)$ 
maps the dot product $V^\top \bh_L$ to the natural parameter space. For example when $y_i$ is in the reals, 
$g(\cdot)$ is identity and if $y_i$ is categorical then $g(\cdot) = \text{softmax}(\cdot)$. 

Neural networks have been shown to have the ability to represent any function~\citep{hornik1989multilayer}. 
They can therefore capture the complex dependencies in data without any feature engineering. 
It is this flexibility that has made \gls{DL} very successful at modeling large-scale complex data. 

However neural networks tend to overfit to data. \gls{DL} approaches often require large datasets to achieve 
good performance. Researchers have developed several regularization methods for neural
 networks~\citep{bishop1995training, maaten2013learning, srivastava2014dropout, gal2016dropout, dieng2018noisin}.  
 
 Neural networks are fit using stochastic gradient descent with backpropagation~\citep{rumelhart1986learning}.

\subsection{Example: Recurrent Neural Networks}
Consider a sequence of observations, $\bx_{1:T} = (\bx_1, ... ,\bx_T)$. A \gls{RNN}
factorizes its joint distribution according to the chain rule of probability,
\begin{align}
  \label{eq:chain_rule}
  p(\bx_{1:T}) = \prod_{t=1}^{T} p(\bx_t | \bx_{1: t-1}).
\end{align}
To capture dependencies, the \gls{RNN} expresses each conditional
probability as a function of a low-dimensional recurrent hidden state,
\begin{align*}
  \bh_t  &= f_W(\bx_{t-1}, \bh_{t-1}) \text{ and }
  p(\bx_{t} | \bx_{1: t-1}) = p(\bx_{t} | \bh_t).
\end{align*}

The likelihood $p(\bx_{t} | \bh_t)$ can be 
of any form. We focus on the exponential family 
\begin{align}\label{eq:likelihood}
	p(\bx_{t} | \bh_t)  &= \nu(\bx_t) \exp \left\{(V^\top \bh_t)^\top \bx_t - A(V^\top \bh_t) \right\}
	,
\end{align}
where $\nu(\cdot)$ is the base measure, $V^\top \bh_t$ is the natural 
parameter---a linear function of the hidden state $\bh_t$---and $A(V^\top \bh_t)$ is the 
log-normalizer. The matrix $V$ is called the \textit{prediction} or \textit{output} matrix 
of the \gls{RNN}. 

The hidden state $\bh_t$ at time $t$ is a parametric
  function $f_W(\bh_{t-1}, \bx_{t-1})$ of the previous hidden state
$\bh_{t - 1}$ and the previous observation $\bx_{t - 1}$; the
parameters $W$ are shared across all time steps. The function $f_W$ is 
the transition function of the \gls{RNN}, it defines a recurrence relation 
for the hidden states and renders $\bh_t$ a function of all the past 
observations $\bx_{1:t-1}$; these properties match the chain rule 
decomposition in Eq.\nobreakspace \ref {eq:chain_rule}. 

The particular form of $f_W$ determines the \gls{RNN}. Researchers
have designed many flavors, including the \gls{ERNN}~\citep{elman1990finding}, 
the \gls{LSTM}~\citep{hochreiter1997long} and the \gls{GRU}~\citep{cho2014learning}. 

\parhead{\Acrlong{ERNN}.} The \gls{ERNN} is the simplest \gls{RNN}. 
In an \gls{ERNN}, the transition function is
\begin{align*}
  f_W(\bx_{t-1}, \bh_{t-1}) =
  s(W_x^\top \bx_{t-1} + W_h^\top \bh_{t-1})
  ,
\end{align*}
where we dropped an intercept term to avoid cluttered notation. Here,
$W_h$ is called the \textit{recurrent weight matrix} and $W_x$ is
called the \textit{embedding matrix} or \textit{input matrix}.~The function $s(\cdot)$ is 
called an \textit{activation} or \textit{squashing} function, which stabilizes the
transition dynamics by bounding the hidden state. Typical choices for
the squashing function include the sigmoid and the hyperbolic tangent.

\parhead{\Acrlong{LSTM}.} 
The \gls{LSTM} was designed to avoid optimization issues, such as
vanishing (or exploding) gradients.  Its transition function composes
four \gls{ERNN}s, three with sigmoid activations and one with a
$\tanh$ activation:
\begin{gather}
  \label{eq:lstm-f}
  f_t = \sigma(W_{x1}^\top\bx_{t-1} + W_{h1}^\top\bh_{t-1}) \\
  i_t = \sigma(W_{x2}^\top\bx_{t-1} + W_{h2}^\top\bh_{t-1})\\
  o_t = \sigma(W_{x4}^\top\bx_{t-1} + W_{h4}^\top\bh_{t-1}) \\
  \bc_t = f_t \odot \bc_{t-1} + i_t \odot \tanh(W_{x3}^\top\bx_{t-1} + W_{h3}^\top\bh_{t-1}) \\
  \bh_t = o_t \odot \tanh(\bc_t) \label{eq:lstm_state}.
\end{gather}
Here $f_t$, $i_t$, and $o_t$ are called \textit{gates}. The \gls{LSTM} state is the pair $(\bc_t, \bh_t)$. 
The state $\bc_t$ is the \textit{memory cell}; it is designed to capture long-term dependencies~\citep{hochreiter1997long}. 
The gate $f_t$ is the \textit{forget gate}; it determines which part of memory to discard. The gate $i_t$ controls 
the amount of new information to add to the memory. Finally, the gate $o_t$ determines how the output state $\bh_t$ 
depends on the memory cell. 

\parhead{\Acrlong{GRU}.} 
\glspl{GRU} provide a simpler way to parameterize a \gls{RNN} than the \gls{LSTM}~\citep{cho2014learning}. 
The hidden state $\bh_t$ of a \gls{GRU} is computed as 
\begin{gather}
  \label{eq:lstm-f}
  u_t = \sigma(W_{x1}^\top\bx_{t-1} + W_{h1}^\top\bh_{t-1}) \\
  r_t = \sigma(W_{x2}^\top\bx_{t-1} + W_{h2}^\top\bh_{t-1})\\
  \bh_t =  u_t \odot \bh_{t-1} + (1 - u_t) \odot \tanh\left( W_{x3}^\top\bx_{t-1} + W_{h3}^\top( r_t \odot \bh_{t-1}) \right)
  \label{eq:gru_state}.
\end{gather}
Here $u_t$ is called an \textit{update} gate, it decides whether to change the 
previous configuration of the hidden state or not. The variable $r_t$ is called a \textit{reset gate} 
it indicates which coordinates of the previous hidden state are to be updated. 

\subsection{Example: Auto-Encoders} 
Auto-encoding is a successful \gls{DL} technique for dimensionality reduction~\citep{hinton2006reducing}. 
The idea is to learn to reconstruct data. Consider a dataset of $N$ i.i.d observations 
$\bx_1, \dots, \bx_N$. An \gls{AE} minimizes reconstruction error,
\begin{align}
	\mathcal{L}(\theta, \phi) 
	&= \sum_{i=1}^{N} \Vert \bx_i - f_{\theta} (g_{\phi}(\bx_i))  \Vert_2^2.
\end{align}
Here $g_{\phi}(\cdot)$ is called an \textit{encoder}, it takes a data point 
$\bx_i$ as input and outputs $\bh_i = g_{\phi}(\bx_i)$, which is a low-dimensional 
representation of $\bx_i$ called a \textit{code}.  The function $f_{\theta}(\cdot)$ 
is called a \textit{decoder}, it maps the code $\bh_i$ to the observation space. 
Its output is $\tilde{\bx}_i$ which is optimized to be close to $\bx_i$. 
\Glspl{AE} have been shown to learn better low-dimensional representations 
of data than \gls{PCA}~\citep{hinton2006reducing}. They have also been 
useful for other applications, e.g. image denoising~\citep{vincent2008extracting}. 

\subsection{Example: Word Embeddings}
Word embeddings provide models of language
that use vector representations of
words~\citep{Rumelhart:1973,Bengio:2003}. The word representations are
fitted to relate to meaning, in that words with similar meanings will
have representations that are close.  (In embeddings, the ``meaning''
of a word comes from the contexts in which it is used~\citep{harris1954distributional}.)

We focus on the \gls{CBOW} variant of word embeddings
\citep{mikolov2013distributed}. In \gls{CBOW}, the likelihood of each
word $w_{dn}$ is
\begin{equation}\label{eq:cbow}
  w_{dn} \sim \textrm{softmax}(\rho^\top \alpha_{dn}).
\end{equation}
The embedding matrix $\rho$ is a $L\times V$ matrix whose columns
contain the embedding representations of the vocabulary,
$\rho_v\in\mathbb{R}^L$. The vector $\alpha_{dn}$ is the \textit{context
  embedding}. The context embedding is the sum of the context
embedding vectors ($\alpha_v$ for each word $v$) of the words surrounding
$w_{dn}$.

\section{Combining Neural Networks and Latent Variables}
\gls{PGM} and \gls{DL} both aim to learn from data. While \gls{PGM} 
uses latent variables to express the structure underlying data in an interpretable way, 
\gls{DL} uses neural networks to express the structure underlying data in a flexible way. 
These two approaches of learning from data are complementary. Researchers 
have developed methods that combine neural networks and latent 
variables~\citep{kingma2013auto, rezende2014stochastic, Johnson2016composing, gao2016linear, krishnan2017structured}. 

\subsection{Probabilistic Conditioning with Neural Networks}
Probabilistic conditioning with neural networks is a way to combine neural networks 
and latent variables. There are two ways to do this:
\begin{itemize}
	\item Use the output of a neural network that takes the latent variables as input to define the parameters of the conditional distribution of the data given the latent variables.
	\item Use the output of a neural network that takes data as input to define the parameters of the posterior distribution of the latent variables given the data. 
\end{itemize}

More concretely, consider our running \gls{PGM} example, the \gls{EF-PCA}. 
It posits a shared global latent structure $\bbeta$ and a per-observation local latent structure $\bz_{1:N}$. 
A data point $\bx$ is drawn by conditioning on both $\bbeta$ and $\bz$. 
In \gls{EF-PCA}, the global and local structure interact linearly in the likelihood $p(\bx \g \bz, \bbeta)$. (See Section\nobreakspace \ref {subsec:efpca}.) 
Allowing non-linear interactions between global and local structure will make the model 
more flexible. This can be achieved using neural networks, the parameters of which will be shared 
across the observations to model the global structure. The conditional distribution 
of $\bx$ given $\bz$ is then
\begin{align}\label{eq:vae_model}
	p_{\bbeta}(\bx, \bz) &= p(\bx \g f_{\bbeta}(\bz)) \cdot p(\bz)
\end{align}
where $f_{\bbeta}(\cdot)$ is the neural network that defines the likelihood. 

On the other hand, the local latent variables $\bz_{1:N}$ can be seen as low-dimensional representations 
of the data. Each observation has its own representation in the latent space. We saw in Section\nobreakspace \ref {sec:dl} 
that \glspl{AE} are good at finding low-dimensional representations of data. We can use auto-encoding 
to define the posterior distributions of the local latent variables,
\begin{align}\label{eq:vae_posterior}
	q_{\phi}(\bz \g \bx) &= q_{\phi}(\bz \g g_{\phi}(\bx)) 
\end{align}
where $g_{\phi}(\cdot)$ is the neural network that parameterizes the posterior, the encoder in an \gls{AE}. Note $q_{\phi}(\bz \g \bx)$ 
is not the true posterior distribution of the latent variables, but we can use it within the framework 
of \gls{VI} to learn approximations of the true posterior. 

\subsection{Variational Auto-Encoders} 
\Glspl{VAE} tie model design and posterior inference within one framework. 
They use Eq.\nobreakspace \ref {eq:vae_model} as a model for data and Eq.\nobreakspace \ref {eq:vae_posterior} as an approximate posterior over latent variables. 
More specifically, the likelihood is an exponential family whose natural parameter $\eta(\bz; \bbeta)$ is computed as follows:
\begin{enumerate}
  \item $\bh^{(1)} = f_{\bbeta_0}(\bz)$ 
  \item $\bh^{(l+1)} = f_{\bbeta{l}}\left(\bh^{(l)}\right) \quad l = 1 \dots L-1$ 
  \item $\eta(\bz; \theta) = f_{\bbeta_L}\left(\bh^{(L)}\right)$.
\end{enumerate}
The parameter $\bbeta$ is the collection $\{\bbeta_0, \dots, \bbeta_L\}$. The output $\bh^{(l+1)}$ of the $(l+1)^{th}$  
layer is computed by composing the output $\bh^{(1)}$ of the previous layer with layer-specific parameters $\bbeta{l}$. 

\Glspl{VAE} use stochastic gradient ascent to learn both sets of parameters $\bbeta$ and $\phi$. The objective is the \gls{ELBO},
\begin{align}\label{eq:elbovae}
	\gls{ELBO}(\bbeta, \phi) 
	&= \mathbb{E}_{q_{\phi}(\bz \g \bx)} \left[
		\log p_{\bbeta}(\bx, \bz) - \log q_{\phi}(\bz \g \bx)
	\right]
	.
\end{align}
For a given setting of $\phi$, maximizing the \gls{ELBO} with respect to $\bbeta$ corresponds to maximizing the likelihood 
of the observations. 
For a given setting of the model parameters $\bbeta$, maximizing the \gls{ELBO} with respect to $\phi$ can be interpreted in two different ways. 

\parhead{\gls{KL} minimization perspective.} 
We can write the \gls{ELBO} as 
\begin{align}
	\gls{ELBO}(\bbeta, \phi) 
	&= \mathbb{E}_{q_{\phi}(\bz \g \bx)} \left[
		\log p_{\bbeta}(\bz \g \bx) + \log p_{\bbeta}(\bx) - \log q_{\phi}(\bz \g \bx)
	\right]
	.
\end{align}
Since $\log p_{\bbeta}(\bx)$ does not depend on $\phi$, maximizing the \gls{ELBO} with respect to $\phi$ is equivalent 
to minimizing the \gls{KL} between $q_{\phi}(\bz \g \bx)$ and the true posterior $p_{\bbeta}(\bz \g \bx)$. 
Indeed, 
\begin{align}
	\gls{ELBO}(\bbeta, \phi) 
	&= -\gls{KL}(q_{\phi}(\bz \g \bx) \vert\vert p_{\bbeta}(\bz \g \bx)) + cst
	.
\end{align}

\parhead{Regularized autoencoder perspective.} 
Maximizing the \gls{ELBO} with respect to $\phi$ can also be seen as regularizing an \gls{AE}. 
Rewrite the \gls{ELBO} as follows
\begin{align}
	\gls{ELBO}(\bbeta, \phi) 
	&= \mathbb{E}_{q_{\phi}(\bz \g \bx)} \left[
		\log p_{\bbeta}(\bx \g \bz) 	\right] - \gls{KL}(q_{\phi}(\bz \g \bx) \vert\vert p(\bz))
	.
\end{align}
Assume without loss of generality that the likelihood is Gaussian with identity variance 
and that $q_{\phi}(\bz \g \bx)$ is also a Gaussian with identity variance. 
Assume we draw one sample $\bz_{\phi}(\bx) = g_{\phi}(\bx) + \bepsilon$ from $q_{\phi}(\bz \g \bx)$ 
where $\bepsilon \sim \mathcal{N}(\bzero, \bI)$. Assume the prior is standard Gaussian. The \gls{ELBO} is 
\begin{align}
	\gls{ELBO}(\bbeta, \phi) 
	&= \Vert \bx - f_{\bbeta} (\bz_{\phi}(\bx))  \Vert_2^2 - \frac{1}{2}\Vert g_{\phi}(\bx) \Vert_2^2
	.
\end{align}
When maximizing the \gls{ELBO} with respect to $\phi$, the first term is the objective of an \gls{AE}, 
it forces to learn settings of $\phi$ that are able to reconstruct the data well. The second term 
regularizes the parameters $\phi$ such that the output of the encoder have bounded  $L_2$ norm. 

In fact there is a second source of regularization in this particular case; noise $\bepsilon$ 
is first added to the code $g_{\phi}(\bx)$ to get the \textit{latent} code $\bz_{\phi}(\bx)$, which is then used as input to the decoder $f_{\bbeta}(\cdot)$. 
This added noise trades off some reconstruction error with the ability to simulate new data 
from the fitted decoder. 

\parhead{Amortized variational inference and mean field variational inference.} 
By using neural networks to define the approximate posterior over the latent variables, 
\glspl{VAE} are doing \gls{AVI}. The term comes from the fact that computing the approximate posterior 
boils down to passing data through a shared neural network, which \textit{amortizes} the cost of inference 
for models with local latent variables when dealing with large datasets. 

How does \gls{AVI} relate to mean field \gls{VI}? 
Consider our canonical \gls{EF-PCA} example. Mean field \gls{VI} uses the factorization 
\begin{align}
	q(\bbeta, \bz_{1:N}; \blambda)
	&= q(\bbeta; \blambda_{\beta}) \cdot \prod_{i=1}^{N} q(\bz_i; \blambda_i)
\end{align}
Mean field assumes all latent variables are independent, both local and global. 
We can relax this a bit and assume the local latent variables $\bz_{1:N}$ are 
conditionally independent given the global latent variables,
\begin{align}
	q(\bbeta, \bz_{1:N}; \blambda)
	&= q(\bbeta; \blambda_{\beta}) \cdot \prod_{i=1}^{N} q(\bz_i \g \bbeta; \blambda_i)
\end{align}
We let each factor explicitly condition on data, 
\begin{align}
	q(\bbeta, \bz_{1:N}; \blambda)
	&= q(\bbeta; \blambda_{\beta}) \cdot \prod_{i=1}^{N} q(\bz_i \g \bx_i, \bbeta; \blambda_i)
\end{align}
A way to get to \gls{AVI} is to assume that the 
global structure $\bbeta$ represents a posteriori a neural network with parameters $\blambda_{\beta}$ 
and let each latent $\bz_i$ have its own distribution only through $\bx_i$. Then we end up with 
\begin{align}
	q(\bbeta, \bz_{1:N}; \blambda)
	&=  \prod_{i=1}^{N} q(\bz_i \g \bx_i, \blambda_{\beta})
\end{align}
where the conditioning on $\bx_i$ and $\blambda_{\beta}$ corresponds to passing $\bx_i$ 
through the neural network $\blambda_{\beta}$. 

\subsection{Challenges: Latent Variable Collapse} 
The \gls{ELBO} in Eq.\nobreakspace \ref {eq:elbovae} is intractable because of the expectations. 
\glspl{VAE} leverage \gls{BBVI}~\citep{Paisley2012, Ranganath2014} and approximate the \gls{ELBO} 
with Monte Carlo samples from the variational distribution. To reduce variance 
of the gradients of the \gls{ELBO}, \glspl{VAE} use reparameterization~\citep{kingma2013auto, rezende2014stochastic}. 
This procedure often empirically leads to a degenerate solution where
\begin{align*}
  q_{\phi}(\bz \g \bx) &\approx p(\bz).
\end{align*}
The variational ``posterior'' does not depend on the data; this is referred to as \textit{latent variable collapse}. 
When the approximate posterior is close to the prior, posterior estimates of the latent variable $\bz$ 
do not represent faithful summaries of their data $\bx$---the \gls{VAE} has not learned good representations. 
This issue is discussed in several papers \citep{bowman2015generating, sonderby2016train, kingma2016improved, chen2016variational, zhao2017towards,yeung2017tackling}. 
In Chapter\nobreakspace \ref {chap:dpgm} we provide a solution to this problem.

\glsresetall

\chapter{Deep Probabilistic Graphical Modeling}\label{chap:dpgm}
The previous chapter laid out the foundations for \gls{DPGM}. 
We reviewed latent-variable graphical models and their inference. 
These models have an interpretable probabilistic structure 
and can be fit using \gls{VI}. However graphical models tend to lack flexibility, 
which hinders their use when it comes to modeling high-dimensional complex data 
and/or performing tasks that require flexibility (e.g. in vision and language applications.) 

We reviewed \gls{DL}, a paradigm that offers flexibility both in terms of model specification 
and model fitting by leveraging neural networks and backpropagation.  
Although flexible, \gls{DL} does not offer the same interpretability as \gls{PGM}. 

Finally, we described ways to combine neural networks and latent variables and the \emph{latent variable collapse} 
issue that might arise from it leading to a non-interpretable latent structure. 

In this chapter, we develop \gls{DPGM}, a set of methodologies that leverage \gls{DL} for \gls{PGM}. 
\gls{DPGM} benefits from the interpretability of \gls{PGM} and the flexibility of \gls{DL}. 
We first discuss three desiderata for \gls{DPGM} before describing several instances of \gls{DPGM}. 
One instance extends \gls{EF-PCA} using deep neural networks while 
preserving the interpretability of the latent factors. Another instance corresponds to a model class 
for sequential data that allows to account for long-range dependencies. Finally, 
we show how \gls{DPGM} solves several problems of probabilistic topic models. 

\section{Desiderata}
The goal of \gls{DPGM} is to make \gls{PGM} more flexible.  
This leads to three desiderata for \gls{DPGM}, which we discuss in more detail. 
\begin{enumerate}
	\item \textit{Generalization}. Data are finite. We require systems built using \gls{DPGM} to 
	generalize beyond the observed data. Here, the term ``generalization" 
	encapsulates two things: (1) the capacity to assign high probability to unobserved data arising from the 
	same distribution as the training data\footnote{this is the usual meaning of the word ``generalization" in Machine Learning.} 
	and (2) the ability to yield good simulations of new data. The latter pertains to the diversity and visual quality 
	of data generated from the fitted model.  
	We will measure generalization using held-out predictive log-likelihood or measures of simulation quality.  
	\item \textit{Interpretability}. \gls{PGM} uncovers the hidden structure underlying data through a set of latent variables. 
	These latent variables capture meaning and are interpretable. We require the same interpretability for \gls{DPGM}. 
	However, flexibility often comes in the way of interpretability. The \textit{latent variable collapse} problem described 
	in Chapter\nobreakspace \ref {chap:foundations} is one manifestation of this. The \gls{DPGM} instances we describe in Section\nobreakspace \ref {sec:skipvae}, 
	Section\nobreakspace \ref {sec:topicrnn}, and Section\nobreakspace \ref {sec:etm} will offer both flexibility and interpretability.  
	We will measure interpretability of the learned latent variables using proxies of mutual 
	information or performance on a downstream classification task.  
	\item \textit{Scalability}. The data we deal with are large-scale and high-dimensional. We require \gls{DPGM} 
	methods to scale both in terms of the number of observations and the dimensionality of each observation. 
\end{enumerate}

\section{From Exponential Family PCA to Deep Generative Skip Models}\label{sec:skipvae}

We introduce \glspl{DGSM}, a class of models that extends \gls{EF-PCA} using neural networks. 
\glspl{DGSM} are efficiently fit using \gls{AVI}~\citep{Gershman2014, kingma2013auto, rezende2014stochastic}. 
When evaluated on image and text data, they achieve higher predictive performance and 
learn more interpretable latent factors than several baselines. 

\subsection{Model Class} 

Assume observed $N$ i.i.d data points $\bx_1, \dots, \bx_N$ where $\bx_i \in \mathbb{R}^D$. 
Consider the data generative process of \gls{EF-PCA}:
\begin{enumerate}
	\item Draw global latents $\bbeta \sim p(\bbeta)$
	\item For each data point $i = 1 \dots N$:
	 \begin{enumerate}
	 	\item Draw local latent variable $\bz_i \sim p(\bz)$
		\item Draw data point $\bx_i \sim \text{EF}(\eta_i = f(\bbeta^\top\bz_i))$
	 \end{enumerate}
\end{enumerate}

As discussed in Chapter\nobreakspace \ref {chap:foundations}, the global latent variables $\bbeta$ capture 
features shared across all the observations. We can infer them using variational inference. 
In this section, we set $\bbeta$ to be deterministic parameters of a shared deep neural network. 
\glspl{DGSM} define the following generative process for data,
\begin{enumerate}
	\item For each data point $i = 1 \dots N$:
	 \begin{enumerate}
	 	\item Draw local latent variable $\bz_i \sim p(\bz)$
		\item Draw data point $\bx_i \sim \text{EF}(\eta_i = f_\bbeta(\bz_i))$.
	 \end{enumerate}
\end{enumerate}
Here, the natural parameter is the output of a neural network $f_\bbeta(\cdot)$ that takes $\bz_i$ as input. 
It is computed through the following chain:
\begin{enumerate}
    \item $\bh_i^{(1)} = f_{\theta_0}(\bz_i)$
    \item $\bh_i^{(l+1)} = g_{W_l}\left(f_{\theta_l}\left(\bh_i^{(l)}\right), \bz_i\right)$ for $l = 1 \dots L-1$
    \item $\eta_i = g_{W_L}\left(f_{\theta_L}\left(\bh_i^{(L)}\right), \bz_i\right)$
.
\end{enumerate}
Here $\bbeta = (W_1, \dots, W_L, \theta_0, \dots, \theta_L)$ 
The functions $g_{W_1}(\cdot), \dots, g_{W_L}(\cdot)$ and $f_{\theta_0}(\cdot), \dots, f_{\theta_L}(\cdot)$ 
define the neural network $f_\bbeta(\cdot)$. Their choices lead to different architectures. 
At a given layer $l$, the hidden state $\bh_i^{(l)}$ of the neural network $f_\bbeta(\cdot)$ 
is a function of both the latent variable $\bz_i$ and the hidden state from the previous layer $\bh_i^{(l-1)}$. 
The dependence on $\bz_i$, via $g_{W_1}(\cdot), \dots, g_{W_L}(\cdot)$, is called a \textit{skip connection}. 
Skip connections are widely used in \gls{DL}, for example, in designing residual, highway, and attention networks~\citep{fukushima1988neocognitron,he2016identity,srivastava2015highway,bahdanau2014neural}. 
Here we use them to define \glspl{DPGM} to enforce a stronger dependence between the latent variables 
and the observations. 

\glspl{DGSM} are amenable to any type of skip functions; in this section we consider 
\begin{align*}
g_{W_l}\left(f_{\theta_l}(\bh_i^{(l)}), \bz_i\right) &=
\sigma_l\left(
W_l^{(h)}f_{\theta_l}(\bh_i^{(l)}) + W_l^{(z)}\bz_i
\right)
\end{align*}
where $\sigma_l$ is a typical nonlinear function such as sigmoid or ReLU. We set $\sigma_L(\cdot)$, the activation at the last layer, 
to identity. The weights $W_l^{(h)} \neq \bzero$ and $W_l^{(z)} \neq \bzero$ are parameters of the model.  
Similarly to other uses of skip connections~\citep{fukushima1988neocognitron,he2016identity,srivastava2015highway,bahdanau2014neural} 
we do not need to explicitly enforce the constraints $W_l^{(h)} \neq \bzero$ and $W_l^{(z)} \neq \bzero$ in practice. 

\glspl{DGSM} are amenable to any neural network architecture. 
Defining $f_{\theta_0}(\cdot), \dots, f_{\theta_L}(\cdot)$ corresponds to specifying a neural network architecture. 
In our empirical study we explore \glspl{CNN} for image applications and \glspl{LSTM} for text applications.  

\subsection{Amortized Variational Inference}

\glspl{DGSM} are fit using \gls{AVI}. For that we define a variational distribution over the latent variables, 
\begin{align}
	q_{\phi}(\bz_{1:N} \vert \bx_{1:N})
	&= \prod_{i=1}^{N} q_{\phi}(\bz_i \vert \bx_i)
\end{align}
We set each factor $q_{\phi}(\bz_i \vert \bx_i)$ as a Gaussian whose mean and Covariance 
are the given by the output of a neural network that takes $\bx_i$ as input, 
\begin{align}
	q_{\phi}(\bz_i \vert \bx_i)
	&= \mathcal{N}(\bmu_i(\bx_i; \phi), \Sigma_i(\bx_i; \phi))
\end{align}
Here $\bmu_i(\bx_i; \phi_{\mu})$ and $\Sigma_i(\bx_i; \phi_{\Sigma})$ are the neural networks 
for the mean and the covariance respectively and $\phi = (\phi_{\mu}, \phi_{\Sigma})$. 
In practice we use one shared neural network whose output is mapped to one of two sets of weights to get the mean or the covariance. 
We use $\text{softplus}(a) = \log(1 + \exp(a))$ as the final activation function when computing the covariance.  

Note the variational distribution can also be parameterized using the same approach we used to define the model. 
More concretely, when computing the mean and the covariance of the variational distributions, 
we can add skip connections from an input $\bx_i$ to each layer of the inference network. 

We now can form the \gls{ELBO}, 
\begin{align}\label{eq:elboskip}
	\gls{ELBO} = \sum_{i=1}^{N}\mathbb{E}_{q_{\phi}(\bz_i \vert \bx_i)} \left[
		\log p_{\bbeta}(\bx_i \vert \bz_i) - \acrshort{KL}(q_{\phi}(\bz_i \vert \bx_i) \vert\vert p(\bz_i))
	\right]
\end{align}
The \gls{ELBO} is intractable but we can estimate it using Monte Carlo with the 
reparameterization trick~\citep{kingma2014autoencoding}. 
We can then optimize the \gls{ELBO} with respect to both the model parameters $\bbeta$ and 
the variational parameters $\phi$.

\begin{figure*}[htp]
  \centering
  {\includegraphics[scale=0.25]{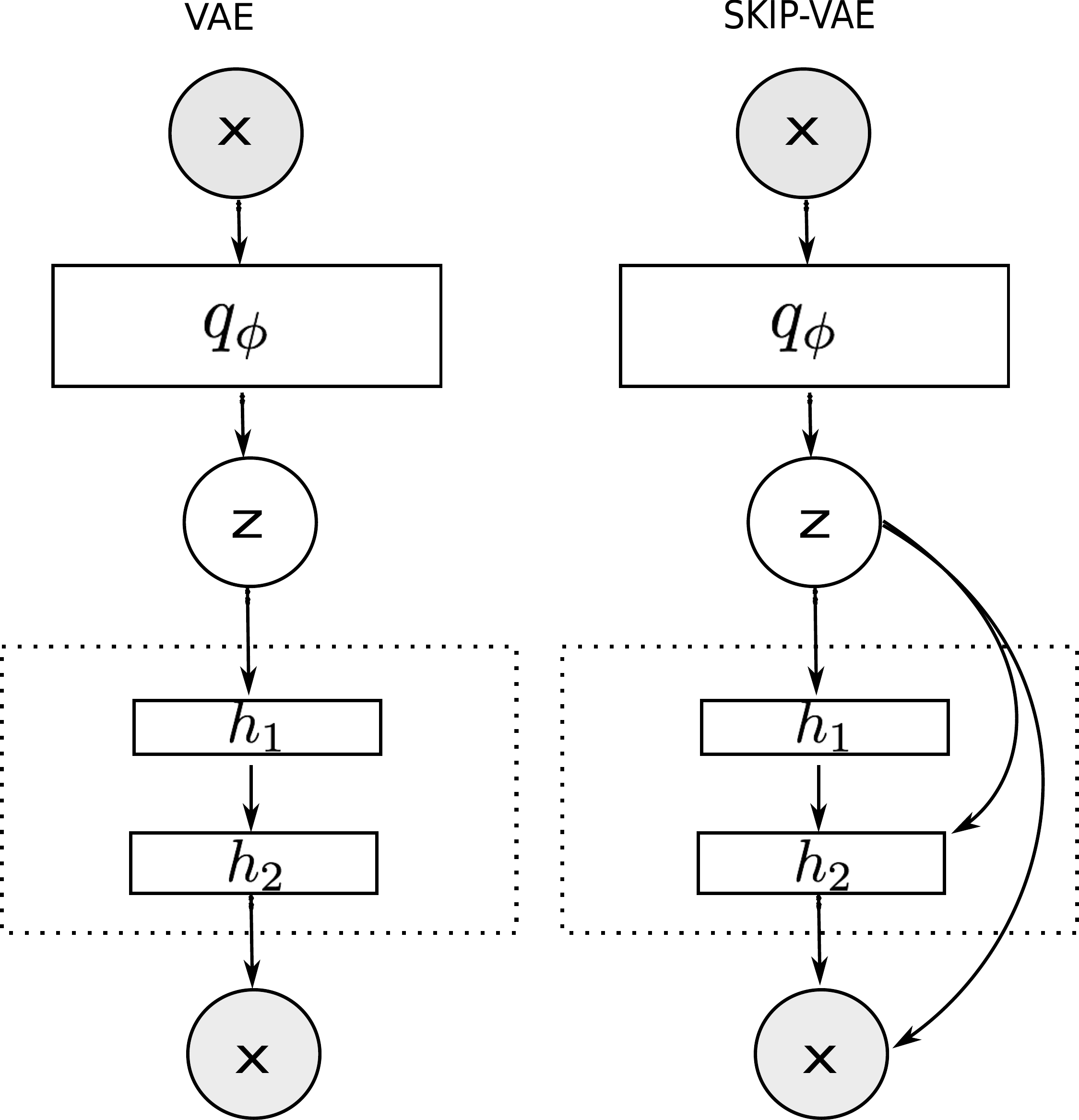}}\quad\quad\quad\quad\quad
  {\includegraphics[scale=0.5]{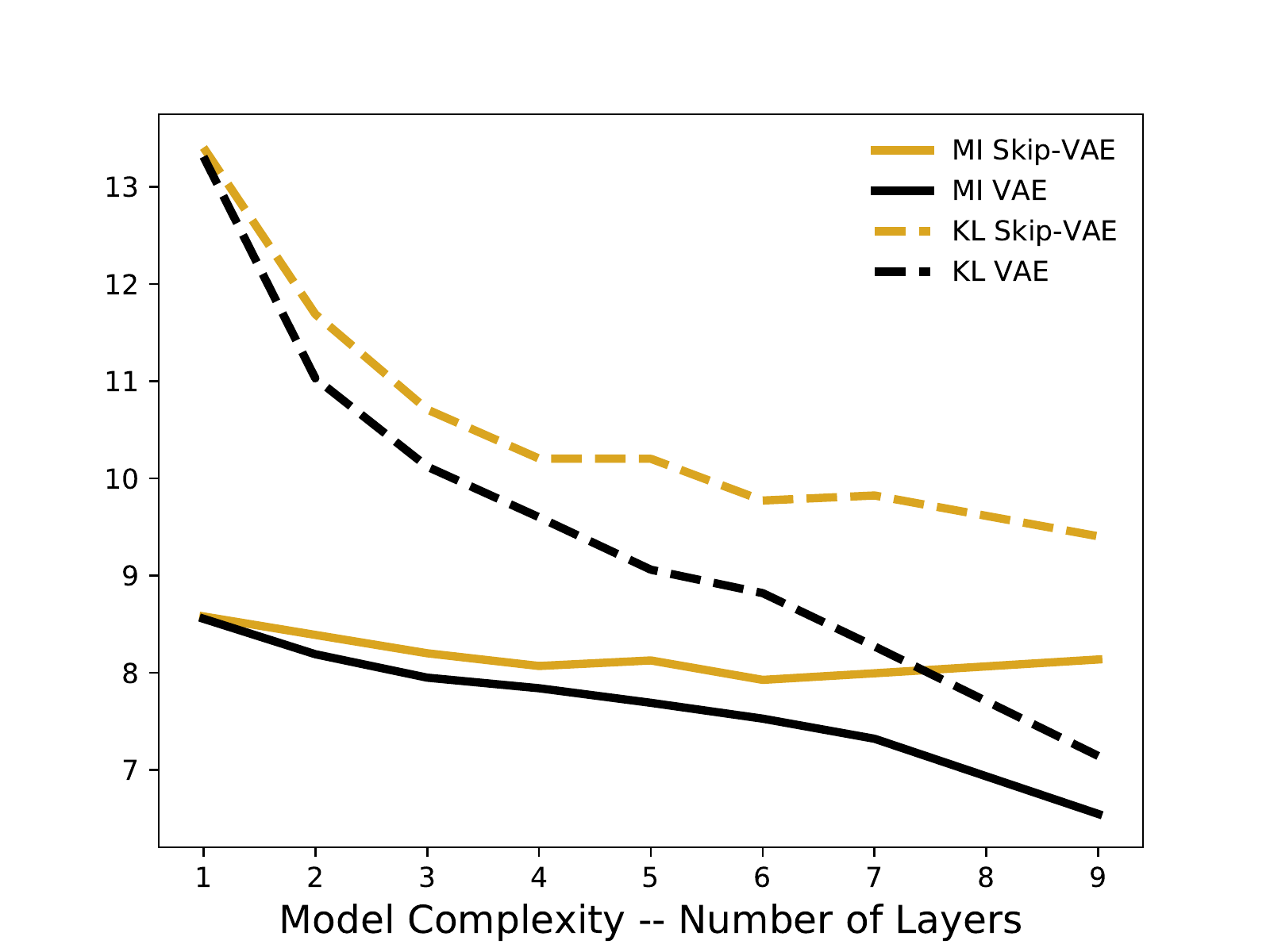}}
  \caption[Comparing the interpretability of the representations learned by a \gls{VAE} and a \gls{SkipVAE} as the decoder gets more flexible]{\textbf{Left:} The \gls{VAE} and \gls{SkipVAE} with a two-layer generative model. 
The function $q_{\phi}$ denotes the variational neural network (here identical for \gls{VAE} and \gls{SkipVAE}). 
The difference is in the generative model class: the \gls{SkipVAE}'s generative model enforces 
residual paths to the latents at each layer. \textbf{Right:} The mutual information induced 
by the variational distribution and $\kl$ from the variational distribution to the prior 
for the \gls{VAE} and the \gls{SkipVAE} on MNIST as we vary the number of 
layers $L$. The \gls{SkipVAE} leads to both higher $\kl$ and higher mutual information.}
\label{fig:model}
\end{figure*}

\begin{figure}[t]
\center
\includegraphics[width=0.49\linewidth, height=5.5cm]{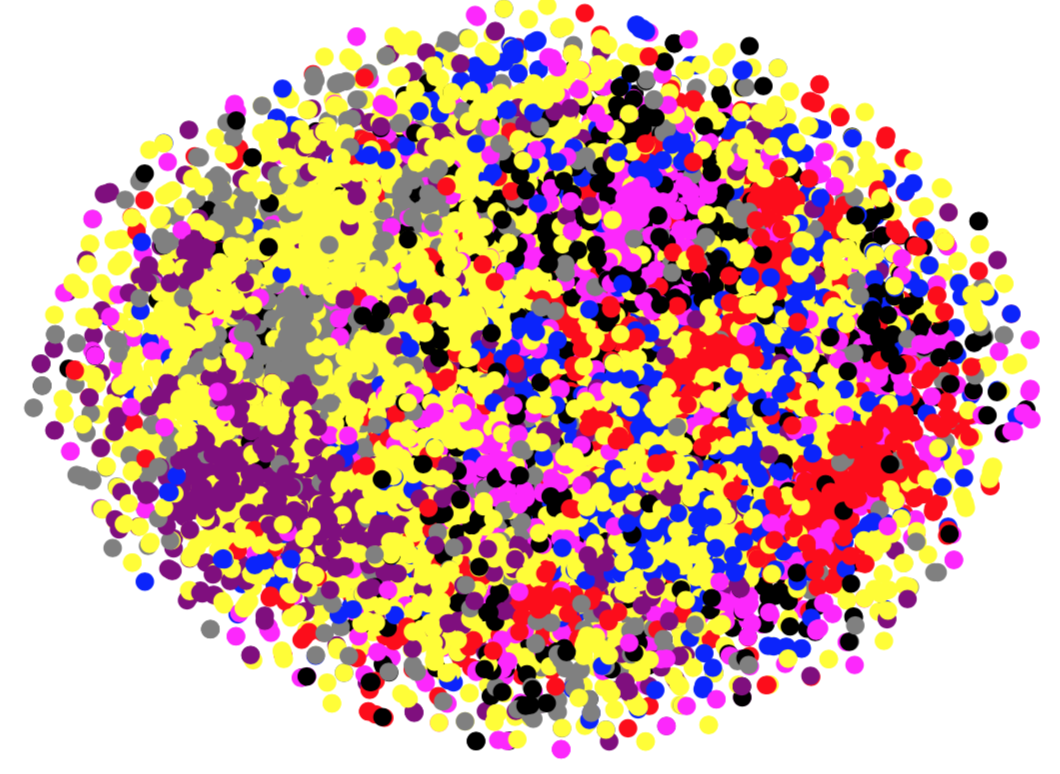}
\includegraphics[width=0.49\linewidth, height=5.5cm]{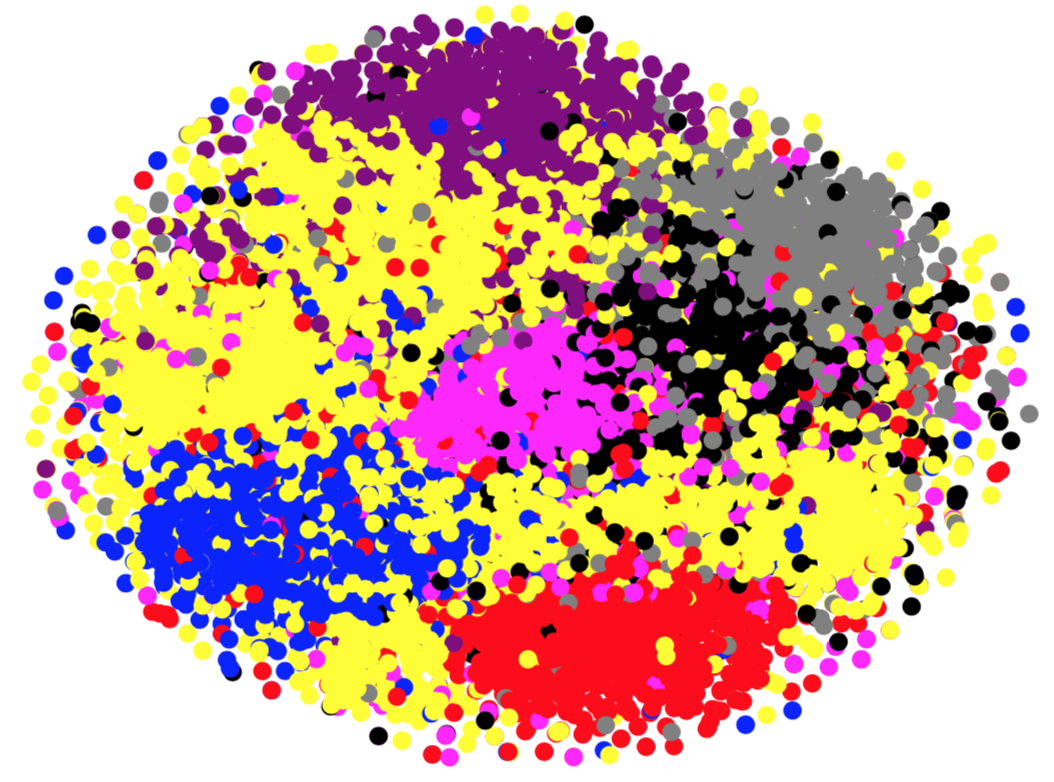}
\caption[Clustering of the representations learned by a \gls{VAE} and a \gls{SkipVAE} on \textsc{mnist}]{Clustering of the latent variables learned by fitting a \gls{VAE} (left) 
and a \gls{SkipVAE} (right) on MNIST and applying T-SNE on the test set. 
The model is a 9-layer PixelCNN and the variational neural network is a 3-layer ResNet. 
The colors represent digit labels. The \gls{SkipVAE} clusters the latent variables 
better than the \gls{VAE}; it discovers $7$ digit classes. The remaining 3 classes 
are covered by the other classes. The latent variables learned by the \gls{VAE} 
are not meaningful as they are spread out. The \gls{SkipVAE} learns more useful latent representations.}
\label{fig:cluster}
\end{figure}

\subsection{Connections \& Related Work}

\gls{DGSM} fit using \gls{AVI} are related to \glspl{VAE}. The difference between 
the approach of \citet{kingma2013auto} and the approach described above 
is the use of skip connections, when defining the model and/or the inference network. 
We verify empirically that these skip connections lead to more interpretable latent factors 
than the \gls{VAE}. To make the connection more apparent, we call \gls{AVI} in 
the context of \glspl{DGSM}, \glspl{SkipVAE}. Figure\nobreakspace \ref {fig:model} highlights the 
methodological differences between \glspl{VAE} and \glspl{SkipVAE} and shows the benefit 
of \glspl{SkipVAE} over \glspl{VAE} in learning more interpretable latent variables. 
This figure is a visualization of two interpretability metrics as a function of the 
depth of the neural network used to define the generative model for data. 
The first interpretability metric is the mutual information between the data and 
the latent variables (MI) whereas the second metric is the \gls{KL} between 
the variational distribution and the prior. The higher these metrics the better; 
higher MI and KL signal stronger correlation between the data and the latent variables.  

There have been many proposals for learning more interpretable latent variables 
with the \gls{VAE}. These approaches tackle the latent variable collapse discussed 
in Chapter\nobreakspace \ref {chap:foundations} from different angles. One approach is to handicap 
the training of the generative model \citep{bowman2015generating} or weaken 
its capacity \citep{gulrajani2016pixelvae}, effectively encouraging better 
representations by limiting the generative model. Another approach 
replaces the simple spherical Gaussian prior with more sophisticated priors. 
For example \cite{van2017neural} and \cite{ tomczak2017vae} propose 
parametric priors, which are learned along with the generative 
model. 
Still another approach uses richer variational 
distributions \citep{rezende2015variational}. In another thread of 
research, \cite{makhzani2015adversarial, mescheder2017adversarial} 
replace the $\kl$ regularization term in the \gls{VAE} objective with adversarial 
regularizers and \cite{higgins2016beta} dampen the effect of the $\kl$ regularization 
term with Lagrange multipliers. Finally, one can appeal to new inference algorithms. 
For example \cite{hoffman2017learning} uses \gls{MCMC} instead of 
variational inference and \cite{kim2018semi} uses stochastic variational 
inference, initialized with the variational neural network parameters, to iteratively 
refine the variational distribution. A very recent approach against posterior collapse relies on ideas from directional statistics. 
More specifically it consists in using the Von Mises-Fisher distribution for both the prior and 
the variational posterior and fixing the dispersion parameter of the Von Mises-Fisher distribution 
to make the $\kl$ term in the \gls{ELBO} constant~\citep{guu2017generating, xu2018spherical}. 
However this practice might result in less expressive approximate posteriors. 

\subsection{Empirical Study}

We first set some definitions
\begin{defn}\label{def:varjoint}
  For any data $\bx$ and variational posterior
  $q_{\phi}(\bz \vert \bx)$, the variational joint $q_{\phi}(\bx , \bz)$ is the joint
  distribution of $\bx$ and $\bz$ induced by
  $q_{\phi}(\bz \g \bx)$. It induces a marginal $q_{\phi}(\bz)$ called 
  the aggregated posterior~\citep{makhzani2015adversarial, mescheder2017adversarial}
\begin{align*}
q_{\phi}(\bx , \bz) &= p(\bx)\cdot q_{\phi}(\bz \vert \bx)
\quad \text{and} \quad
q_{\phi}(\bz) = E_{p(\bx)}q_{\phi}(\bz \vert \bx).
\end{align*}
The mutual information $\mathcal{I}_q(\bx, \bz)$ induced by the variational joint is 
\begin{align*}
  \mathcal{I}_q(\bx, \bz) &=
      E_{p(\bx)}E_{q_{\phi}(\bz \g \bx)}\log q_{\phi}(\bz \g \bx) - 
      E_{q_{\phi}(\bz)} \log q_{\phi}(\bz) 
      .
\end{align*}
\end{defn}

We next assess the performance of \glspl{SkipVAE} on learning latent representations 
of data by applying it to both a standard \gls{VAE}~\citep{kingma2013auto,rezende2014stochastic} 
and to the recently introduced \gls{SAVAE}~\citep{kim2018semi}. We use 
standard benchmark datasets for images and text: MNIST, Omniglot, and 
the Yahoo corpus. Text datasets have been shown to be particularly sensitive 
to latent variable collapse \citep{bowman2015generating}. 

The prior for all experiments is a spherical Gaussian, and the variational posterior is a diagonal Gaussian. 
Experiments compare \gls{SkipVAE} to baselines when 
varying the dimensionality of the latent variable and the complexity of the generative model.

\begin{table*}[t]
\center
\begin{small}
\caption[Comparing the performance of a \gls{VAE} and a \gls{SkipVAE} as the dimensionality of the latent space increases]{Performance of \gls{SkipVAE} vs \gls{VAE} on MNIST as the dimensionality 
of the latent variable increases. \gls{SkipVAE} outperforms \gls{VAE} on all collapse 
metrics while achieving a similar log likelihood---as measured by \gls{ELBO}.}
\begin{tabular}{c  c c c c c c c c c c c}
\toprule
  &  \multicolumn{2}{c}{\gls{ELBO}} & & \multicolumn{2}{c}{ $\kl$ } & & \multicolumn{2}{c}{MI} & & \multicolumn{2}{c}{AU}  \\
Dim &  \gls{VAE} & \gls{SkipVAE} & & \gls{VAE} & \gls{SkipVAE} & & \gls{VAE} & \gls{SkipVAE} & & \gls{VAE} & \gls{SkipVAE}\\
\midrule
$2$ & $\textbf{-84.27}$ & $-84.30$ & &$3.13$ & $\textbf{3.54}$  &  & $3.09$ & $\textbf{3.46}$ & & $2$ & $2$\\
$10$ & $-83.01$ &  $\textbf{-82.87}$ & & $8.29$ & $\textbf{9.41}$ & & $7.35$ & $\textbf{7.81}$ & & $9$ &  $\textbf{10}$\\
$20$ & $-83.06$ &  $\textbf{-82.55}$ & &$7.14$ & $\textbf{9.33}$    &  & $6.55$ & $\textbf{7.80}$  & & $8$ &  $\textbf{13}$\\
$50$ & $-83.31$ & $\textbf{-82.58}$ & &$6.22$ & $\textbf{8.67}$  &  & $5.81$ & $\textbf{7.49}$ & & $8$ & $\textbf{12}$\\
$100$ & $-83.41$ &  $\textbf{-82.52}$ & & $5.82$ & $\textbf{8.45}$ & & $5.53$ & $\textbf{7.38}$ & & $5$ &  $\textbf{9}$\\
\bottomrule
\end{tabular}\label{tab:mnist1}
\end{small}
\end{table*}

\begin{table*}[t]
\center
\begin{small}
\caption[Comparing the performance of a \gls{VAE} and a \gls{SkipVAE} as the flexibility of the decoder increases]{Performance of \gls{SkipVAE} vs \gls{VAE}  on MNIST (Top) and 
Omniglot (Bottom) as the complexity of the decoder increases. Skip-VAE 
outperforms VAE on all collapse metrics while achieving a similar log likelihood---as 
measured by \gls{ELBO}. In particular, this advantage widens as the number of 
layers increases. The number of latent dimensions is $20$---they are all active under \gls{SkipVAE} on Omniglot.}
\begin{tabular}{c c c c c c c c c c c c}
\toprule
  &  \multicolumn{2}{c}{\gls{ELBO}} & & \multicolumn{2}{c}{ $\kl$ } & & \multicolumn{2}{c}{MI} & & \multicolumn{2}{c}{AU}  \\
Layers & \gls{VAE} & \gls{SkipVAE} & & \gls{VAE} & \gls{SkipVAE} & & \gls{VAE} & \gls{SkipVAE} & & \gls{VAE} & \gls{SkipVAE}\\
\midrule
$1$ & $-89.64$ & \textbf{-89.22} & &$13.31$ & $\textbf{13.40}$  &  & $8.56$ & $8.56$ & & $20$ & $20$\\
$3$ & $-84.38$ &  $\textbf{-84.03}$ & & $10.12$ & $\textbf{10.71}$ & & $7.95$ & $\textbf{8.20}$ & & $16$ &  $16$\\
$6$ & $-83.19$ &  $\textbf{-82.81}$ & &$8.82$ & $\textbf{9.77}$    &  & $7.53$ & $\textbf{7.93}$  & & $11$ &  $\textbf{13}$\\
$9$ & $-83.06$ & $\textbf{-82.55}$ & & $7.14$ &  $\textbf{9.34}$   & & $6.55$ & $\textbf{7.80}$ & & $8$ &  $\textbf{13}$\\
\midrule
$1$ & $-97.69$ & $\textbf{-97.66}$  & & $\textbf{8.42}$ & $8.37$  &  & $\textbf{7.09}$ & $7.08$  & &$20$ & $20$\\
$3$ & $-93.95$ & $\textbf{-93.75}$ & &$6.43$  & $\textbf{6.58}$ &  & $5.88$ & $\textbf{5.97}$ & &$20$ &   $20$\\
$6$ & $-93.23$ & $\textbf{-92.94}$  & &$5.24$ & $\textbf{5.78}$ &  & $4.94$ & $\textbf{5.43}$ & &$20$ &  $20$\\
$9$ &$-92.79$ & $\textbf{-92.61}$ & & $4.41$ & $\textbf{6.12}$ & & $4.24$ & $\textbf{5.65}$  & & $11$ &  $\textbf{20}$\\
\bottomrule
\end{tabular}\label{tab:mnist2}
\end{small}
\end{table*}

\begin{table*}[t]
\center
\begin{small}
\caption[Comparing the performance of a \gls{VAE} and a \gls{SkipVAE} for MLP-based decoders as the flexibility of the decoder increases]{VAE and SkipVAE on MNIST using $50$ latent dimensions. The encoder is a 2-layer MLP with 512 units in each layer. The decoder is also an MLP. The results below correspond to different number of layers for the decoder.}
\begin{tabular}{c c c c c c c c c c c c}
\toprule
  &  \multicolumn{2}{c}{\acrshort{ELBO}} & & \multicolumn{2}{c}{ $\kl$ } & & \multicolumn{2}{c}{MI} & & \multicolumn{2}{c}{AU}  \\
Layers & \acrshort{VAE} & \acrshort{SkipVAE} & & \acrshort{VAE} & \acrshort{SkipVAE} & & \acrshort{VAE} & \acrshort{SkipVAE} & & \acrshort{VAE} & \acrshort{SkipVAE}\\
\midrule
$2$ & $-94.88$ &  $\textbf{-94.80}$ & & $24.23$ & $\textbf{26.35}$ & & $\textbf{9.21}$ & $9.20$ & & $17$ &  $\textbf{24}$\\
$3$ & $-95.38$ &  $\textbf{-94.17}$ & & $21.87$ & $\textbf{26.15}$ & & $9.20$ & $\textbf{9.21}$ & & $13$ &  $\textbf{21}$\\
$4$ & $-97.09$ &  $\textbf{-93.79}$ & & $20.95$ & $\textbf{25.63}$ & & $\textbf{9.21}$ & $\textbf{9.21}$ & & $11$ &  $\textbf{21}$\\
\bottomrule
\end{tabular}\label{tab:mnist3}
\end{small}
\end{table*}

\begin{table*}[t]
\center
\begin{small}
\caption[Comparing the performance of different \gls{VAE} and \gls{SkipVAE} variants on a language modeling task on the Yahoo corpus]{\gls{SkipVAE} and \textsc{skip-sa-vae} perform better than their 
counterparts (\gls{VAE},  \gls{SAVAE}) on the Yahoo corpus under all latent 
variable collapse metrics while achieving similar log-likelihoods. In particular, all 
latent dimensions are active when using \textsc{skip-sa-vae}. Perplexity (PPL) for 
the variational models  is estimated by calculating the log marginal likelihood with 
200 samples from $q(\zvec \g \bx; \phi)$. }
\begin{tabular}{l c c c c c  c  }
\toprule
Model & Dim& PPL & $\gls{ELBO}$ & $\kl$ &  MI & AU  \\
\midrule
\textsc{language model} &- & $61.60$ & - & - & - & -\\
\hline
\gls{VAE}   & $32$ & $ 62.38$ & $-330.1$ & $0.005$ & $0.002$ & $0$  \\
\gls{SkipVAE} &$32$ & $ 61.71$  & $-330.5$ & $\textbf{0.34}$ & $\textbf{0.31}$ & $\textbf{1}$  \\
 \gls{SAVAE} & $32$&$ 59.85$ & $-327.5$ & $5.47$ & $4.98$ & $14$ \\
\textsc{skip-sa-vae}  &$32$ & $ 60.87$ &$-330.3$ & $\textbf{15.05}$ & $\textbf{7.47}$ & $\textbf{32}$   \\
\gls{SAVAE}  & $64$&  $60.20$ & $-327.3$ & $3.09$ & $2.95$ & $10$ \\
\textsc{skip-sa-vae} &$64$ & $ 60.55$ & $-330.8$ & $\textbf{22.54}$ & $\textbf{9.15}$ & $\textbf{64}$  \\
\bottomrule
\end{tabular}\label{tab:yahoo}
\end{small}
\end{table*}

\parhead{Evaluation metrics} 
Our evaluation metrics assess both the performance---as given by some 
measure of log-likelihood---as well as latent variable collapse.
Performance is measured using standard metrics: for image datasets 
we report the \gls{ELBO} as a measure of log-likelihood, for text we report 
both the \gls{ELBO} and perplexity estimated using importance sampling. 

Quantitatively assessing latent variable collapse is more difficult. 
We employ three metrics KL, MI, and AU.~The first 
metric is the $\kl$ regularization term of the \gls{ELBO} as written 
in Eq.\nobreakspace \ref {eq:elboskip}. The second measure of latent variable collapse is the mutual 
information induced by the variational joint $\mathcal{I}_q(\xvec, \zvec)$. 
We follow \cite{hoffman2016elbo} and approximate this mutual 
information using Monte Carlo estimates of the two $\kl$ terms. 
In particular $\kl(q(\zvec; \phi) \, \Vert \, p(\zvec))$ is approximated as
\begin{align*}
\kl(q(\zvec; \phi) \, \Vert \, p(\zvec)) &= 
\mathbb{E}_{q(\zvec; \phi)} \left[ \log q(\zvec; \phi) - \log p(\zvec)\right] \\
&\approx 
\frac{1}{S}\sum_{s=1}^{S}
\log q(\bz^{(s)}; \phi) - \log p(\bz^{(s)})
\end{align*} 
where each aggregated posterior $q(\bz^{(s)}; \phi)$ is also approximated by Monte Carlo.

The third measure of latent variable collapse is the number of "active" dimensions of 
the latent variable $\bz$. This is defined in \cite{Burda2015} as
\begin{align*}
\text{AU} &= \sum_{d=1}^{D} \mathbbm{1}{\left\{\text{Cov}_{p(\xvec)}\left(\mathbb{E}_{q(\bz \vert \xvec; \phi)}[z_d]
\right) \geq \epsilon
\right\}},
\end{align*}
where $z_d$ is the $d^{th}$ dimension of $\bz$ and $\epsilon$ is a 
threshold. ($\mathbbm{1}\{\cdot\}$ is an indicator giving $1$ when 
its argument is true and $0$ otherwise.)
We follow \cite{Burda2015} and use a 
threshold of $\epsilon = 0.01$. We observe the same phenomenon: the 
histogram of the number of active dimensions of $\bz$ is bi-modal, which 
means that it is not highly sensitive to the chosen threshold.

\parhead{Images}
We studied MNIST and Omniglot. We use a 3-layer 
ResNet \citep{He2016} (with $3\times 3$ filters and $64$ feature maps in 
each layer) as the variational neural network and a 9-layer Gated 
PixelCNN \citep{Oord2016b} (with $3 \times 3$ filters and $32$ feature maps) 
as the likelihood. For the baseline approach without skip connections we 
apply a linear map to the sample from the variational posterior (to 
project out to the image spatial resolution), concatenate this with the 
original image, and feed this to the PixelCNN. This set up 
reflects the set up of current state-of-the-art settings for modeling images
with \glspl{VAE}~\citep{gulrajani2016pixelvae,chen2016variational}. 
For the generative skip model, we apply a linear map to the sample and concatenate 
it with the output from each layer of the PixelCNN (before feeding it to the next layer). 
This results in more parameters for the \gls{SkipVAE} model but we will see shortly that 
the baseline \gls{VAE}'s performance on the collapse metrics worsens more quickly 
than the \gls{SkipVAE} as the capacity of the model increases. 

Table\nobreakspace \ref {tab:mnist1} shows the results on MNIST when varying the size of the latent dimension. 
In all scenarios, the generative skip model yields 
higher $\kl$ between the variational posterior and the prior, higher mutual 
information, and uses more latent dimensions (as measured by AU). 

Table\nobreakspace \ref {tab:mnist2} varies the generative 
model's complexity by increasing its depth. We used $20$-dimensional latent variables. 
With the \gls{VAE}, as the generative model becomes more expressive the model
becomes less reliant on $\zvec$ as evidence by the poor performance 
on the collapse metrics. The generative skip model mitigates this issue and 
performs better on all latent-variable-collapse metrics. Note 
the \gls{ELBO} is similar for both models. These results indicate that the 
family of generative skip models has a strong inductive bias to share more 
mutual information between the observation and the latent variable.
Similar results are observed when using weaker models. For example in Table\nobreakspace \ref {tab:mnist3} 
we used \glspl{MLP} for both the variational neural network and the generative model. We 
set the dimensionality of the latent variables to $50$. Even with this weaker setting 
the \gls{SkipVAE} leads to less collapse than the \gls{VAE}. 

\parhead{Quality of the learned latent variables.}
To measure the quality of the learned latent variables for both the \gls{VAE} 
and the \gls{SkipVAE} we ran two sets of analyses: one quantitative and one qualitative. 
Qualitatively we cluster the latent variables using the learned variational neural network 
and the test set. Figure\nobreakspace \ref {fig:cluster} illustrates this. It shows a clear clustering of the 
MNIST digits with the latent space learned by the generative skip model. This is not 
the case for the latent variables learned by the \gls{VAE} which are more spread out. 
Note we did not fit a \gls{VAE} and a \gls{SkipVAE} with 2-dimensional latents for 
the visualization as this is not a realistic setting in practice. Instead we fit the 
\gls{VAE} and the \gls{SkipVAE} on $50$-dimensional latents---as is usual in 
state-of-the-art image modeling with \glspl{VAE}---and used t-SNE to project 
the learned latents on a two-dimensional space. 

Quantitatively we performed a classification experiment on MNIST 
using the latent variables learned by the variational neural networks of 
\gls{VAE} and \gls{SkipVAE} as features. This experiment uses $50$ latent dimensions, 
a $9$-layer PixelCNN as the generative model, a $3$-layer ResNet as the variational neural 
network, and a simple $2$-layer \gls{MLP} over the posterior means as the classifier. 
The \gls{MLP} has $1024$ hidden units, ReLU activations, and a dropout rate of $0.5$. 
The classification accuracy of the \gls{VAE} is $97.19\%$ which is 
lower than the accuracy of the \gls{SkipVAE} which is $98.10\%$. 
We also studied this classification performance on a weaker model. 
We replaced the $9$-layer PixelCNN and the $3$-layer ResNet above by two \glspl{MLP}.
The \gls{VAE} achieved an accuracy of $97.70\%$  whereas the \gls{SkipVAE} 
achieved an accuracy of $98.25\%$. 

\parhead{Text}
Next we analyze the Yahoo Answers dataset from \cite{Yang2017}, a 
common benchmark for deep generative models of text. Successfully training \glspl{VAE} for 
text with flexible autoregressive likelihoods such as LSTMs remains an important 
open issue in the field. In many cases the generative model learns to ignore
the latent variable (setting $\kl(q(\zvec \, | \, \xvec; \phi) \, \Vert \, p(\zvec) )$ close to zero) 
and collapses to a deterministic language model \cite{bowman2015generating}.

We use the same training setup as the current best model from \cite{kim2018semi}. 
Concretely, the variational neural network is a $1$-layer LSTM with $1024$ hidden 
units, whose last hidden state is used to predict the mean vector and the
(log) variance vector of the variational posterior. The generative model is also a $1$-layer LSTM 
with $1024$ hidden units. In the non-skip generative model the sample from the variational 
posterior is used to predict the initial hidden state of the decoder and also 
fed as input at each time step. In the generative skip model we also concatenate the 
sample with the decoder's hidden state before projecting out to the vocabulary 
space. In both cases we apply an additional softmax layer to approximate 
the predictive distribution over the next word. 

In addition to the vanilla \gls{VAE}, we also study the \gls{SAVAE} \citep{kim2018semi}, 
which proposes a different optimization-based strategy for targeting
the latent variable collapse issue when training \glspl{VAE} for text. \gls{SAVAE} combines 
stochastic variational inference \citep{Hoffman2013} with amortized variational inference 
by first using an inference network over $\xvec$ to predict the initial variational parameters 
and then subsequently running iterative inference on the \gls{ELBO} to refine 
the initial variational parameters. In our experiments we used $10$ steps of iterative 
refinement for \gls{SAVAE} and \acrshort{SkipSAVAE}.

Table\nobreakspace \ref {tab:yahoo} shows the results. Here we compare performance of adding the skip connections to both 
\gls{VAE} and SA-VAE. Table\nobreakspace \ref {tab:yahoo} shows that \gls{SkipVAE} is better than 
\gls{VAE} at avoiding latent variable collapse for similar log likelihoods.
The same conclusion holds when comparing SA-VAE and Skip-SA-VAE. 
Without any skip connections the generative model learns to ignore the latent variable 
and the mutual information is lower. Adding skip connections increases the mutual 
information. 
All in all \textsc{skip-sa-vae} outperforms all models and achieves perfect latent variable usage when 
the number of latent dimensions is either $32$ or $64$.

\subsection{Conclusion}
We proposed \acrlong{DGSM} a class of models that extend \gls{EF-PCA}. 
When fit using \gls{AVI} to scale inference, \glspl{DGSM} lead to more interpretable
 latent variables. This is verified on both image and text datasets. 
The  approach consists in using skip connections to promote a stronger dependence between 
the observations and their associated latent variables. 

One interesting line of future work is to study the constraints that should be imposed on 
the skip model to achieve a good performance---as measured by predictive log-likelihood---while 
also yielding more expressive latent representations. 
 
\section{Deep Sequential Models with Long-Range Latent Context}\label{sec:topicrnn}
One challenge in modeling sequential data is the difficulty to capture long-term dependencies. 
\gls{PGM} approaches, such as the \gls{HMM}, make Markov assumptions 
that prune these long-term dependencies. \gls{DL} approaches such as \glspl{RNN} and their 
variants (\gls{LSTM}, \gls{GRU}) have unlimited memory, in theory, but face optimization 
challenges in practice that hinder their ability to capture long-term dependencies~\citep{bengio1994learning, pascanu2013difficulty}.

In this section, we introduce a new class of model for sequential data, called \acrlong{TopicRNN}, 
that allows to capture long-range dependencies. 
\acrlong{TopicRNN} marries latent variables with neural networks. The latent variables model the 
structure shared by all the elements in a sequence whereas the neural network focuses 
on capturing local dependencies. This marriage has been shown useful for language modeling~\citep{dieng2016topicrnn}, 
conversation modeling~\citep{wen2018latent}, patient representation learning for hospital readmission~\citep{xiao2018readmission}, 
and unsupervised document representation learning~\citep{dieng2016topicrnn}. 

\begin{figure}[t]    \centering
    {\includegraphics[width=6.5cm, height=5.cm]{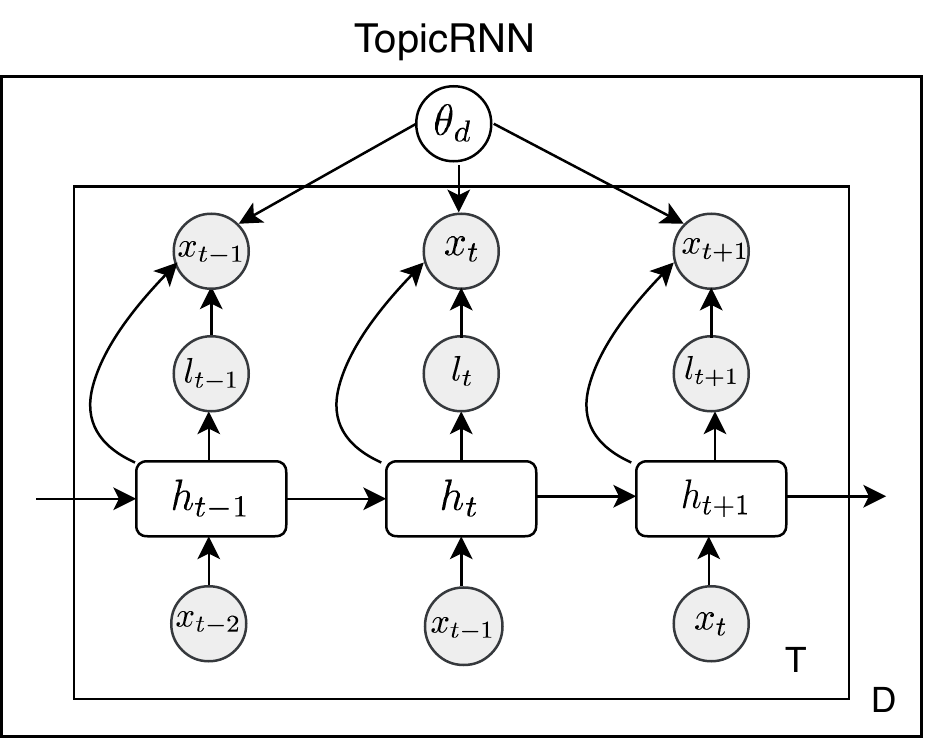} }    \caption[Graphical model of \acrlong{TopicRNN}]{Graphical model of \acrlong{TopicRNN}. There are $D$ documents. 
    Each document has $T$ words $\bx_{1:T}$. Observations are shaded in grey, 
    deterministic variables are represented in squares, and latent variables are represented 
    using unshaded circles. The unobserved $\bh_{1:T}$ represent the hidden states of the \gls{RNN}. 
    The observed variables $l_{1:T}$ correspond to stop word indicators: $l_t = 1$ if $\bx_t$ is a 
    stop word and $0$ otherwise. The latent variable $\btheta_d$ is shared by all the words in 
    document $d$. The observation model over a word $\bx_t$ is $p(\bx_t \g l_t, \btheta_d) = \text{softmax}(V^\top\bh_t + (1 - l_t) B^\top \btheta_d)$. 
    }
    \label{fig:topicrnn}\end{figure}

\subsection{Inductive Biases for Sequential Data Modeling} 
One advantage of \gls{PGM} is that it makes it easy to include inductive biases 
when specifying a model. There are many known inductive biases for discrete sequences. 
To illustrate what these inductive biases are, let's consider the following excerpt 
from the CNN news network.

``\textit{The \textcolor{blue}{U.S.}\textcolor{blue}{presidential}
  \textcolor{blue}{race} isn't only drawing attention and controversy
  in the \textcolor{blue}{United States} -- it's being closely watched
  across the globe. But what does the rest of the world think about a
  \textcolor{blue}{campaign} that has already thrown up one surprise
  after another?  CNN asked 10 journalists for their take on the
  \textcolor{blue}{race} so far, and what their  \textcolor{blue}{country} might be
hoping for in \textcolor{blue}{America}'s next
\textcolor{blue}{\bf{---}}}'' 

The missing word in this excerpt can be predicted with high accuracy by accounting for 
two types of contexts: \textit{local} context and  \textit{global} context. Local context 
is defined as the set of few words preceding the word to be predicted. Order matters 
when defining local context, it defines \emph{syntax} in language. Going back to our example 
above the phrase ``America's next" represents local context for the word we want to predict. 
It already tells us what the function of the word we want to predict is. Here we know 
we have to predict a noun. Global context defines the semantics of the word of interest. 
Order does not matter. The semantic is defined by the semantics of the words that appear in the 
same paragraph. Here such words are ``America", ``United States", ``race", ``country", ``campaign", 
and ``presidential". Accounting for this global context narrows down the search for eligible words 
to ``President" and its synonyms.  A good language model, a model of sequences of words, 
should capture at least these two important properties of natural language: syntax (local context) and semantics (global context). 

When should we account for which types of context? Local context should always be accounted for. An element 
of a sequence depends on the elements immediately preceding it. 
Global context is needed to predict certain elements in the sequence but not all elements in a sequence 
exhibit long-range dependencies. A priori we do not know which elements in a sequence do require long-term context 
and which elements do not. Ultimately, we want to discover those elements after we fit a model to data. 
In what follows, we make the simple assumption that these elements correspond to \textit{stop words} in language. 
We will justify this assumption later. 

\subsection{Model Class}

We now describe \acrlong{TopicRNN}. Consider $D$ i.i.d pairs $(\bx^{(d)}, \bl^{(d)})$ for $d = 1 \dots D$. 
Here $\bx^{(d)} = \bx_{1:T}^{(d)}$ is a document and $\bl^{(d)} = \bl_{1:T}^{(d)}$ is a vector indicating 
whether each word in document $d$ is a stop word or not. We determine this using a predefined stop word list. 
Under \acrlong{TopicRNN}, a given pair is generated as follows:
\begin{enumerate}
  \item Draw a global context vector $\theta_d \sim \mathcal{N}(0,I)$
  \item For each word in the sequence, $t = 1 \dots T$:
\begin{enumerate}
  \item Compute local context $\bh^{(d)}_t = f_{\eta}\left(\bx^{(d)}_{t-1}, \bh^{(d)}_{t-1}\right)$  
  \item Draw stop word indicator $l^{(d)}_t \sim {\rm Bernoulli}\left(\sigma\left(\Gamma^\top \bh^{(d)}_t \right)\right)$ 
  \item Draw word $\bx^{(d)}_t \sim \text{Cat}\left(\bp^{d}_t\right)$  where  $\bp^{d}_t = \text{softmax}\left(\bV^\top\bh^{(d)}_t + (1 - l^{(d)}_t) \cdot \bbeta^\top\theta_d\right)$
\end{enumerate}
\end{enumerate}
Figure\nobreakspace \ref {fig:topicrnn} shows the graphical model corresponding to this generative process. 
Here $\theta_d$ represents global context, it is shared across all the words in the document. 
We chose its prior to be a standard Gaussian. The function $f_{\eta}(\cdot)$ is a neural network 
that takes as input the previous input $\bx^{(d)}_{t-1}$ and its own previous output $\bh^{(d)}_{t-1}$. 
It can be implemented using any of the \gls{RNN} cells described in Chapter\nobreakspace \ref {chap:foundations}. 
The function $\sigma(\cdot)$ is the logistic function. The stop word indicator $l^{(d)}_t$ controls how 
the latent global context $\theta_d$ affects the output. If $l^{(d)}_t = 1$ (indicating $\bx^{(d)}_t$ is a stop word), the global context 
$\theta_d$ has no contribution to the output. Otherwise, we add a bias
to favor those words that are more likely to appear when mixing with $\theta_d$, as measured by the dot product 
between $\theta$ and the latent word vector $b_i$ for the $i$th vocabulary word. 

To understand the bias term $\bbeta^\top\theta_d$ recall \gls{LDA}, a probabilistic topic model described in Chapter\nobreakspace \ref {chap:foundations}. 
Marginalize out its per-word discrete topic assignments, the conditional distribution of the words in the document given 
the topics and topic proportions under \gls{LDA} is
\begin{align}
	p(\bx^{(d)}_{1:T} \vert \theta_d, \bbeta) 
	&= \prod_{t=1}^{T} \sum_{k=1}^{K}  \theta_{dk} \bbeta_{\bx^{(d)}_t} = \prod_{t=1}^{T} \bbeta^\top\theta_d\Big\vert_{\bx^{(d)}_t}
\end{align}
This implies $\bx^{(d)}_t \vert \theta_d, \bbeta \sim \bbeta^\top\theta_d$. Although this term has the same form as the bias term in \acrlong{TopicRNN} 
the constraints put on the matrices $\bbeta$ and $\theta_d$ are different in \gls{LDA} and in \acrlong{TopicRNN}. 
In \gls{LDA} both $\bbeta$ and $\theta_d$ are modeled using the Dirichlet distribution whereas in 
\acrlong{TopicRNN} they are modeled as a deterministic model parameter and a standard Gaussian. 
Adding a simplex constraints to $\theta_d$ enforces interpretability for $\bbeta$~\citep{donoho2004does} 
We can achieve this in \acrlong{TopicRNN} by simply mapping the global context vector $\theta_d$ to $\text{softmax}(\cdot)$. 
In our empirical study we use a standard Gaussian for simplicity and found the matrix $\bbeta$ still captures interpretable word clusters.  

\subsection{Amortized Variational Inference}

We fit \acrlong{TopicRNN} using \gls{AVI}. For that we have to specify a variational family over the global context $\theta_d$. 
Denote by $q_{\phi}\left(\theta_{1:D} \Big\vert \bx^{(1:D)}_{1:T}, \bl^{(1:D)}_{1:T}\right)$ the variational family; it is indexed by $\phi$. 
We factorize it as
\begin{align*}
	q_{\phi}\left(\theta_{1:D} \Big\vert \bx^{(1:D)}_{1:T}, \bl^{(1:D)}_{1:T}\right) 
	&= q_{\phi}\left(\theta_d \Big\vert \bx^{(d)}_{1:T}, \bl^{(d)}_{1:T}\right)
\end{align*}
Each factor $q_{\phi}\left(\theta_d \Big\vert \bx^{(d)}_{1:T}, \bl^{(d)}_{1:T}\right)$ is a Gaussian whose mean and covariance 
are given by the outputs of neural networks,
\begin{align*}
	q_{\phi}\left(\theta_d \Big\vert \bx^{(d)}_{1:T}, \bl^{(d)}_{1:T}\right)
	&= \mathcal{N}\left(
		\bmu^{(d)}\left(\bx^{(d)}_{1:T}, \bl^{(d)}_{1:T}; \phi_{\mu}\right), \bSigma^{(d)}\left(\bx^{(d)}_{1:T}, \bl^{(d)}_{1:T}; \phi_{\Sigma}\right)
	\right)
\end{align*}
Here $\phi = \left(\phi_{\mu}, \phi_{\Sigma}\right)$. 
In practice we use one shared neural network whose output we map to the mean and the covariance using 
two different sets of weight matrices. The input of this shared neural network is the bag-of-word representation 
of the document $\bx^{(d)}_{1:T}$ multiplied by $1 - \bl^{(d)}_{1:T}$. This multiplication zeroes out 
the contributions of the stop words in the inference of $\theta_d$. 

We can now form the \gls{ELBO}, 
\begin{align*}
	\gls{ELBO} 
	&= \sum_{d=1}^{D}\mathbb{E}_{q_{\phi}\left(\theta_d \Big\vert \bx^{(d)}_{1:T}, \bl^{(d)}_{1:T}\right)}\left\{
		\log p\left(\bx^{(d)}_t \vert  \bl^{(d)}_t, \theta_d \right) 
	\right\} - \gls{KL}\left( 
		q_{\phi}\left(\theta_d \Big\vert \bx^{(d)}_{1:T}, \bl^{(d)}_{1:T}\right) \Big\vert\Big\vert p(\theta_d)
	 \right).
\end{align*}
The \gls{ELBO} is intractable. We approximate it using Monte Carlo with the 
reparameterization trick~\citep{kingma2013auto,rezende2014stochastic}. 
Importantly, we apply truncated backpropagation through time, which 
unrolls the \gls{RNN} $f_{\eta}(\cdot)$ a fixed number of time steps 
(instead of accounting for all the elements in the sequence of words 
when computing the states of the \gls{RNN}, which would cause optimization issues~\citep{bengio1994learning, pascanu2013difficulty}.)

\subsection{Application to Language Modeling}

We first tested \acrlong{TopicRNN} on the word prediction 
task using the Penn Treebank (PTB) 
portion of the Wall Street Journal. We use the 
standard split, where sections 0-20 (930K tokens) 
are used for training, sections 21-22 (74K tokens) for 
validation, and 
sections 23-24 (82K tokens) for 
testing~\citep{mikolov2010recurrent}. 
We use a vocabulary of size $10K$ that includes 
the special token \textit{unk} for 
rare words and \textit{eos} that indicates the 
end of a sentence. \acrlong{TopicRNN} takes documents 
as inputs. We split the PTB data into blocks 
of 10 sentences to constitute documents 
as done by~\citep{mikolov2012context}. 
The inference network takes as input the bag-of-words 
representation of the input document. 
For that reason, the vocabulary size of the inference network is reduced 
to $9551$ after excluding $449$ pre-defined stop words. 

In order to compare with previous work on 
contextual \acrshort{RNN}s (e.g.~\citet{mikolov2012context}), we trained \acrlong{TopicRNN}
 using different network sizes. We performed 
 word prediction using a recurrent neural 
 network with 10 neurons, 100 neurons and 300 
 neurons. For these experiments, we used 
 a multilayer perceptron with 2 hidden layers 
 and 200 hidden units per layer for the inference network. 
 The dimensionality $K$ of $\theta_d$ was tuned depending on 
 the size of the \acrshort{RNN}. For 10 neurons we used $K=18$. For $100$ and $300$ neurons we chose $K=50$. 
 We used the validation set to tune the hyperparameters of the model (including $K$).
 We used a maximum of $15$ epochs for the experiments and performed early stopping 
 using the validation set. For comparison purposes we did not apply regularization and used 1 layer 
 for the \acrshort{RNN} and its counterparts in all the experiments. 
\begin{table*}[t]
\center
\begin{small}
\caption[Performance of \acrlong{TopicRNN} and \gls{RNN} on a next word prediction task on the Penn Treebank dataset]{\acrlong{TopicRNN} and its counterparts exhibit lower perplexity scores across different network sizes. 
These results prove TopicRNN has more generalization capabilities:
    for example we only need a TopicGRU with 100 neurons 
    to achieve a better perplexity than stacking 2 LSTMs with 
    200 neurons each: 112.4 vs 115.9)}
\begin{tabular}{ccccccc}
\toprule
  &  \multicolumn{2}{c}{$10$ Neurons} & \multicolumn{2}{c}{ $100$ Neurons } & \multicolumn{2}{c}{300 Neurons}  \\
  \midrule
Method & Val & Test & Val & Test & Val & Test  \\
\midrule
\acrshort{RNN} & $239.2$ & $225.0$ & $150.1$ & $142.1$ & $-$ & $124.7$\\
\acrshort{RNN} + \acrshort{LDA} & $197.3$ &$187.4$ & $132.3$& $126.4$& $-$ & $113.7$\\
\hline
\acrlong{TopicRNN} & $184.5$ & $172.2$ & $128.5$  & $122.3$  & $118.3$ & $112.2$ \\
\acrlong{TopicLSTM} & $188.0$ & $175.0$ & $126.0$  & $118.1$  & $104.1$  & $99.5$ \\
\acrlong{TopicGRU} & $178.3$ & $166.7$ & $118.3$  & $112.4$  & $99.6$ & $97.3$ \\
\bottomrule
\end{tabular}\label{tab:pp}
\end{small}
\end{table*}
 
Table\nobreakspace \ref {tab:pp} reports perplexity on the validation set and the test set for different network sizes. 
Perplexity can be thought of as a measure of surprise for a language model. It is 
defined as the exponential of the average negative log likelihood. 
We learn three things from Table\nobreakspace \ref {tab:pp}. First, the perplexity is reduced the larger the 
network size. Second, \acrshort{RNN}s with global context features perform better than \acrshort{RNN}s 
without context features. Third, we see that \acrlong{TopicRNN} gives better perplexity than the previous baseline result 
reported by~\citet{mikolov2012context}. Note we compute the perplexity scores for word prediction 
using a sliding window, to compute $\theta$ as we move along the sequences. The topic vector $\theta$ that is used 
from the current batch of words is estimated from the previous batch of words. This enables fair comparison 
to previously reported results \citep{mikolov2012context}.

\subsection{Unsupervised Feature Learning and Application to Sentiment Analysis}

We performed sentiment analysis using \acrlong{TopicRNN} 
as a feature extractor on 
the IMDB 100K dataset. This data consists 
of 100,000 movie reviews from 
the Internet Movie Database (IMDB) website. The 
data is split into $75\%$ for training and $25\%$ for testing. 
Among the 75K training reviews, 50K are unlabelled 
and 25K are labelled as carrying either a positive or a negative sentiment. 
All 25K test reviews are labelled. We trained 
\acrlong{TopicRNN} on 65K random training reviews and used 
the remaining 10K reviews for validation. To learn a classifier, we passed 
the 25K labelled training reviews through the learned \acrlong{TopicRNN} model. We 
then concatenated the output of the inference network and the last 
state of the \acrshort{RNN} for each of these 25K reviews to compute the feature vectors. 
We then used these feature vectors to train a neural network with one hidden 
layer, 50 hidden units, and a sigmoid activation function to predict sentiment, exactly as 
 done in~\citet{le2014distributed}.
 
\begin{figure}[t]
\centering
\includegraphics[width=.95\textwidth]{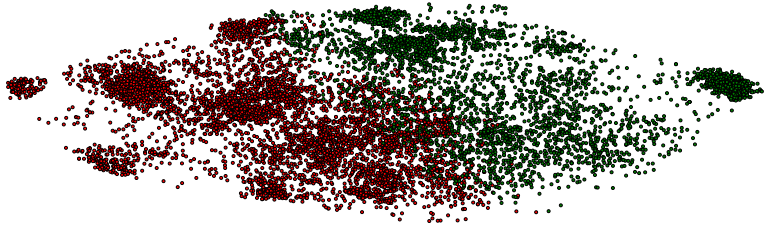}
\caption[Clustering of the representations learned by \acrlong{TopicRNN} on the \textsc{imdb} movie review dataset]{Clusters of a sample of $10000$ movie reviews from the IMDB $100$K 
dataset using \acrlong{TopicRNN} as feature extractor. We used K-Means to cluster
 the feature vectors. We then used PCA to reduce 
 the dimension to two for visualization purposes.
 \red{red} is a negative review and \green{green} is a positive review.}
\label{fig:clusters}
\end{figure}

To train the \acrlong{TopicRNN} model, we used a vocabulary 
of size 5,000 and mapped all other words to the \textit{unk} token. 
We took out 439 stop 
words to create the input of the inference network. 
We used 500 units and 2 layers 
for the inference network, and used 2 layers 
and 300 units per-layer for the \acrshort{RNN}. 
We chose a step size of 5 and defined 200 topics. 
We did not use any regularization such as dropout. 
We trained the model for 13 epochs and used the 
validation set to tune the 
hyperparameters of the model and track perplexity 
for early stopping. This experiment took close to 78 hours
on a MacBook pro quad-core with 16GHz of RAM. 

Table\nobreakspace \ref {tab:imdb} summarizes sentiment classification results from
\acrlong{TopicRNN} and other methods.  Our error rate is
$6.28\%$.\footnote{The experiments were solely based on TopicRNN.
Experiments using TopicGRU/TopicLSTM are being carried out and will be
added as an extended version of this paper.}  This is close to the
state-of-the-art $5.91\%$~\citep{miyato2016adversarial} despite that
we do not use the labels and adversarial training in the feature
extraction stage.  Our approach is most similar
to~\citet{le2014distributed}, where the features were extracted in a
unsupervised way and then a one-layer neural net was trained for
classification.

\begin{table}[t]
\caption[Sentiment classification error rate on the \textsc{imdb} movie review dataset]{Classification error rate on IMDB 100k dataset. 
\acrlong{TopicRNN} achieves state of the art error rate amongst methods that first perform 
unsupervised feature extraction before doing sentiment classification.} 
\label{tab:imdb}
\centering
\resizebox{0.8\columnwidth}{!}{\begin{tabular}{lr}
\toprule
Model & Reported Error rate\\\midrule
BoW (bnc) (Maas et al., 2011) & $12.20\%$\\
BoW ($b\Delta$ t\'c) (Maas et al., 2011) & $11.77\%$\\
LDA (Maas et al., 2011) & $32.58\%$\\
Full +  BoW (Maas et al., 2011) & $11.67\%$ \\
Full + Unlabelled + BoW (Maas et al., 2011) & $11.11\%$ \\
WRRBM  (Dahl et al., 2012) & $12.58\%$ \\
WRRBM + BoW (bnc) (Dahl et al., 2012) & $10.77\%$ \\
MNB-uni (Wang \& Manning, 2012) & $16.45\%$ \\
MNB-bi (Wang \& Manning, 2012) & $13.41\%$ \\
SVM-uni (Wang \& Manning, 2012) & $13.05\%$ \\
SVM-bi (Wang \& Manning, 2012) & $10.84\%$ \\
NBSVM-uni (Wang \& Manning, 2012) & $11.71\%$ \\
seq2-bown-CNN (Johnson \& Zhang, 2014) & $14.70\%$ \\
NBSVM-bi (Wang \& Manning, 2012) & $8.78\%$ \\
Paragraph Vector (Le \& Mikolov, 2014) & $7.42\%$ \\
SA-LSTM  with joint training (Dai \& Le, 2015) & $14.70\%$ \\
LSTM with tuning and dropout (Dai \& Le, 2015)  & $13.50\%$ \\
LSTM initialized with word2vec embeddings (Dai \& Le, 2015) & $10.00\%$ \\
SA-LSTM  with linear gain (Dai \& Le, 2015) & $9.17\%$ \\
LM-TM  (Dai \& Le, 2015) & $7.64\%$ \\
SA-LSTM (Dai \& Le, 2015)  & $7.24\%$ \\
\textbf{Virtual Adversarial (Miyato et al. 2016)} & \textbf{5.91\%}\\
\midrule
\textbf{TopicRNN } & \textbf{6.28\%} \\\bottomrule
\end{tabular}}
\end{table}

Figure\nobreakspace \ref {fig:clusters} shows the ability of \acrlong{TopicRNN} to cluster documents 
using the feature vectors as created during the sentiment analysis task. Reviews with positive sentiment are colored 
in green while reviews carrying negative sentiment are shown in red. 

\subsection{Conclusion} 
We introduced \acrlong{TopicRNN}, a class of model for sequential data 
that marries latent variables and neural networks. The latent variables 
model the structure shared between all the elements of a sequence 
whereas the neural network models local dependencies. 
\acrlong{TopicRNN} yields competitive per-word perplexity on the Penn Treebank benchmark dataset. 
It is effective at learning unsupervised document features for sentiment classification on the \acrshort{IMDB} 
benchmark dataset. \acrlong{TopicRNN} has also been applied to healthcare data 
where learning meaningful patient representations can help predict hospital readmission~\citep{xiao2018readmission}. 
Finally, it has been used for conversation modeling by~\citet{wen2018latent}. 
Future work can study the performance of \acrlong{TopicRNN} when words that do not need global context
are learned instead of chosen to be stop words of language. 
 
\section{Topic Modeling in Embedding Spaces}\label{sec:etm}

Topic models are statistical tools for discovering the hidden semantic
structure in a collection of documents
\citep{blei2003latent,blei2012probabilistic}. Topic models and their
extensions have been applied to many fields, such as marketing,
sociology, political science, and the digital humanities.
\citet{boydgraber2017applications} provide a review. 

Most topic models build on \gls{LDA} \citep{blei2003latent}, which we described in Chapter\nobreakspace \ref {chap:foundations}. 
\gls{LDA} is a powerful model and it is widely used.  However, it suffers from a pervasive technical 
problem---it fails in the face of large vocabularies. Practitioners must severely prune their
vocabularies in order to fit good topic models, i.e., those that are both predictive and interpretable.
This is typically done by removing the most and least frequent words (called \textit{stop words} and \textit{rare words} respectively.)
On large collections, this pruning may remove important terms and limit the scope of the models. 
The problem of topic modeling with large vocabularies has yet to be addressed in the research literature. 

In this section we describe how we leverage word embeddings, a successful advance of \gls{DL}, to 
solve the problems described above. We develop the \gls{ETM}~\citep{dieng2019topic}. The \gls{ETM} marries \gls{LDA} 
and word embeddings to enable flexible topic modeling at large scale. The flexibility provided 
by the word embeddings will allow us to not have to prune stop words and rare words to learn interpretable topics. 
Furthermore, we devise an efficient \gls{AVI} procedure to fit the \gls{ETM}. The recognition network 
used for \gls{AVI} will allow us to perform evaluation on new documents without running 
an optimization loop, as required in \gls{LDA}. Other benefits brought in by the use of the word embeddings 
in the context of topic modeling is that novel unseen words can be assigned to a given topic. 

\begin{figure}[t]
	\centering
	\includegraphics[width=0.5\textwidth]{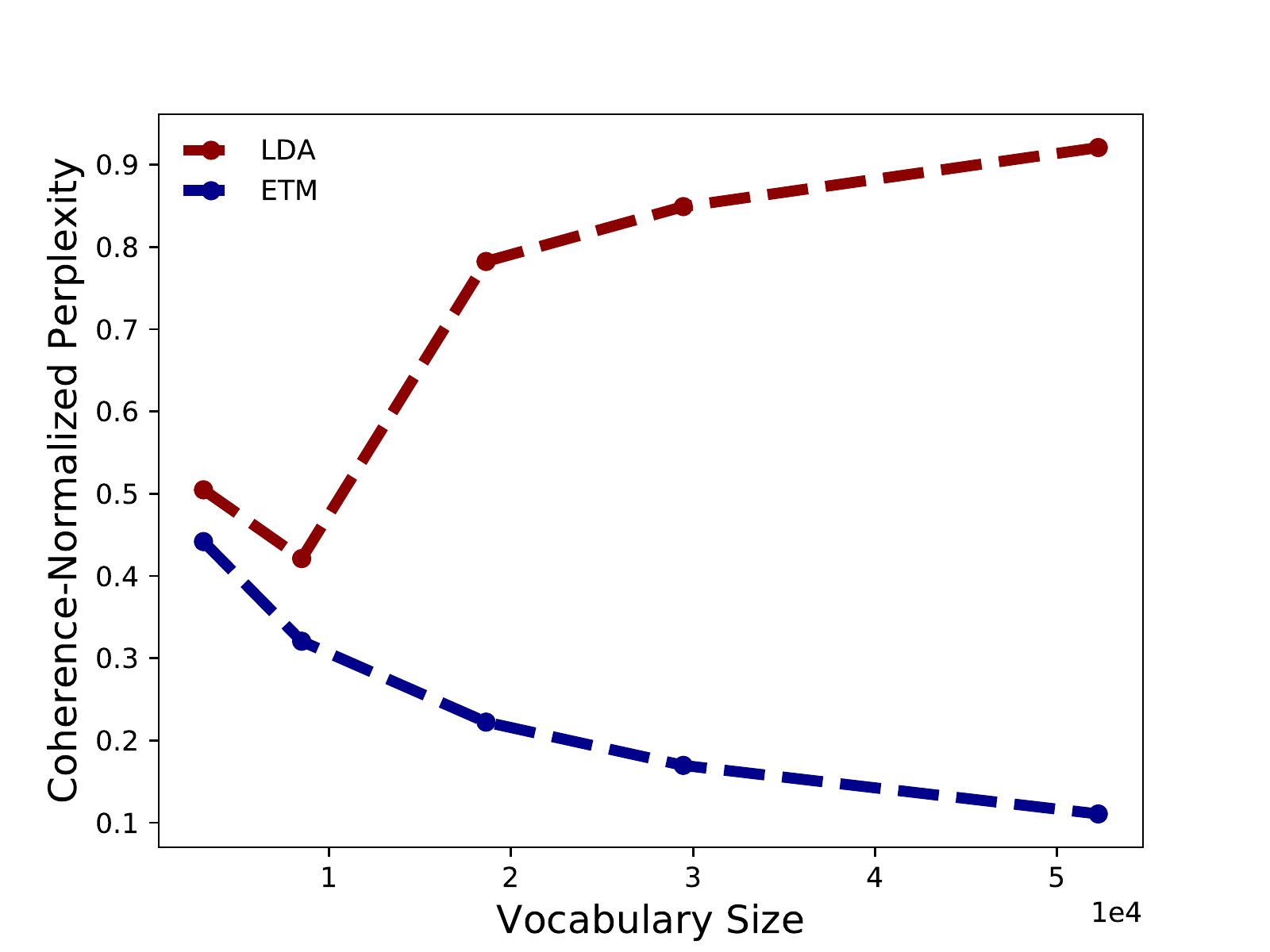}
	\caption[Comparing the \acrshort{ETM} and \acrshort{LDA} on the \textit{20NewsGroup} corpus]{Ratio of the held-out perplexity on a document completion task
	and the topic coherence as a function of the vocabulary size for the \acrshort{ETM} and \acrshort{LDA} on the \textit{20NewsGroup}  corpus. 
	The perplexity is normalized by the size of the vocabulary. 
	While the performance of \gls{LDA} deteriorates for large vocabularies, the \gls{ETM} maintains good performance.
	\label{fig:log_lik_intro}}
\end{figure}

Figure\nobreakspace \ref {fig:log_lik_intro} illustrates one of the advantages of the \gls{ETM} over \gls{LDA}. This figure shows the
ratio between the perplexity on held-out documents (a measure of predictive performance) 
and the topic coherence (a measure of the quality of the topics), as a function of the size of the vocabulary.
(The perplexity has been normalized by the vocabulary size.)
This is for a corpus of $11.2$K articles from the \textit{20NewsGroup} and for $100$ topics.  
The red line is \gls{LDA}; its performance deteriorates as the vocabulary size
increases---the predictive performance and the quality of the topics get worse. 
The blue line is the \gls{ETM}; it maintains good performance, even as the vocabulary size gets large.

\begin{figure*}[t]
\centering
\begin{minipage}{.45\textwidth}
  \centering
  \includegraphics[width=.8\linewidth]{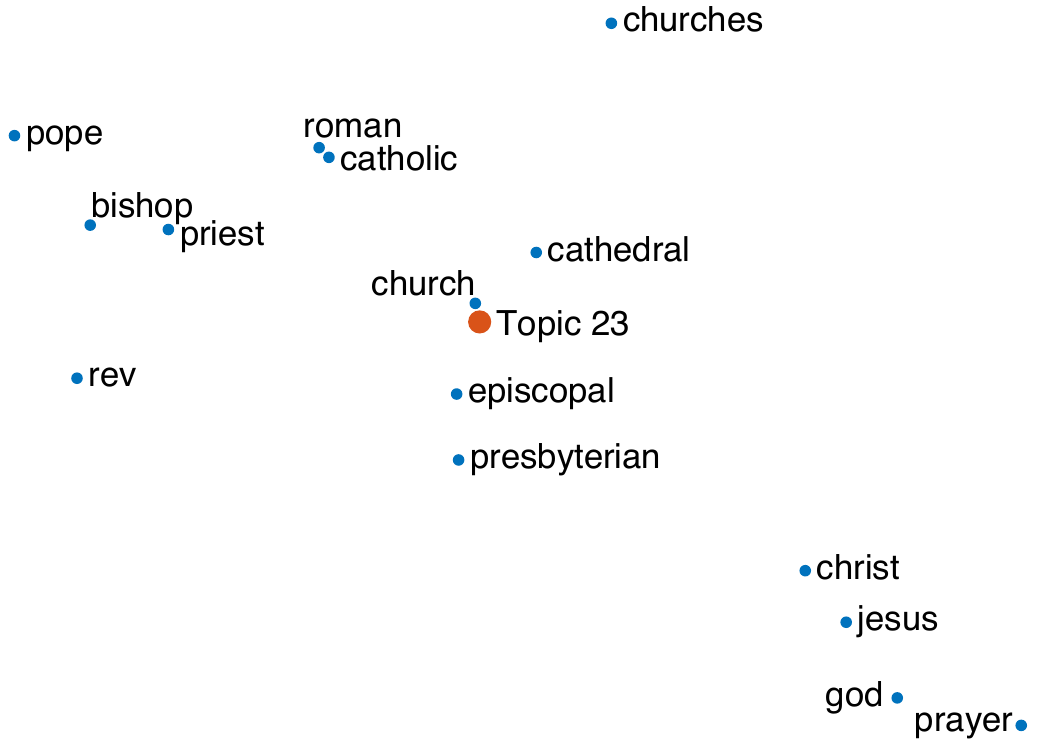}
  \captionof{figure}[A topic about Christianity found by the \acrshort{ETM} on \textit{The New York Times}]{A topic about Christianity found by the \acrshort{ETM} on \textit{The New York Times}. 
  The topic is a point in the word embedding space.}
  \label{fig:topic_embedding_intro1}
\end{minipage}\hspace*{1cm}
\begin{minipage}{.45\textwidth}
  \centering
  \includegraphics[width=0.91\linewidth]{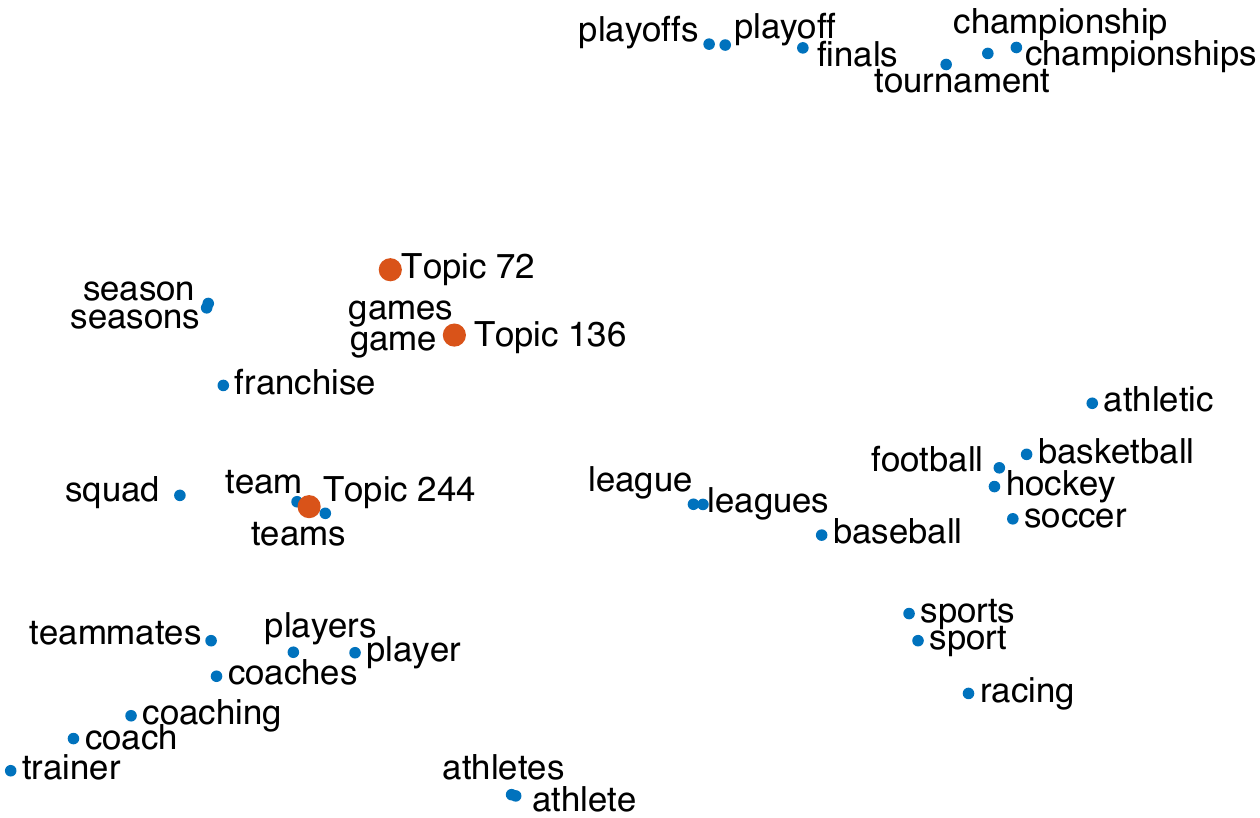}
  \captionof{figure}[Topics about sports found by the \acrshort{ETM} on \textit{The New York Times}]{Topics about sports found by the \acrshort{ETM} on \textit{The New York Times}. Each topic is a point in the word embedding space.}
  \label{fig:topic_embedding_intro2}
\end{minipage}
\end{figure*}

Figures\nobreakspace \ref {fig:topic_embedding_intro1} and\nobreakspace  \ref {fig:topic_embedding_intro2}
show topics in the embedding space of words from a $300$-topic \gls{ETM} of \textit{The New York Times}. 
These topics are about Christianity and sports.

We now describe the \gls{ETM} in detail. 

\subsection{The Embedded Topic Model}\label{sec:model}

The \gls{ETM} is a topic model that uses embedding representations of
both words and topics.  It contains two notions of latent
dimension. First, it embeds the vocabulary in an $L$-dimensional
space.  These embeddings are similar in spirit to classical word
embeddings.  Second, it represents each document in terms of $K$
latent topics.

In traditional topic modeling, each topic is a full distribution over
the vocabulary. In the \gls{ETM}, however, the $k^{\textrm{th}}$ topic is a vector
$\alpha_k\in\mathbb{R}^L$ in the embedding space. We call $\alpha_k$ a
\textit{topic embedding}---it is a distributed representation of the
$k^{\textrm{th}}$ topic in the semantic space of words.

In its generative process, the \gls{ETM} uses the topic embedding to
form a per-topic distribution over the vocabulary. Specifically, the
\gls{ETM} uses a log-linear model that takes the inner product of the
word embedding matrix and the topic embedding.  With this form, the
\gls{ETM} assigns high probability to a word $v$ in topic $k$ by
measuring the agreement between the word's embedding and the topic's
embedding.

Denote the $L \times V$ word embedding matrix by $\rho$; the column
$\rho_v$ is the embedding of term $v$.  Under the \gls{ETM}, the generative
process of the $d^{\textrm{th}}$ document is the following:
\begin{compactitem}
\item[1.] Draw topic proportions $\theta_d \sim \mathcal{LN}(0,I).$
\item[2.] For each word $n$ in the document:
  \begin{compactitem}
    \setlength{\itemindent}{-0.3cm}
  \item[a.] Draw topic assignment $z_{dn} \sim \text{Cat}(\theta_d).$
  \item[b.] Draw the word $w_{dn} \sim \text{softmax}(\rho^\top\alpha_{z_{dn}})$.
  \end{compactitem}
\end{compactitem}
In Step 1, $\mathcal{LN}(\cdot)$ denotes the logistic-normal
distribution~\citep{Aitchison:1980,blei2007correlated}; it transforms
a standard Gaussian random variable to the simplex. A draw $\theta_d$
from this distribution is obtained as
\begin{align}
  \label{eq:logistic-normal}
  \delta_d \sim \Ncal\left(0,I\right); \quad  \theta_d =
  \text{softmax}(\delta_d).
\end{align}
(We replaced the Dirichlet with the logistic normal to easily use
reparameterization in the inference algorithm; see
Section\nobreakspace \ref {sec:inference}.)

Steps 1 and 2a are standard for topic modeling: they represent
documents as distributions over topics and draw a topic assignment for
each observed word. Step 2b is different; it uses the embeddings of
the vocabulary $\rho$ and the assigned topic embedding
$\alpha_{z_{dn}}$ to draw the observed word from the assigned topic, as
given by $z_{dn}$.

The topic distribution in Step 2b mirrors the \gls{CBOW} likelihood in
Eq.\nobreakspace \ref {eq:cbow}. Recall \gls{CBOW} uses the surrounding words to form
the context vector $\alpha_{dn}$.  In contrast, the \gls{ETM} uses the
topic embedding $\alpha_{z_{dn}}$ as the context vector, where the
assigned topic $z_{dn}$ is drawn from the per-document variable
$\theta_d$.  The \gls{ETM} draws its words from a document context,
rather than from a window of surrounding words.

The \gls{ETM} likelihood uses a matrix of word embeddings $\rho$, a
representation of the vocabulary in a lower dimensional space.  In
practice, it can either rely on previously fitted embeddings or learn
them as part of its overall fitting procedure.  When the \gls{ETM}
learns the embeddings as part of the fitting procedure, it
simultaneously finds topics and an embedding space.

When the \gls{ETM} uses previously fitted embeddings, it learns the
topics of a corpus in a particular embedding space.  This strategy is
particularly
useful when there are words in the embedding that are not used in the
corpus.  The \gls{ETM} can hypothesize how those words fit in to the
topics because it can calculate $\rho_v^\top \alpha_k$, even for words
$v$ that do not appear in the corpus.

\subsection{Inference and Estimation}\label{sec:inference}

We are given a corpus of documents $\{\bw_{1}, \ldots, \bw_{D}\}$,
where $\bw_d$ is a collection of $N_d$ words.  How do we fit the
\gls{ETM}?

\parhead{The marginal likelihood.} The parameters of the \gls{ETM} are
the embeddings $\rho_{1:V}$ and the topic embeddings
$\alpha_{1:K}$; each $\alpha_k$ is a point
in the embedding space.  We maximize the marginal likelihood of the
documents,
\begin{align}
  \label{eq:marginal}
  \cL(\alpha, \rho) = \sum_{d=1}^{D} \log p(\bw_d \g \alpha, \rho).
\end{align}

The problem is that the marginal likelihood of each document is
intractable to compute.  It involves a difficult integral over the
topic proportions, which we write in terms of the untransformed
proportions $\delta_d$ in Eq.\nobreakspace \ref {eq:logistic-normal},
\begin{align}
  \label{eq:integral}
  p(\bw_d \g \alpha, \rho) =
  \int p(\delta_d)
  \prod_{n=1}^{N_d}
  p(w_{dn} \g \delta_d, \alpha, \rho) \dd \delta_d.
\end{align}
The conditional distribution of each word marginalizes out the topic
assignment $z_{dn}$,
\begin{align}
  \label{eq:likelihood}
  p(w_{dn} \g \delta_d, \alpha, \rho)
  &= \sum_{k=1}^{K} \theta_{dk} \beta_{k,w_{dn}}.
\end{align}
Here, $\theta_{dk}$ denotes the (transformed) topic proportions
(Eq.\nobreakspace \ref {eq:logistic-normal}) and $\beta_{kv}$ denotes a traditional
``topic,'' i.e., a distribution over words, induced by the word embeddings
$\rho$ and the topic embedding $\alpha_k$,
\begin{align}
  \label{eq:topic}
  \beta_{kv} = \textrm{softmax}(\rho^\top \alpha_k)\big|_v.
\end{align}
Eqs.\nobreakspace  \ref {eq:integral} to\nobreakspace  \ref {eq:topic}  flesh out the likelihood in
Eq.\nobreakspace \ref {eq:marginal}.

\parhead{Variational inference.} We sidestep the intractable integral
with variational inference, which we reviewed in Chapter\nobreakspace \ref {chap:foundations}. 
Variational inference optimizes a sum of per-document bounds on the log of the
marginal likelihood of Eq.\nobreakspace \ref {eq:integral}.  There are two sets of
parameters to optimize: the model parameters, as described above, and
the variational parameters, which tighten the bounds on the marginal
likelihoods.

To begin, posit a family of distributions of the untransformed topic
proportions $q(\delta_d\prm \bw_d, \nu)$.  We use \gls{AVI}, where the variational distribution 
of $\delta_d$ depends on both the document $\bw_d$ and shared variational parameters $\nu$.  In
particular $q(\delta_d\prm \bw_d, \nu)$ is a Gaussian whose mean and
variance come from an ``inference network,'' a neural network
parameterized by $\nu$ \citep{kingma2014autoencoding}.
The inference network ingests the document $\bw_d$ and outputs a
mean and variance of $\delta_d$. (To accommodate documents of varying
length, we form the input of the inference network by normalizing the
bag-of-word representation of the document by the number of words $N_d$.)

We use this family of variational distributions to bound the log-marginal
likelihood.  The \gls{ELBO} is a function of the model
parameters and the variational parameters,
\begin{align} \label{eq:rem_elbo}
  \cL (\alpha, \rho, \nu) & =
        \sum_{d=1}^{D} \sum_{n=1}^{N_d}
        \E{q}{\log p(w_{nd} \g \delta_d, \rho, \alpha)}  - \sum_{d=1}^{D} \mathrm{KL}(q(\delta_d ; \bw_d, \nu) \; || \; p(\delta_d)).
\end{align}

The first term of the \gls{ELBO} (Eq.\nobreakspace \ref {eq:rem_elbo}) encourages variational distributions 
$q(\delta_d\prm \bw_d, \nu)$ that place mass on unnormalized topic proportions $\delta_d$ that explain the
observed words while the second term encourages $q(\delta_d\prm \bw_d, \nu)$ to be close 
to the prior $p(\delta_d)$. Maximizing the \gls{ELBO} with respect to the model parameters $(\alpha, \rho)$ 
is equivalent to maximizing the expected complete log-likelihood,
$\sum_{d} \log p(\delta_d, \bw_d \g \alpha, \rho)$.

The \gls{ELBO} in Eq.\nobreakspace \ref {eq:rem_elbo} is intractable because the expectation is intractable. 
However we can use Monte Carlo to approximate the \gls{ELBO},
\begin{align}
  \label{eq:elbo2}
  \tilde{\cL} (\alpha, \rho, \nu) & =
    \frac{1}{S} \sum_{d=1}^{D} \sum_{n=1}^{N_d}
       \sum_{s=1}^{S}{\log p(w_{nd} \g \delta_d^{(s)}, \rho, \alpha)} 
        - \sum_{d=1}^{D} \mathrm{KL}(q(\delta_d ; \bw_d, \nu) \; || \; p(\delta_d))
\end{align}
where $ \delta_d^{(s)} \sim q(\delta_d ; \bw_d, \nu) $ for $s = 1 \dots S.$
To reduce variance we use the reparameterization trick when sampling the unnormalized 
proportions  $\delta_d^{(1)}, \dots, \delta_d^{(S)}$\citep{kingma2014autoencoding,Titsias2014_doubly,rezende2014stochastic}. 
That is, we sample $\delta_d^{(s)}$ from $q(\delta_d ; \bw_d, \nu)$ as 
\begin{align}\label{eq:sample}
	\epsilon_d^{(s)} &\sim \mathcal{N}(0, I) \text{ and } \delta_d^{(s)} = \mu_d + \Sigma_d^{\frac{1}{2}} \odot \epsilon_d^{(s)}
\end{align}
where $\mu_d$ and $\Sigma_d$ are the mean and covariance of $q(\delta_d ; \bw_d, \nu)$ respectively. 

We also use data subsampling to handle large collections of documents
\citep{Hoffman2013} and set $S=1$. Denote by $\mathcal{B}$ a minibatch of documents. Then 
the approximation of the \gls{ELBO} using data subsampling is 
\begin{align}
  \label{eq:elbo3}
  \tilde{\cL} (\alpha, \rho, \nu) & =
	\frac{D}{\vert\mathcal{B}\vert} \sum_{d\in\mathcal{B}} \sum_{n=1}^{N_d}
       {\log p(w_{nd} \g \delta_d, \rho, \alpha)} - \frac{D}{\vert\mathcal{B}\vert}
        \sum_{d\in\mathcal{B}} \mathrm{KL}(q(\delta_d ; \bw_d, \nu) \; || \; p(\delta_d))
\end{align}
Finally, given the prior $p(\delta_d)$ and $q(\delta_d ; \bw_d, \nu)$ are both Gaussians, 
the $\mathrm{KL}$ is closed-form,
\begin{align}
  \label{eq:kl}
  &\mathrm{KL}(q(\delta_d ; \bw_d, \nu) \; || \; p(\delta_d)) =
     \frac{1}{2}\left\{
     	\mathrm{tr}(\Sigma_d) + \mu_d^\top\mu_d - \log \text{det}(\Sigma_d) - K
     \right\}.
\end{align}
Here both $\mu_d$ and $\Sigma_d$ depend implicitly on $\nu$ and $\bw_d$ via the inference network.

We optimize the \gls{ELBO} with respect to both the model parameters $(\alpha, \rho)$ 
and the variational parameters $\nu$. We set the learning rate with Adam \citep{Kingma2015}.
The procedure is shown in Algorithm\nobreakspace \ref {alg:etm}, where the notation
$\textrm{NN}(\mathbf{x}\prm \nu)$ represents a neural network with input
$\mathbf{x}$ and parameters $\nu$.

\begin{algorithm}[t]
  \caption{Flexible topic modeling with the \gls{ETM}}\label{alg:etm}
  \begin{algorithmic}
    \STATE Initialize model and variational parameters
    \FOR{iteration $i = 1, 2, \ldots$}
    \STATE Compute $\beta_k = \text{softmax}(\rho^\top \alpha_k)$ for each topic $k$
    \STATE Choose a minibatch $\mathcal{B}$ of documents
    \FOR{each document $d$ in $\mathcal{B}$}
    \STATE Get normalized bag-of-word representat.\ $\bx_d$
    \STATE Compute $\mu_d = \textrm{NN}(\bx_d\prm \nu_{\mu})$
    \STATE Compute $\Sigma_d = \textrm{NN}(\bx_d\prm \nu_{\Sigma})$
    \STATE Sample $\theta_d \sim \Lcal\Ncal(\mu_d, \Sigma_d)$
    \FOR{each word in the document}
    \STATE Compute $p(w_{dn} \g \theta_d) = \theta_d^\top\beta_{\cdot,w_{dn}}$
    \ENDFOR
    \ENDFOR
    \STATE Estimate the \acrshort{ELBO} and its gradient (backprop.)
    \STATE Update model parameters $\alpha_{1:K}$
    \STATE Update variational parameters ($\nu_\mu$, $\nu_\Sigma$)
    \ENDFOR
  \end{algorithmic} 
\end{algorithm}

\subsection{Related Work}\label{sec:related}

One of the goals in developing the \gls{ETM} is to incorporate word
similarity into the topic model, and there is previous research that
shares this goal.  These methods either modify the topic
priors~\citep{petterson2010word, zhao2017metalda, shi2017jointly,
  zhao2017word} or the topic assignment
priors~\citep{xie2015incorporating}. For example
\citet{petterson2010word} use a word similarity graph (as given by a
thesaurus) to bias \gls{LDA} towards assigning similar words to
similar topics. As another example, \citet{xie2015incorporating} model
the per-word topic assignments of \gls{LDA} using a Markov random
field to account for both the topic proportions and the topic
assignments of similar words.  These methods use word
similarity as a type of ``side information'' about language; in
contrast, the \gls{ETM} directly models the similarity (via
embeddings) in its generative process of words.

Other work has extended \gls{LDA} to directly involve word embeddings.
One common strategy is to convert the discrete text into continuous
observations of embeddings, and then adapt \gls{LDA} to generate
real-valued data~\citep{das2015gaussian, xun2016topic,
  batmanghelich2016nonparametric, xun2017correlated}.  With this
strategy, topics are Gaussian distributions with latent means and
covariances, and the likelihood over the embeddings is modeled with a
Gaussian~\citep{das2015gaussian} or a Von-Mises Fisher
distribution~\citep{batmanghelich2016nonparametric}.  The \gls{ETM}
differs from these approaches in that it is a model of categorical
data, one that goes through the embeddings matrix.  Thus it does not
require pre-fitted embeddings and, indeed, can learn embeddings as
part of its inference process.

There have been a few other ways of combining \gls{LDA} and
embeddings. \citet{nguyen2015improving} mix the likelihood defined by
\gls{LDA} with a log-linear model that uses pre-fitted word
embeddings; \citet{bunk2018welda} randomly replace words drawn from a
topic with their embeddings drawn from a Gaussian; and
\citet{xu2018distilled} adopt a geometric perspective, using
Wasserstein distances to learn topics and word embeddings jointly.

Another thread of recent research improves topic modeling inference
through deep neural networks \citep{srivastava2017autoencoding,
  card2017neural,cong2017deep, zhang2018whai}. Specifically, these
methods reduce the dimension of the text data through amortized
inference and the variational
auto-encoder~\citep{kingma2014autoencoding, rezende2014stochastic}.
To perform inference in the \gls{ETM}, we also avail ourselves of
amortized inference methods \citep{Gershman2014}.

Finally, as a document model, the \gls{ETM} also relates to works that
learn per-document representations as part of an embedding
model~\citep{Le2014, moody2016mixing, miao2016neural}. In contrast to
these works, the document variables in the \gls{ETM} are part of a larger
probabilistic topic model.

\subsection{Empirical Study}\label{sec:experiments}

We study the performance of the \gls{ETM} and compare it to other
unsupervised document models.  A good document model should provide
both coherent patterns of language and an accurate distribution of
words, so we measure performance in terms of both predictive accuracy
and topic interpretability.  We measure accuracy with log-likelihood
on a document completion task \citep{rosenzvi2004author,wallach2009evaluation};
we measure topic interpretability as a
blend of topic coherence and diversity.  We find that, of the interpretable
models, the \gls{ETM} is the one that provides better predictions and
topics.

\begin{table*}[t]
  \centering \captionof{table}[Word embeddings learned by different document models]{Word embeddings learned by all document
    models (and skip-gram) on the \textit{New York Times} with
    vocabulary size $118{,}363$.}  \vskip 0.1in
  \makebox[\textwidth][c]{
  \begin{tabular}{llllllll}
\multicolumn{4}{c}{Skip-gram embeddings} & \multicolumn{4}{c}{\gls{ETM} embeddings} \\
\cmidrule(r){1-4} \cmidrule(r){5-8}
\bf{love} & \bf{family} & \bf{woman} & \bf{politics}  & \bf{love} &  \bf{family} & \bf{woman} & \bf{politics}\\
loved & families  & man & political & joy & children & girl & political\\
passion &grandparents & girl & religion & loves & son & boy & politician \\
loves & mother & boy & politicking & loved & mother &  mother & ideology \\
affection & friends & teenager & ideology & passion & father & daughter  & speeches\\
adore & relatives  & person & partisanship & wonderful & wife & pregnant & ideological\\
\cmidrule(r){1-4} \cmidrule(r){5-8}
 \end{tabular}
 }
   \vskip 0.1in
   \makebox[\textwidth][c]{
  \begin{tabular}{llllllll}
\multicolumn{4}{c}{\acrshort{NVDM} embeddings} & \multicolumn{4}{c}{$\Delta$-\acrshort{NVDM} embeddings} \\
\cmidrule(r){1-4} \cmidrule(r){5-8}
\bf{love} & \bf{family} & \bf{woman} & \bf{politics}  & \bf{love} &  \bf{family} & \bf{woman} & \bf{politics}\\
loves & sons & girl & political & miss & home & life &  political   \\
passion & life & women & politician & young & father & marriage &  faith  \\
wonderful & brother & man & politicians & born & son &  women & marriage   \\
joy & son & pregnant & politically & dream & day &  read &  politicians  \\
beautiful &  lived & boyfriend & democratic & younger & mrs &  young & election   \\
\cmidrule(r){1-4} \cmidrule(r){5-8}
 \end{tabular}
 }
  \vskip 0.1in
   \makebox[\textwidth][c]{
  \begin{tabular}{llll}
\multicolumn{4}{c}{\acrshort{PRODLDA} embeddings}  \\
\cmidrule(r){1-4}
\bf{love} & \bf{family} & \bf{woman} & \bf{politics}  \\
loves & husband & girl & political \\
affection & wife & boyfriend & politician  \\
sentimental & daughters &  boy & liberal \\
dreams & sister & teenager & politicians \\
laugh & friends & ager & ideological  \\
\cmidrule(r){1-4} 
 \end{tabular}
 }
\label{tab:embeddings}
\end{table*}

In a separate analysis, we study the robustness of each method in the presence of stop words. 
Standard topic models fail in this regime---since stop words appear in many documents,
every learned topic includes some stop words, leading to poor topic interpretability.
In contrast, the \gls{ETM} is able to use the information from the word
embeddings to provide interpretable topics. 

\begin{table*}[!hbpt]
  \centering \small \captionof{table}[Topics discovered by different document models]{Top five words of seven most
    used topics from different document models on $1.8$M documents of
    the \textit{New York Times} corpus with vocabulary size $212{,}237$
    and $K=300$ topics.}  \vskip 0.1in
 \begin{tabular}{llllllll}
 \toprule
\multicolumn{7}{c}{LDA}\\
 \hline
     time & year & officials & mr & city & percent & state  \\
     day & million & public & president & building & million & republican  \\
     back & money & department & bush & street  & company  & party  \\
     good & pay & report  & white & park & year  & bill \\
     long & tax & state  & clinton & house & billion  & mr  \\
     \midrule 
\multicolumn{7}{c}{\acrshort{NVDM}}\\
\hline
     scholars & japan & gansler & spratt & assn & ridership & pryce \\
     gingrich & tokyo & wellstone & tabitha & assoc & mtv & mickens  \\
     funds & pacific & mccain & mccorkle & qtr & straphangers & mckechnie   \\
     institutions & europe & shalikashvili & cheetos & yr & freierman & mfume \\
     endowment & zealand & coached & vols  & nyse  & riders & filkins \\
\midrule
\multicolumn{7}{c}{$\Delta$-\acrshort{NVDM}}\\
\hline
     concerto & servings & nato & innings & treas & patients & democrats  \\
     solos & tablespoons & soviet & scored & yr & doctors  & republicans \\
     sonata & tablespoon & iraqi & inning & qtr & medicare & republican \\
     melodies & preheat  &  gorbachev & shutout & outst  & dr & senate \\
     soloist & minced &  arab & scoreless & telerate & physicians & dole \\
 \midrule 
\multicolumn{7}{c}{\acrshort{PRODLDA}}\\
\hline
temptation & grasp & electron & played & amato & briefly & giant \\
repressed & unruly & nuclei & lou & model & precious & boarding \\
drowsy & choke &  macal & greg & delaware & serving & bundle \\
addiction & drowsy & trained & bobby & morita & set & distance\\
conquering & drift & mediaone & steve & dual & virgin & foray \\
 \midrule 
\multicolumn{7}{c}{Labelled \acrshort{PRODLDA}}\\
\hline
mercies & cheesecloth & scoreless & chapels & distinguishable & floured & gillers  \\
lockbox & overcook & floured & magnolias & cocktails  & impartiality & lacerated  \\
pharm & strainer & hitless & asea  & punishable  & knead & polshek  \\
shims & kirberger & asterisk & bogeyed & checkpoints  & refrigerate  & decimated   \\
cp & browned & knead & birdie & disobeying & tablespoons & inhuman  \\
\midrule
\multicolumn{7}{c}{Labelled \gls{ETM}}\\
\hline
     music & republican & yankees & game &  wine &  court & company  \\
     dance & bush & game & points & restaurant &  judge & million \\
     songs & campaign & baseball & season & food &  case & stock \\
     opera & senator & season & team & dishes &  justice & shares \\
     concert & democrats & mets & play & restaurants &  trial & billion  \\
\midrule 
\multicolumn{7}{c}{\gls{ETM}}\\
\hline
     game & music & united & wine & company & yankees &  art\\
     team & mr & israel & food & stock & game & museum    \\
     season & dance & government & sauce &  million & baseball & show \\
     coach & opera & israeli & minutes & companies & mets &  work \\
     play &  band & mr &  restaurant & billion &  season & artist  \\
 \bottomrule
 \end{tabular}
\label{tab:topics}
\end{table*}

\parhead{Corpora.} We study the \textit{20Newsgroups} corpus and the
\textit{New York Times} corpus.

The \textit{20Newsgroup} corpus is a collection of newsgroup posts. We
preprocess the corpus by filtering stop words, words with document
frequency above 70\%, and tokenizing.  To form the vocabulary, we keep
all words that appear in more than a certain number of documents, and
we vary the threshold from 100 (a smaller vocabulary, where
$V = 3{,}102$) to 2 (a larger vocabulary, where $V=52{,}258$). After
preprocessing, we further remove one-word documents from the
validation and test sets.
We split the corpus into a training set of $11{,}260$ documents, a test
set of $7{,}532$ documents, and a validation set of $100$ documents.

\begin{figure*}[t]
  \centering
  \caption[Generalization performance and interpretability of different document models on the \textit{20NewsGroup} corpus]{Interpretability as measured by the exponentiated product of topic coherence and topic diversity (the higher the better) vs.\ predictive performance as measured by log-likelihood on document completion (the higher the better) on the \textit{20NewsGroup} dataset. Both interpretability and predictive power metrics are normalized by subtracting the mean and dividing by the standard deviation across models. Better models are on the top right corner. Overall, the \gls{ETM} is a better topic model.}
  \label{fig:scatter_20ng}
  \centering{\includegraphics[width=1.\linewidth]{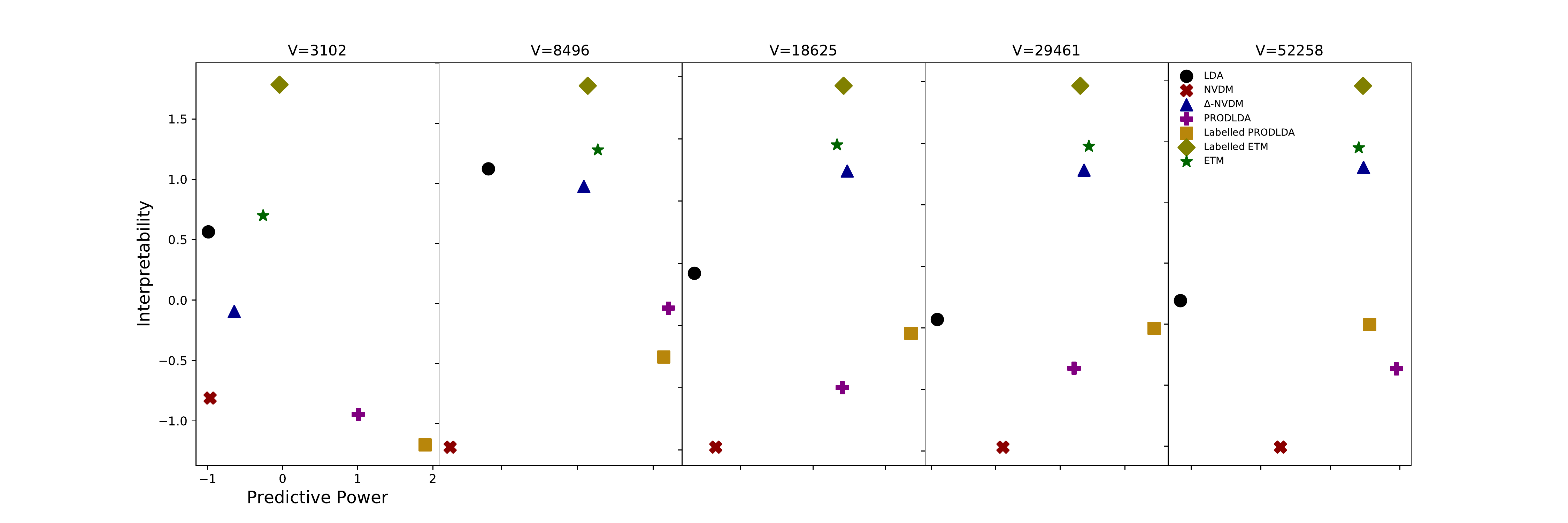}} 
 \end{figure*}

\begin{figure*}[t]
  \centering
  \caption[Generalization performance and interpretability of different document models on the \textit{New York Times} corpus]{Interpretability as measured by the exponentiated product of topic coherence and topic diversity (the higher the better) vs.\ predictive performance as measured by log-likelihood on document completion (the higher the better) on the \textit{New York Times} dataset. Both interpretability and predictive power metrics are normalized by subtracting the mean and dividing by the standard deviation across models. Better models are on the top right corner. Overall, the \gls{ETM} is a better topic model.}
  \label{fig:scatter_nyt}
  \centering{\includegraphics[width=1.\linewidth]{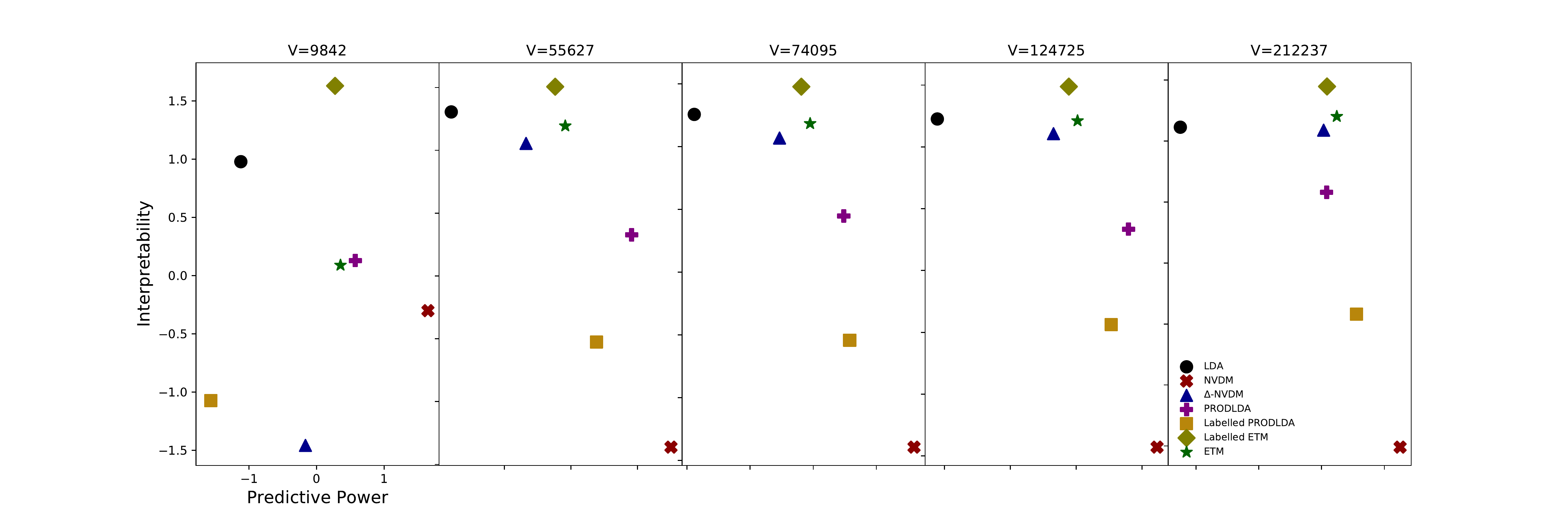}} 
 \end{figure*}
 
The \textit{New York Times} corpus is a larger collection of news
articles.  It contains more than $1.8$ million articles, spanning the
years 1987--2007.  We follow the same preprocessing steps as for
\textit{20Newsgroups}.  We form versions of this corpus with
vocabularies ranging from $V=5{,}921$ to $V=212{,}237$.  After
preprocessing, we use $85\%$ of the documents for training, $10\%$ for
testing, and $5\%$ for validation.

\glsreset{LDA}
\glsreset{NVDM}

\parhead{Models.} We compare the performance of the \gls{ETM} against several document models. 
We briefly describe each below.

We consider \gls{LDA} \citep{blei2003latent}, a standard topic model that posits
Dirichlet priors for the topics $\beta_k$ and topic proportions
$\theta_d$. (We set the prior hyperparameters to $1$.) It is a
conditionally conjugate model, amenable to variational inference with
coordinate ascent.  We consider \gls{LDA} because it is the most
commonly used topic model, and it has a similar generative process as
the \gls{ETM}.
 
We also consider the \gls{NVDM} \citep{miao2016neural}. The \gls{NVDM} is a multinomial 
factor model of documents; it posits the likelihood
$w_{dn}\sim \text{softmax}(\beta^\top \theta_d)$, where the
$K$-dimensional vector $\theta_d\sim \mathcal{N}(\bzero, \bI_K)$ is a
per-document variable, and $\beta$ is a real-valued matrix of size
$K \times V$.  The \gls{NVDM} uses a per-document real-valued latent vector
$\theta_d$ to average over the embedding matrix $\beta$ in the logit
space.  Like the \gls{ETM}, the \gls{NVDM} uses amortized variational
inference to jointly learn the approximate posterior over the document
representation $\theta_d$ and the model parameter $\beta$.

\gls{NVDM} is not interpretable as a topic model; its latent variables
are unconstrained.  We study a more interpretable variant of the
\gls{NVDM} which constrains $\theta_d$ to lie in the simplex,
replacing its Gaussian prior with a logistic
normal~\citep{Aitchison:1980}.  (This can be thought of as a
semi-nonnegative matrix factorization.)  We call this document model
$\Delta$-\gls{NVDM}.

We also consider \acrshort{PRODLDA}~\citep{srivastava2017autoencoding}. It posits the likelihood
$w_{dn}\sim \text{softmax}(\beta^\top \theta_d)$ where the topic proportions 
$\theta_d$ are from the simplex. Contrary to \gls{LDA}, the topic-matrix $\beta$ 
is unconstrained. \acrshort{PRODLDA} is fit using amortized variational inference 
with batch normalization~\citep{ioffe2015batch} and dropout~\citep{srivastava2014dropout}. 

Finally, we consider a document model that combines \acrshort{PRODLDA} 
with pre-fitted word embeddings. We call this document model Labelled \acrshort{PRODLDA}. 

We study two variants of the \gls{ETM}, one where the word embeddings
are pre-fitted and one where they are learned jointly with the rest of
the parameters.  The variant with pre-fitted embeddings is called the
``labelled \gls{ETM}.'' We use skip-gram embeddings
\citep{mikolov2013distributed}.

\parhead{Algorithm settings.}  Given a corpus, each model comes with an
approximate posterior inference problem.  We use variational inference
for all of the models and employ \gls{SVI} \citep{Hoffman2013} to
speed up the optimization. The minibatch size is $1{,}000$ documents.
For \gls{LDA}, we set the learning rate as suggested by \citet{Hoffman2013}: the
delay is $10$ and the forgetting factor is $0.85$.

Within \gls{SVI}, \gls{LDA} enjoys coordinate ascent variational
updates, with $5$ inner steps to optimize the local variables.  For
the other models, we use amortized inference over the local variables
$\theta_d$.  We use $3$-layer inference networks and we set the local
learning rate to $0.002$. We use $\ell_2$ regularization on the
variational parameters (the weight decay parameter is
$1.2\times 10^{-6}$).

\parhead{Qualitative results.}  We first examine the embeddings.  The
\gls{ETM}, \gls{NVDM}, $\Delta$-\gls{NVDM}, and \acrshort{PRODLDA} all involve a word
embedding.  We illustrate them by fixing a set of terms and
calculating the words that occur in the neighborhood around them.  For
comparison, we also illustrate word embeddings learned by the
skip-gram model.
 
Table\nobreakspace \ref {tab:embeddings} illustrates the embeddings of the different
models.  All the methods provide interpretable embeddings---words with
related meanings are close to each other. The \gls{ETM}, the
\gls{NVDM}, and \acrshort{PRODLDA} learn embeddings that are similar to those from the
skip-gram.  The embeddings of $\Delta$-\gls{NVDM} are different; the
simplex constraint on the local variable changes the nature of the
embeddings.

We next look at the learned topics. Table\nobreakspace \ref {tab:topics} displays the $7$
most used topics for all methods, as given by the average of the topic
proportions $\theta_d$.  \gls{LDA} and the \gls{ETM} both provide
interpretable topics. The rest of the models do not 
provide interpretable topics; their model parameters $\beta$ 
are not interpretable as distributions over the vocabulary that mix to form documents.

\parhead{Quantitative results.} We next study the models
quantitatively.  We measure the quality of the topics and the
predictive performance of the model.  We found that among models with
interpretable topics, the \gls{ETM} provides the best predictions.

We measure topic quality by blending two metrics: topic coherence and
topic diversity.  Topic coherence is a quantitative measure of the
interpretability of a topic \citep{mimno2011optimizing}.  It is the average pointwise mutual
information of two words drawn randomly from the same document \citep{lau2014machine},
\begin{equation*}
  \textrm{TC} = \frac{1}{K}\sum_{k=1}^{K} \frac{1}{45} \sum_{i=1}^{10}
  \sum_{j=i+1}^{10} f(w_i^{(k)}, w_j^{(k)}),
\end{equation*}
where $\{w_1^{(k)},\ldots,w_{10}^{(k)}\}$ denotes the top-$10$ most
likely words in topic $k$.  Here, $f(\cdot,\cdot)$ is the normalized
pointwise mutual information,
\begin{equation*}
  f(w_i, w_j) = \frac{\log\frac{P(w_i,w_j)}{P(w_i)P(w_j)}}{-\log
    P(w_i,w_j)}.
\end{equation*}
The quantity $P(w_i,w_j)$ is the probability of words $w_i$ and $w_j$
co-occurring in a document and $P(w_i)$ is the marginal probability of
word $w_i$.  We approximate these probabilities with empirical counts.

The idea behind topic coherence is that a coherent topic will display
words that tend to occur in the same documents. In other words, the
most likely words in a coherent topic should have high mutual
information. Document models with higher topic coherence are more
interpretable topic models.

We combine coherence with a second metric, topic diversity.  We define
topic diversity to be the percentage of unique words in the top $25$
words of all topics. Diversity close to $0$ indicates redundant
topics; diversity close to $1$ indicates more varied topics.  

We define the overall metric for the quality of a model's topics as the 
exponentiated product of its topic diversity and topic coherence.  

A good topic model also provides a good distribution of language.  To
measure predictive quality, we calculate log likelihood on a document
completion task~\citep{rosenzvi2004author,wallach2009evaluation}. We
divide each test document into two sets of words.  The first half is
observed: it induces a distribution over topics which, in turn,
induces a distribution over the next words in the document.  We then
evaluate the second half under this distribution. A good document
model should provide higher log-likelihood on the second half. (For
all methods, we approximate the likelihood by setting $\theta_d$ to
the variational mean.)

We study both corpora and with different
vocabularies. Figure\nobreakspace \ref {fig:scatter_20ng} and Figure\nobreakspace \ref {fig:scatter_nyt} show interpretability 
of the topics as a function of predictive power. (To ease visualization, we normalize both metrics by
subtracting the mean and dividing by the standard deviation.) The best
models are on the upper right corner.

\gls{LDA} predicts worst in almost all settings. On the 
\textit{20NewsGroups}, the \gls{NVDM}'s predictions are in general
better than \gls{LDA} but worse than for the other methods; on the
\textit{New York Times}, the \gls{NVDM} gives the best
predictions. However, topic quality for the \gls{NVDM} is far below
the other methods.  (It does not provide ``topics'', so we assess the
interpretability of its $\beta$ matrix.)  In prediction, both versions
of the \gls{ETM} are at least as good as the simplex-constrained
$\Delta$-\gls{NVDM}. More importantly, both versions of the \gls{ETM} 
outperform the Labelled \acrshort{PRODLDA}; signaling the \gls{ETM} 
provides a better way of integrating word embeddings into a topic model. 

\begin{figure}[t]
 \centering
 \includegraphics[width=0.5\linewidth]{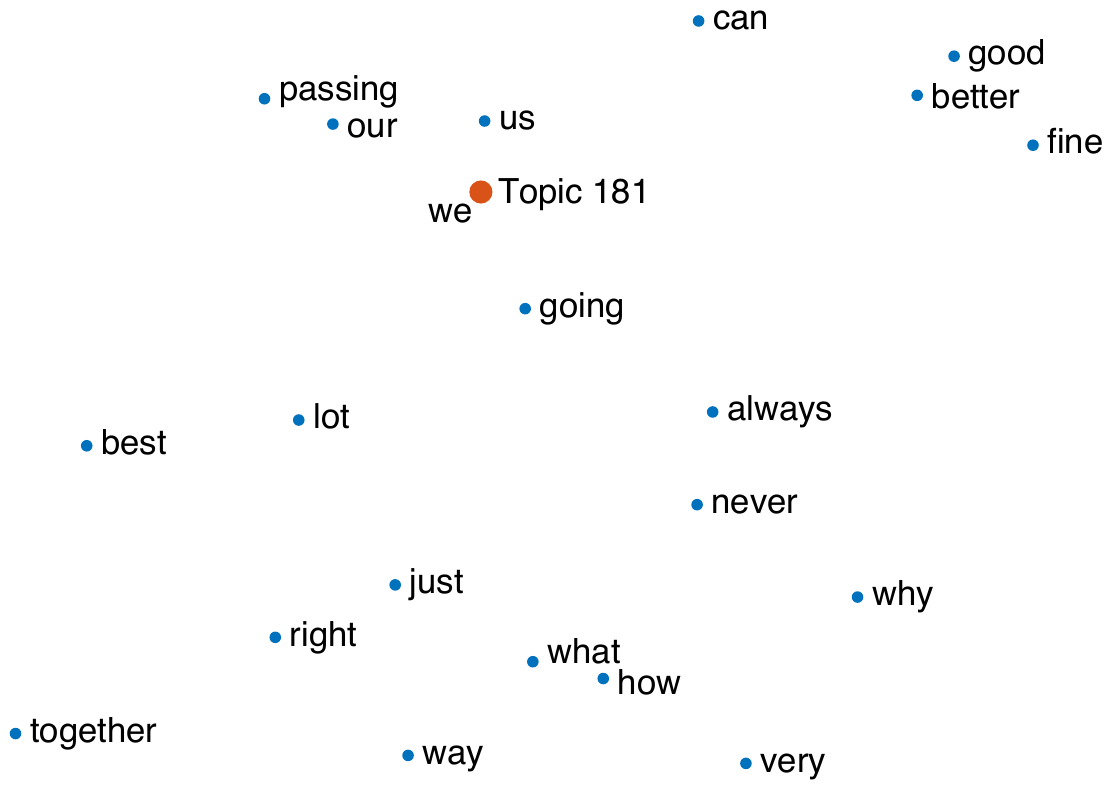}
 \captionof{figure}[The \acrshort{ETM} assigns stop words to their own topic; illustration on \textit{The New York Times} corpus]{A topic containing stop words found by the \acrshort{ETM} on \textit{The New York Times}. The \gls{ETM} is robust even in the presence of stop words.}
 \label{fig:topic_embedding_stops}
\end{figure}

These figures show that, of the interpretable models, the \gls{ETM}
provides the best predictive performance while keeping interpretable
topics.  It is robust to large vocabularies.

\parhead{Stop words}\label{subsec:stopwords}

We now study a version of the \textit{New York Times} corpus that
includes all stop words. We remove infrequent words to form a vocabulary
of size $10{,}283$.  Our goal is to show that the labeled \gls{ETM}
provides interpretable topics even in the presence of stop words,
another regime where topic models typically fail. In particular, given that
stop words appear in many documents, traditional topic models learn topics
that contain stop words, regardless of the actual semantics of the topic.
This leads to poor topic interpretability.

We fit \gls{LDA}, the $\Delta$-\gls{NVDM}, the labelled \acrshort{PRODLDA}, and the labelled \gls{ETM}
with $K=300$ topics. (We do not report the \gls{NVDM} because it does
not provide interpretable topics.) Table\nobreakspace \ref {tab:tc_td_stopwords} shows the logarithm of the 
topic quality (the product of topic coherence and topic
diversity). Overall, the labelled \gls{ETM} gives the best performance
in terms of topic quality.

While the \gls{ETM} has a few ``stop topics'' that are specific for
stop words (see, e.g., Figure\nobreakspace \ref {fig:topic_embedding_stops}),
$\Delta$-\gls{NVDM} and \gls{LDA} have stop words in almost every
topic.  (The topics are not displayed here for space constraints.) The
reason is that stop words co-occur in the same documents as every
other word; therefore traditional topic models have difficulties
telling apart content words and stop words. The labelled \gls{ETM}
recognizes the location of stop words in the embedding space;
its sets them off on their own topic.

\begin{table}[t]
  \centering
  \caption[Topic quality on the \textit{New York Times} data in the
    presence of stop words for different document models]{Topic quality on the \textit{New York Times} data in the
    presence of stop words. Topic quality here is given by the
    product of topic coherence and topic diversity (higher is
    better). The labeled \gls{ETM} is robust to stop words; it achieves
    similar topic coherence than when there are no stop words.
    \label{tab:tc_td_stopwords}}
  {\small
    \begin{tabular}{cccc} \toprule
      & \textsc{tc}  &   \textsc{td}  & Quality\\ \midrule
      \acrshort{LDA}           &  $0.13$ & $0.14$ & $0.0182$  \\
      $\Delta$-\acrshort{NVDM} &   $0.17$        &    $0.11$   & $0.0187$    \\
      Labelled \acrshort{PRODLDA}          &   $0.03$        &    $0.53$    & $0.0159$   \\
      Labeled \acrshort{ETM}    & $\mathbf{0.18}$ & $\mathbf{0.22}$ & $\mathbf{0.0396}$\\ \bottomrule
    \end{tabular}}
\end{table}

\subsection{Conclusion}

We developed the \gls{ETM}, a generative model of documents that
marries \gls{LDA} with word embeddings. The \gls{ETM} assumes that
topics and words live in the same embedding space, and that words are
generated from a categorical distribution whose natural parameter is
the inner product of the word embeddings and the embedding of the
assigned topic.

The \gls{ETM} learns interpretable word embeddings and topics, even in
corpora with large vocabularies. We studied the performance of the
\gls{ETM} against several document models. The \gls{ETM} learns both
coherent patterns of language and an accurate distribution of words. 

The construct used to define the \gls{ETM} can be used to extend all versions 
of \gls{LDA}, e.g. dynamic \gls{LDA}~\citep{blei2006dynamic}, 
supervised \gls{LDA}~\citep{mcauliffe2008supervised}, and correlated \gls{LDA}~\citep{Lafferty2005correlated}. 
In the next section we will apply the \gls{ETM} technique for flexible dynamic topic modeling. 
 
\section{Dynamic Embedded Topic Modeling}\label{sec:detm}

Here we develop the \gls{DETM}, a model that combines the advantages
of \gls{DLDA} and the \gls{ETM}. Like \gls{DLDA}, it allows the topics
to vary smoothly over time to accommodate datasets that span a large
period of time. Like the \gls{ETM}, the \gls{DETM} uses word
embeddings, allowing it to generalize better than \gls{DLDA} and
improving its topics.  We describe the model in Section\nobreakspace \ref {subsec:model}
and then we develop an efficient structured variational inference algorithm in
Section\nobreakspace \ref {subsec:inference}.

\subsection{Model Description}
\label{subsec:model}

The \gls{DETM} is a dynamic topic model that uses embedding
representations of words and topics.  For each term $v$, it considers
an $L$-dimensional embedding representation $\rho_v$.  The \gls{DETM}
posits an embedding $\alpha_k^{(t)}\in\mathbb{R}^L$ for each topic $k$
at a given time stamp $t=1,\ldots,T$.  That is, the \gls{DETM}
represents each topic as a time-varying real-valued vector, unlike
traditional topic models (where topics are distributions over the
vocabulary). We refer to $\alpha_k^{(t)}$ as \textit{topic embedding}
\citep{dieng2019topic}; it is a distributed representation of the
$k^\mathrm{th}$ topic in the semantic space of words.

The \gls{DETM} forms distributions over the vocabulary using the word
and topic embeddings. Specifically, under the \gls{DETM}, the
probability of a word under a topic is given by the (normalized)
exponentiated inner product between the embedding representation of
the word and the topic's embedding at the corresponding time
step, 
\begin{equation}\label{eq:detm_conditional_lik}
  p(w_{dn}=v\g z_{dn}=k, \alpha_k^{(t_d)})\propto \exp\{ \rho_v^\top \alpha_k^{(t_d)} \}.
\end{equation}
The probability of a particular term is higher when the term's
embedding and the topic's embeddings are in agreement. Therefore,
semantically similar words will be assigned to similar topics, since
their representations are close in the embedding space.

\begin{figure}[t]
	\centering
	\includegraphics[width=0.45\textwidth]{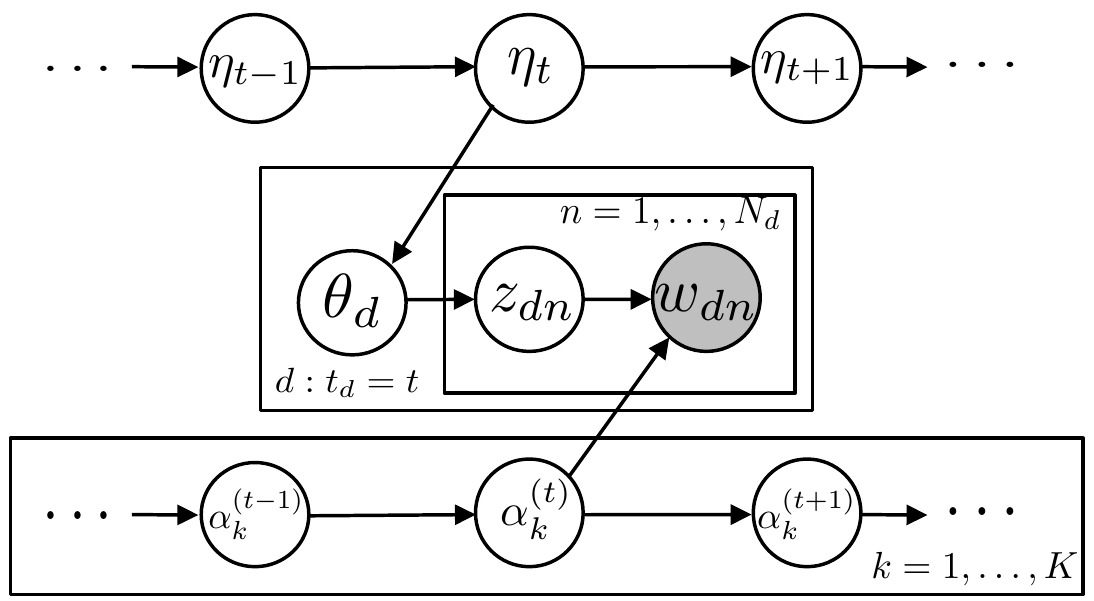}
	\caption[Graphical representation of \gls{DETM}]{Graphical representation of \gls{DETM}. The topic
          embeddings $\alpha_k^{(t)}$ and the latent means $\eta_t$
          evolve over time. For each document at time step $t$, the
          prior over the topic proportions $\theta_d$ depends on
          $\eta_t$. The variables $z_{dn}$ denote the topic
          assignment; the variables $w_{dn}$ denote the words.}
	\label{fig:graphical_model}
\end{figure}

The \gls{DETM} enforces smooth variations of the topics by using a
Markov chain over the topic embeddings $\alpha_k^{(t)}$. The topic
representations evolve under Gaussian noise with variance
$\gamma^2$,~\looseness=-1
\begin{equation}
	p(\alpha_k^{(t)}\g \alpha_k^{(t-1)})=\mathcal{N}(\alpha_k^{(t-1)}, \gamma^2 I).
\end{equation}
Similarly to \gls{DLDA}, the \gls{DETM} considers time-varying priors
over the topic proportions $\theta_d$. In addition to time-varying
topics, this construction allows the model to capture how the general
topic usage evolves over time.  The prior over $\theta_d$ depends on a
latent variable $\eta_{t_d}$ (recall that $t_d$ is the time
stamp of document $d$),
\begin{align*}
  p(\theta_d\g \eta_{t_d})&=\mathcal{LN}(\eta_{t_d}, a^2 I)  \text{ where } 
  p(\eta_t\g \eta_{t-1}) = \mathcal{N}(\eta_{t-1}, \delta^2 I).
\end{align*}
Figure\nobreakspace \ref {fig:graphical_model} depicts the graphical model for the
\gls{DETM}.  The full generative process is as follows:
\begin{compactenum}
\item Draw initial topic embedding $\alpha_k^{(1)} \sim \mathcal{N}(0,I)$.
\item Draw initial topic proportion mean $\eta_1 \sim \mathcal{N}(0,I)$.
\item For time step $t = 2 ,\ldots, T$:
\begin{compactenum}
	\item Draw topic embeddings $\alpha_k^{(t)} \sim \mathcal{N}(\alpha_k^{(t-1)}, \gamma^2 I)$ for $k=1,\ldots,K$.
	\item Draw topic proportion means $\eta_t\sim \mathcal{N}(\eta_{t-1}, \delta^2 I)$.
 \end{compactenum}
\item For each document $d$:
  \begin{compactenum}
    \setlength{\itemindent}{-0.1cm}
  \item Draw topic proportions $\theta_d \sim \mathcal{LN}(\eta_{t_d}, a^2 I)$.
  \item For each word $n$ in the document:
    \begin{compactenum}
      \setlength{\itemindent}{-0.3cm}
    \item Draw topic assignment $z_{dn} \sim \text{Cat}(\theta_d)$.
    \item Draw word $w_{dn} \sim \text{Cat}(\textrm{softmax}(\rho^\top \alpha_{z_{dn}}^{(t_d)}))$.
    \end{compactenum}
  \end{compactenum}
\end{compactenum}
Steps 1 and 3a give the prior over the topic embeddings; they encourage
smoothness on the resulting topics. Steps 2 and 3b are shared with \gls{DLDA};
they describe the evolution of the prior mean over the topic
proportions. Steps 4a and 4b-i are standard for topic modeling; they
represent documents as distributions over topics and draw a topic
assignment for each word. Step 4b-ii is different---it uses the
$L\times V$ word embedding matrix $\rho$ and the assigned topic
embedding $\alpha_{z_{dn}}^{(t_d)}$ at time instant $t_d$ to form a
categorical distribution over the vocabulary.

Since the \gls{DETM} uses embedding representations of the words, it
learns the topics in a particular embedding space. This aspect of the
model is useful when the embedding of a new word is available, i.e., a
word that does not appear in the corpus. Specifically, consider a term
$v^\star$ that was not seen in the corpus. The \gls{DETM} can assign
it to topics by computing the inner products
$\rho_{v^\star}^\top \alpha_k^{(t)}$, thus leveraging the semantic
information of the word's embedding.

\subsection{Structured Amortized Variational Inference}
\label{subsec:inference}

We observe a dataset $\mathcal{D}$ of documents
$\{\bw_1,\ldots,\bw_D \}$ and their time stamps $\{t_1,\ldots,t_D\}$.
Fitting a \gls{DETM} involves finding the posterior distribution over
the model's latent variables, $p(\theta,\eta,\alpha\g \mathcal{D})$,
where we have marginalized out the topic assignments $z$ from
Eq.\nobreakspace \ref {eq:detm_conditional_lik} for convenience,\footnote{
  Marginalizing $z_{dn}$ reduces the number of variational parameters
  and avoids discrete latent variables in the inference procedure,
  which is useful to form reparameterization gradients.}
\begin{equation}
	p(w_{dn}\g \alpha_{1:K}^{(t_d)}) = \sum_{k=1}^{K} p(w_{dn}\g z_{dn}=k, \alpha_k^{(t_d)}).
\end{equation}
The posterior is intractable.  We approximate it with variational
inference \citep{Jordan1999,Blei2017}.

Variational inference approximates the posterior using a family of
distributions $q_{\nu}(\theta,\eta,\alpha)$. The parameters $\nu$ that
index this family are called variational parameters, and are optimized
to minimize the \gls{KL} divergence between the approximation and the
posterior. Solving this optimization problem is equivalent to
maximizing the \gls{ELBO},
\begin{equation}\label{eq:elbo}
  \Lcal(\nu) = \E{q}{\log p(\mathcal{D},\theta,\eta,\alpha) - \log q_{\nu}(\theta,\eta,\alpha)}.
\end{equation}

The model's log-joint distribution in Eq.\nobreakspace \ref {eq:elbo} is
\begin{align}\label{eq:logjoint}
	\log p(\mathcal{D},\theta,\eta,\alpha)
	&= \sum_{k=1}^{K} \sum_{t=1}^{T}  \log p(\alpha_k^{(t)} \g \alpha_k^{(t-1)}) 
   + \sum_{t=1}^{T} \log p(\eta_t\g \eta_{t-1})
	+ \sum_{d=1}^{D} \log p(\theta_d\g\eta_{t_d}) \nonumber \\
	& \quad +  \sum_{d=1}^{D}  \sum_{n=1}^{N_d} \log \left(\sum_{k=1}^{K} \theta_{dk} \beta_{k,w_{dn}}^{(t_d)}\right),
\end{align}
where $\beta_{k,w_{dn}}^{(t_d)}\triangleq \text{softmax}(\rho^\top \alpha_k^{(t_d)})|_{w_{dn}}$,  
$w_{dn}$ denotes the $n^{\textrm{th}}$ word in the $d^{\textrm{th}}$ document,
and $N_d$ is the total number of words of the $d^{\textrm{th}}$ document.

To reduce the number of variational parameters and speed-up the
inference algorithm, we use an amortized variational distribution,
i.e., we let the parameters of the approximating distributions be
functions of the data
\citep{Gershman2014,kingma2014autoencoding}. Additionally, we use a
structured variational family to preserve some of the conditional
dependencies of the graphical model \citep{Saul1996}. The specific
variational family in the \gls{DETM} takes the form
\begin{align}\label{eq:q_factorization}
  q_{\nu}(\theta,\eta,\alpha)
  =&
    \prod_d q(\theta_d\g\eta_{t_d},\bw_d) \times
    \prod_t q(\eta_t\g \eta_{1:t-1}, \widetilde{\bw}_t) \times
    \prod_k \prod_t q(\alpha_k^{(t)} \g \alpha_k^{(1:t-1)}, \widetilde{\bw}_t).
\end{align}
(To avoid clutter, we suppress the notation for the variational
parameters.)

The distribution over the topic proportions
$q(\theta_d\g\eta_{t_d},\bw_d)$ is a logistic-normal whose mean and
covariance parameters are functions of both the latent mean
$\eta_{t_d}$ and the bag-of-words representation of document $d$.  In
particular, these functions are parameterized by feed-forward neural
networks that input both $\eta_{t_d}$ and the normalized bag-of-words
representation.
The distribution over the latent means
$q(\eta_t\g \eta_{1:t-1}, \widetilde{\bw}_t)$ depends on all previous
latent means $\eta_{1:t-1}$. We use an \gls{LSTM} to capture this
temporal dependency. We choose a Gaussian distribution
$q(\eta_t\g \eta_{1:t-1}, \widetilde{\bw}_t)$ whose mean and
covariance are given by the output of the \gls{LSTM}.  The input to
the \gls{LSTM} at time $t$ is formed by the concatenation of $\eta_{t-1}$
and the average of the bag-of-words
representation of all documents whose time stamp is $t$. Here,
$\widetilde{\bw}_t$ denotes the normalized bag-of-words representation
of all such documents. Finally, the distribution over the topic
embeddings $q(\alpha_k^{(t)}\g \alpha_k^{(1:t-1)}, \widetilde{\bw}_t)$
is built analogously, using an \gls{LSTM} to capture the temporal
dependencies.

\begin{algorithm}[tb]
  \caption{Flexible dynamic topic modeling with the \gls{DETM}}
  \label{alg:detm}
  \begin{algorithmic}
     \STATE {\bfseries Input:} Documents $\{\bw_1,\ldots,\bw_D\}$ and their time stamps $\{t_1,\ldots,t_D\}$
     \STATE Initialize all variational parameters
     \FOR{iteration $1,2,3,\ldots$}
      \STATE Sample the latent means and the topic embeddings,\\ \quad $\eta \sim q(\eta\g \widetilde{\bw})$ and $\alpha\sim q(\alpha\g \widetilde{\bw})$
      \STATE Compute the topics $\beta_k^{(t)} = \textrm{softmax}(\rho^\top \alpha_k^{(t)})$ for \\ \quad $k=1,\ldots,K$ and $t=1,\ldots,T$
      \STATE Obtain a minibatch of documents
      \FOR{each document $d$ in the minibatch}
        \STATE Sample the topic proportions $\theta_d \sim q(\theta_d\g \eta_{t_d}, \bw_d)$
        \FOR{each word $n$ in the document}
          \STATE Compute $p(w_{dn} \g \theta_d) = \sum_k \theta_{dk}\beta_{k,w_{dn}}^{(t_d)}$
        \ENDFOR
      \ENDFOR
      \STATE Estimate the \acrshort{ELBO} in Eq.\nobreakspace \ref {eq:mcelbo} and its gradient w.r.t.\\ \quad the variational parameters (backpropagation)
      \STATE Update the model and variational parameters (Adam)
     \ENDFOR
  \end{algorithmic}
\end{algorithm}

We optimize the \gls{ELBO} with respect to the variational
parameters. Because the expectations in Eq.\nobreakspace \ref {eq:elbo} are
intractable, we use black box variational inference, obtaining
unbiased gradient estimators with a Monte Carlo method. In particular, we use one sample 
from the variational distribution to form reparameterization gradients
\citep{kingma2014autoencoding,Titsias2014_doubly,rezende2014stochastic}.

To sample from $q_{\nu}(\theta,\eta,\alpha)$ using reparameterization, we first sample
a set of standard Gaussian auxiliary latent variables $\varepsilon\sim \Ncal(0,I)$
and we then use a deterministic transformation $h_{\nu}(\varepsilon)$ that gives
the samples $(\theta,\eta,\alpha)$. Therefore, the realized values of the latent
variables are now functions of the variational parameters $\nu$, since 
$(\theta,\eta,\alpha) = h_{\nu}(\varepsilon)$. Given these samples, we
estimate the \gls{ELBO} in Eq.\nobreakspace \ref {eq:elbo} as
\begin{align}\label{eq:mcelbo}
	& \Lcal(\nu) 
	\approx \sum_{d=1}^{D}  \sum_{n=1}^{N_d} \log \left(\sum_{k=1}^{K} \theta_{dk} \beta_{k,w_{dn}}^{(t_d)} \right) 
	- \sum_{k=1}^{K} \sum_{t=1}^{T}  \textsc{kl}\left(q(\alpha_k^{(t)} \g \alpha_k^{(1:t-1)}, \widetilde{\bw}_t) \;\vert\vert\; p(\alpha_k^{(t)} \g \alpha_k^{(t-1)}) \right) \nonumber\\
  	&- \sum_{t=1}^{T} \textsc{kl}\left(q(\eta_t\g \eta_{1:t-1}, \widetilde{\bw}_t) \;\vert\vert\; p(\eta_t\g \eta_{t-1})  \right)
	- \sum_{d=1}^{D} \textsc{kl}\left(q(\theta_d\g\eta_{t_d},\bw_d) \;\vert\vert\; p(\theta_d\g\eta_{t_d}) \right).
\end{align}
Here, each \gls{KL} divergence corresponds to the \gls{KL} between two Gaussian
distributions whose parameters are functions of the latent variables in the
conditioning set. Therefore, the \gls{KL} terms can be obtained in closed form
as a function of these latent variables.

The variational optimization problem reduces to a stochastic optimization method
that approximates the gradients $\nabla_{\nu} \Lcal(\nu)$ by differentiating
through Eq.\nobreakspace \ref {eq:mcelbo} w.r.t.\ $\nu$.
To speed up the algorithm, we estimate the sum over documents
by taking a minibatch of documents at each iteration;
this allows to handle large collections of documents
\citep{Hoffman2013}.  We set the learning rate with Adam
\citep{Kingma2015}.  Algorithm\nobreakspace \ref {alg:detm} summarizes the procedure.

\subsection{Related Work}\label{sec:related}

The \gls{DETM} builds on word embeddings, topic models, and dynamic
topic models.

Word embeddings are low-dimensional continuous representations of
words that capture their
semantics~\citep{Rumelhart:1973,Bengio:2003,bengio2006neural,mikolov2013efficient,
  mikolov2013distributed,pennington2014glove,levy2014neural}.  Some
recent work finds embedding representations that vary over time
\citep{Bamler2017,Rudolph2018}. Despite incorporating a time-varying
component, these works have a different goal than the
\gls{DETM}. Rather than modeling the temporal evolution of documents,
they model how the meaning of words shifts over time.  (In future
research, the \gls{DETM} developed here could be used in concert with
these methods.)

There has been a surge of methods that combine word embeddings and
probabilistic topic models.  Some methods modify the prior
distributions over topics in \gls{LDA}~\citep{petterson2010word,
  xie2015incorporating, shi2017jointly, zhao2017word,
  zhao2017metalda}.  These methods use word embeddings as a type of
``side information.''  There are also methods that combine \gls{LDA}
with word embeddings by first converting the discrete text into
continuous observations of embeddings \citep{das2015gaussian,
  xun2016topic, batmanghelich2016nonparametric, xun2017correlated}.
These works adapt \gls{LDA} for real-valued observations, for example
using a Gaussian likelihood.  Still other ways of combining \gls{LDA}
and word embeddings modify the likelihood \citep{nguyen2015improving},
randomly replace words drawn from a topic with the embeddings drawn
from a Gaussian \citep{bunk2018welda}, or use Wasserstein distances to
learn topics and embeddings jointly \citep{xu2018distilled}.  In
contrast to all these methods, the \gls{DETM} uses sequential priors
and is a probabilistic model of discrete data that directly models the
words.

Another line of research improves topic modeling inference through
deep neural networks; these are called neural topic models
\citep{miao2016neural,srivastava2017autoencoding,card2017neural,
  cong2017deep,zhang2018whai}. Most of these works are based on the
variational autoencoder \citep{kingma2014autoencoding} and use
amortized inference \citep{Gershman2014}.  Finally, the \gls{ETM}
\citep{dieng2019topic} is a probabilistic topic model that also makes
use of word embeddings and uses amortization in its inference
procedure.

The first and most common dynamic topic model is \gls{DLDA}
\citep{blei2006dynamic}.  \citet{Bhadury2016} scale up the inference
method of \gls{DLDA} using a sampling procedure.  Other extensions of
\gls{DLDA} use stochastic processes to introduce stronger correlations
in the topic dynamics \citep{Wang2006,Wang2008,Jahnichen2018}.  The
\gls{DETM} is also an extension of \gls{DLDA}, but developed for a
different purpose. The \gls{DETM} better fits the distribution of
words via the use of distributed representations for both the words
and the topics.

\subsection{Empirical Study}\label{sec:experiments}

We use the \gls{DETM} to analyze the transcriptions of the \gls{UN}
general debates from $1970$ to $2015$, a corpus of \textsc{acl} abstracts from $1973$ to $2006$, and a set of articles from Science Magazine from $1990$ to $1999$. 
We found the \gls{DETM} provides better predictive power and higher topic quality in general on these datasets when compared to \gls{DLDA}. 

On the transcriptions of the \gls{UN} general debates, we additionally carried out a qualitative analysis of the results. We found that
the \gls{DETM} reveals the temporal evolution of the topics discussed in the debates (such as
climate change, war, poverty, or human rights).

We compared the \gls{DETM} against two versions of \gls{DLDA}, labeled as \gls{DLDA}
and \gls{DLDA}-\textsc{rep}, which differ only in the inference method (the details are below). 
The comparison of the \gls{DETM} against \gls{DLDA}-\textsc{rep}
reveals that the key to the \gls{DETM}'s performance is the model and not simply the scalable inference procedure.

\begin{table*}[t]
	\centering
	\captionof{table}[Summary statistics of the \textsc{un}, \textsc{science}, and \textsc{acl} datasets for dynamic topic modeling.]{Summary statistics of the \textsc{un}, \textsc{science}, and \textsc{acl} datasets.}
	\begin{tabular}{cccccc}
	\toprule
	 Dataset & \# Docs Train & \# Docs Val & \# Docs Test & \# Timestamps & Vocabulary \\
	 \midrule
	 \textsc{un}    & $196{,}290$ & $11{,}563$ & $23{,}097$ & $46$ & $12{,}466$ \\
	 \textsc{science}  &  $13{,}894$ &      $819$ &  $1{,}634$ & $10$ & $25{,}987$ \\
	 \textsc{acl}    &   $8{,}936$ &      $527$ &  $1{,}051$ & $31$ & $35{,}108$ \\
	 \bottomrule
	\end{tabular}
	\label{tab:datasets}
\end{table*}

\begin{table*}[t]
	\centering
	\captionof{table}[Comparing predictive power, interpretability, and runtime of different dynamic topic models on the \textsc{un} dataset]{Performance as measured by perplexity (\textsc{ppl}), topic coherence (\textsc{tc}), topic diversity (\textsc{td}), topic quality (\textsc{tq}), and runtime (in minutes per epoch) on the \textsc{un} dataset. The \acrshort{DETM} achieves better predictive and qualitative performance than \gls{DLDA} and \gls{DLDA}-\textsc{rep} and runs significantly faster than \gls{DLDA}.\label{tab:quantitative}}
		\centering
		\begin{tabular}{cccccc}
		 \midrule
		 \textsc{method} & \textsc{ppl} & \textsc{tc} & \textsc{td} & \textsc{tq}  & \textsc{runtime} \\
		 \midrule
		 \gls{DLDA} \citep{blei2006dynamic} & $2393.5$ & $\textbf{0.1317}$ & $0.6065$ & $0.0799$ & $28.70$ \\
		 \gls{DLDA}-\textsc{rep}  & $2931.3$ & $0.1180$ & $0.2691$ & $0.0318$ & $6.00$ \\
		  \acrshort{DETM}  & $\textbf{1970.7}$  & $0.1206$ & $\textbf{0.6703}$ & $\textbf{0.0809}$ & $3.70$ \\
		 \bottomrule
		\end{tabular}
\end{table*}
	
\begin{table*}[t]
	\centering
	\captionof{table}[Comparing predictive power, interpretability, and runtime of different dynamic topic models on the \textsc{science} dataset]{Performance as measured by perplexity (\textsc{ppl}), topic coherence (\textsc{tc}), topic diversity (\textsc{td}), topic quality (\textsc{tq}), and runtime (in minutes per epoch) on the \textsc{science} dataset. The \acrshort{DETM} achieves better predictive and qualitative performance than \gls{DLDA} and \gls{DLDA}-\textsc{rep} and runs significantly faster than \gls{DLDA}.\label{tab:quantitative}}
		\centering
		\begin{tabular}{cccccc}
		 \midrule
		 \textsc{method} & \textsc{ppl} & \textsc{tc} & \textsc{td} & \textsc{tq} & \textsc{runtime}  \\
		 \midrule
		 \gls{DLDA} \citep{blei2006dynamic} &  $\textbf{3600.7}$ & $\textbf{0.2392}$ & $0.6502$ & $0.1556$  & $15.03$ \\
		 \gls{DLDA}-\textsc{rep}  & $8377.4$ & $0.0611$ & $0.2290$ & $0.0140$ & $0.36$ \\
		  \acrshort{DETM}  & $4206.1$ &  $0.2298$ & $\textbf{0.8215}$ & $\textbf{0.1888}$ & $0.47$ \\
		 \bottomrule
		\end{tabular}
\end{table*}

\begin{table*}[t]
	\centering
	\captionof{table}[Comparing predictive power, interpretability, and runtime of different dynamic topic models on the \textsc{acl} dataset]{Performance as measured by perplexity (\textsc{ppl}), topic coherence (\textsc{tc}), topic diversity (\textsc{td}), topic quality (\textsc{tq}), and runtime (in minutes per epoch) on the \textsc{acl} dataset. The \acrshort{DETM} achieves better predictive and qualitative performance than \gls{DLDA} and \gls{DLDA}-\textsc{rep} and runs significantly faster than \gls{DLDA}.\label{tab:quantitative}}
		\centering
		\begin{tabular}{cccccc}
		 \midrule
		 \textsc{method} & \textsc{ppl} & \textsc{tc} & \textsc{td} & \textsc{tq} & \textsc{runtime} \\
		 \midrule
		 \gls{DLDA} \citep{blei2006dynamic} & $4324.2$ &  $0.1429$ & $0.5904$ & $0.0844$ & $26.30$ \\
		 \gls{DLDA}-\textsc{rep} & $5836.7$ & $0.1011$ & $0.2589$ & $0.0262$ & $1.60$ \\
		  \acrshort{DETM}  & $\textbf{4120.6}$ &$\textbf{0.1630}$ & $\textbf{0.8286}$ & $\textbf{0.1351}$ & $0.75$ \\
		 \bottomrule
		\end{tabular}
\end{table*}

\parhead{Datasets.}  We study the \gls{DETM} on three datasets.  The
\gls{UN} debates corpus\footnote{See
  \url{https://www.kaggle.com/unitednations/un-general-debates}.}
spans $46$ years \citep{Baturo2017understanding}.  Each year, leaders
and other senior officials deliver statements that present their
government's perspective on the major issues in world politics.  The
corpus contains the transcriptions of each country's statement at the
\gls{UN} General Assembly.  We follow \citet{Lefebure2018} and split
the speeches into paragraphs, treating each paragraph as a separate
document.

The second dataset contains ten years of \textsc{science} articles, from $1990$
to $1999$. The articles are from \textsc{jstor}, an on-line archive of
scholarly journals that scans bound volumes and runs optical character
recognition algorithms on the scans.  This data was used by
\citet{blei2007correlated}.

The third dataset is a collection of articles from $1973$ to $2006$
from the \textsc{acl} Anthology \citep{Bird2008}.  This anthology is a
repository of computational linguistics and natural language
processing papers.

For each dataset, we apply standard preprocessing techniques, such as
tokenization and removal of numbers and punctuation marks.  We also
filter out stop words, i.e., words with document frequency above
$70\%$, as well as standard stop words from a list. Additionally, we
remove low-frequency words, i.e., words that appear in less than a
certain number of documents ($30$ documents for \gls{UN} debates, 
$100$ for the \textsc{science} corpus, and $10$ for the \textsc{acl} dataset). 
We use $85\%$ randomly chosen documents for training, $10\%$ for testing, and $5\%$ for validation,
and we remove one-word documents from the validation and test
sets. Table\nobreakspace \ref {tab:datasets} summarizes the characteristics of each dataset.

\parhead{Methods.} We compare the \gls{DETM} against two variants of
\gls{DLDA}. One variant is the original model and algorithm of
\citet{blei2006dynamic}. The other variant, which we call \gls{DLDA}-\textsc{rep}, is the 
\gls{DLDA} model fitted using mean-field variational inference with the reparameterization trick. 
The comparison against \gls{DLDA}-\textsc{rep} helps us delineate between performance due to the model
and performance due to the inference algorithm.

\begin{figure*}[t]
	\centering
	\includegraphics[width=1.0\linewidth]{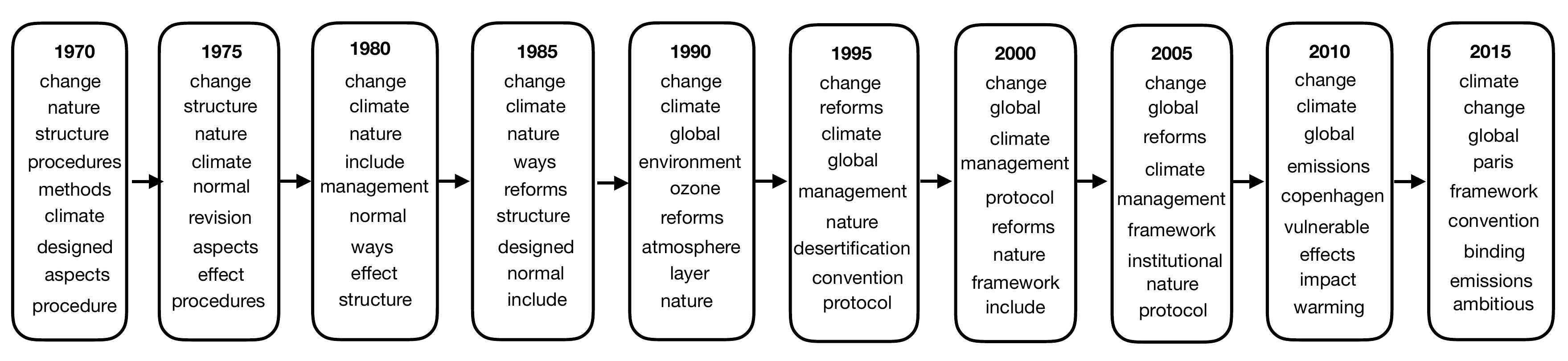}
	\vspace{-0.1in}
	\caption[Temporal evolution of the top-$10$ words from a topic about climate change learned by the \acrshort{DETM}]{Temporal evolution of the top-$10$ words from a topic about climate change learned by the \acrshort{DETM}. This topic is in agreement with historical events. In the 1990s the destruction of the ozone layer was of major concern. More recently the concern is about global warming. Events such as the Kyoto protocol and the Paris convention are also reflected in this topic's evolution.}
	\label{fig:topic_evolution_climate_change}
\end{figure*}

\begin{figure*}[!hbpt]
	\centering
	\includegraphics[width=1.0\linewidth]{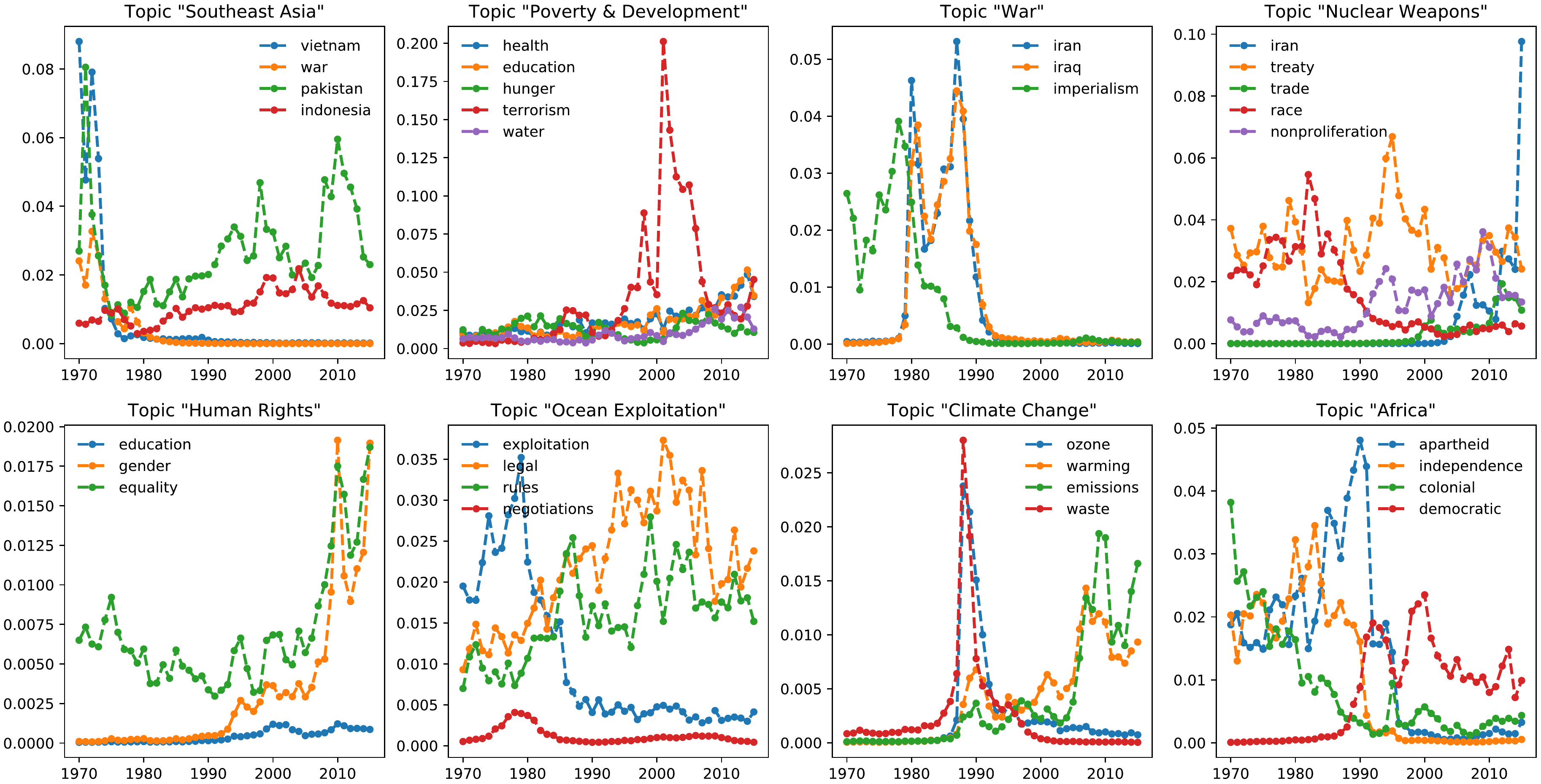}
	\vspace{-0.1in}
	\caption[Language evolution over time discovered by the \acrshort{DETM}]{Evolution of word probability across time for eight different topics learned by the \acrshort{DETM}. For each topic, we choose a set of words whose probability shift aligns with historical events (these are not the words with the highest probability in each topic). For example, one interesting finding is the increased relevance of the words ``gender'' and ``equality'' in a topic about human rights.}
	\label{fig:word_evolution}
\end{figure*}

\parhead{Settings.} We use $50$ topics for all the experiments and
follow \citet{blei2006dynamic} to set the variances of the different priors as
$\delta^2=\sigma^2=\gamma^2=0.005$ and $a^2=1$.

For the \gls{DETM}, we first fit $300$-dimensional word embeddings using skip-gram
\citep{mikolov2013distributed}.\footnote{More advanced methods can be used to learn word embeddings. 
We used skip-gram for simplicity and found it leads to good performance.}
We apply the algorithm in Section\nobreakspace \ref {subsec:inference} using a batch size of $200$
documents for all datasets except for \textsc{acl}, for which we used $100$.
To parameterize the variational distribution,
we use a fully connected feed-forward inference network for the topic proportions
$\theta_d$. The network has ReLU activations and $2$ layers of $800$ hidden units each. 
We set the mean and log-variance for $\theta_d$ as linear maps of the output. We applied a small 
dropout rate of $0.1$ to the output of this network before using it to compute the mean and the log-variance. 
For the latent means
$\eta_{1:T}$, each bag-of-word representation $\widetilde{\bw}_t$ is first linearly
mapped to a low-dimensional space of dimensionality $400$. This conforms the input of
an \gls{LSTM} that has $4$ layers of $400$ hidden units
each. The \gls{LSTM} output is then concatenated with the previous latent mean
$\eta_{t-1}$, and the result is linearly mapped to a $K$-dimensional space to get the mean
and log-variance for $\eta_t$.
We apply a weight decay of $1.2\cdot10^{-6}$ on all network parameters. 
We run Algorithm\nobreakspace \ref {alg:etm} for a maximum of $1000$ epochs on \textsc{science} and \textsc{acl} 
and for $400$ epochs on the \textsc{un} dataset; the stopping criterion is based on the
held-out log-likelihood on the validation set. The learning rate is set to $0.001$ for 
the \textsc{un} and \textsc{science} datasets and to $0.0008$ on the \textsc{acl} corpus. 
We fixed the learning rate throughout training. We clip the norm of the gradients of the \gls{ELBO} to $2.0$ to stabilize training. 

We fit \gls{DLDA} using the published code of \citet{blei2006dynamic}.\footnote{See \url{https://github.com/blei-lab/dtm}.}
To fit \gls{DLDA}, \citet{blei2006dynamic} derived a bound of the \gls{ELBO} to enable
a coordinate-ascent inference algorithm that also uses Kalman filtering and smoothing as a 
subroutine. Besides loosening the variational bound on the log-marginal likelihood of the data,
this algorithm presents scalability issues both in terms of the number of topics and in terms of the vocabulary size. (See Table\nobreakspace \ref {tab:quantitative} for a comparison of the runtime across methods.)
To fit \gls{DLDA}, we follow \citet{blei2006dynamic} and initialize the algorithm with \gls{LDA}.
In particular, we run $25$ epochs of $\gls{LDA}$ followed by $100$ epochs of \gls{DLDA}.

We also fit \gls{DLDA}-\textsc{rep} to overcome the scalability issues of \gls{DLDA} by leveraging recent advances in variational inference.
We use stochastic optimization based on reparameterization gradients and we draw 
batches of $1{,}000$ documents at each iteration. We collapse the discrete latent topic 
indicators $z_{dn}$ to enable the reparameterization gradients, and we use a fully factorized 
Gaussian approximation for the rest of the latent variables, except for $\eta_{1:T}$, for which we use a
full-covariance Gaussian for each of its dimensions. We run $5$ epochs of \gls{LDA} to initialize \gls{DLDA}-\textsc{rep} and then run $120$ epochs of the \gls{DLDA}-\textsc{rep} inference algorithm. For \gls{DLDA}-\textsc{rep}, we use RMSProp \citep{Tieleman2012} to set the step size,
setting the learning rate to $0.05$ for the mean parameters and to $0.005$ for the
variance parameters.

\parhead{Quantitative results.}
We compare the \gls{DETM}, \gls{DLDA}, and \gls{DLDA}-\textsc{rep} according to two metrics: perplexity on a
document completion task and topic quality. The perplexity is obtained by computing the
probability of each word in the second half of a test document, conditioned on the
first half \citep{rosenzvi2004author,wallach2009evaluation}. To obtain the topic quality,
we combine two metrics. The first metric is topic coherence; it provides a quantitative
measure of the interpretability of a topic \citep{mimno2011optimizing}. We obtain the
topic coherence by taking the average pointwise mutual information of two words drawn randomly
from the same document \citep{lau2014machine}; this requires to approximate word
probabilities with empirical counts. The second metric is topic diversity; it is the
percentage of unique words in the top $25$ words of all topics \citep{dieng2019topic}.
Diversity close to $0$ indicates redundant topics. We obtain both topic coherence and topic
diversity by averaging over time. Finally, topic quality is defined as
the product between topic coherence and diversity \citep{dieng2019topic}.~\looseness=-1

Table\nobreakspace \ref {tab:quantitative} shows that the \gls{DETM} outperforms both \gls{DLDA} and \gls{DLDA}-\textsc{rep} 
according to both perplexity and topic quality on almost all datasets. In particular, the \gls{DETM} finds more diverse and coherent topics. 
We posit this is due to its use of embeddings. 

\parhead{Qualitative results.}
The \gls{DETM} finds that the topics' evolution over time are in agreement with historical
events. As an example, Figure\nobreakspace \ref {fig:topic_evolution_climate_change} shows the trajectory of 
a topic on climate change (a topic that \gls{DLDA}-\textsc{rep} did not discover).
In the 1990s, protecting the ozone layer was the primary
concern; more recently the topic has shifted towards global warming and reducing 
the greenhouse gas emissions. Some events on
climate change, such as the Kyoto protocol (1997) or the Paris convention (2016),
are also reflected in the topic's evolution.

We now examine the evolution of the probability of individual words. 
Figure\nobreakspace \ref {fig:word_evolution} shows these probabilities for a variety
of words and topics. For example, the probability of the word ``Vietnam'' in a topic on Southeast
Asia decays after the end of the war in 1975. In a topic about nuclear weapons, the
concern about the arms ``race'' between the USA and the Soviet Union eventually decays,
and ``Iran'' becomes more relevant in recent years. Similarly, words like ``equality''
and ``gender'' become more important in recent years within a topic about human rights.
Note that the names of the topics are subjective; we assigned the names inspired by the top
words in each topic (the words in Figure\nobreakspace \ref {fig:word_evolution} are not necessarily the most likely words within
each topic). One example is the topic on climate change, whose top words are shown in Figure\nobreakspace \ref {fig:topic_evolution_climate_change}. Another example is the topic on human rights, which
exhibits the words ``human'' and ``rights'' consistently at the top across all time steps.

\subsection{Conclusion}\label{sec:conclusion}

We developed the \gls{DETM}, a probabilistic model of documents that combines word embeddings and dynamic latent Dirichlet allocation (\gls{DLDA}). 
The \gls{DETM} models each word with a categorical distribution parameterized by the dot product between the embedding of the word and 
an embedding representation of its assigned topic. Each topic embedding is a time-varying vector in the embedding
space of words. Using a random walk prior over these topic embeddings, the \gls{DETM} uncovers smooth topic trajectories. 
We applied the \gls{DETM} to analyze three different corpora and found that the \gls{DETM} outperforms \gls{DLDA} both in terms of 
predictive performance and topic quality while requiring significantly less time to fit.

\chapter{Learning via Reweighted Expectation Maximization}
\label{chap:rem}

The models described in Chapter\nobreakspace \ref {chap:dpgm} were fitted using \gls{AVI}. 
\gls{AVI} scales learning by using recognition networks to define the variational family. 
It then maximizes the \gls{ELBO}, a lower bound of the log marginal likelihood of the data. 
Because the \gls{ELBO} is often intractable, \gls{AVI} uses Monte Carlo to approximate it. 
Monte Carlo estimates of the \gls{ELBO} are biased and lead to a loose bound of the log marginal likelihood.  
To address this, several other learning algorithms have been proposed that maximize 
a tighter lower bound than the \gls{ELBO} (e.g.~\cite{bornschein2014reweighted, burda2015importance}.) 

In this chapter, we develop an algorithm for fitting \gls{DPGM}s called \gls{REM}.  
\gls{REM} optimizes an asymptotically unbiased approximation of the log marginal likelihood of the data. 
This procedure involves learning a proposal distribution over the latent variables. 
We propose to leverage moment matching to learn expressive proposals. 
Because \gls{REM} optimizes a better approximation to the log marginal likelihood of the data, 
it generalizes better to unseen data than approaches such as the \gls{VAE}. 

\section{Rethinking \acrshort{ELBO} Maximization for Fitting \acrshort{DPGM}s} 
For simplicity, we focus on the simplest \gls{DPGM}. We consider a set of $N$ i.i.d datapoints $\bx_1, \dots, \bx_N$. 
We posit each observation $\bx_i$ is drawn by first sampling a latent variable $\bz_i$ from 
some fixed prior $p(\bz)$ and then sampling $\bx_i$ from $p_{\theta}(\bx_i \g \bz_i)$---the 
conditional distribution of $\bx_i$ given $\bz_i$. We define the conditional $p_{\theta}(\bx_i \g \bz_i)$ 
using a deep neural network with parameters $\theta$. Our goal is to learn the parameters $\theta$ and perform 
posterior inference over the latent variables. 

One way to achieve this goal is to use \gls{VI} and maximize the \gls{ELBO},
\begin{align}\label{eq:elbo}
	\gls{ELBO} 
	&=  \mathbb{E}_{q_{\phi}(\bz)}\left[ \log p_{\theta}(\bx , \bz) - \log q_{\phi}(\bz) \right].
\end{align} 
This is the approach of the \gls{VAE}~\citep{kingma2013auto, rezende2014stochastic}, 
which maximizes the \gls{ELBO} with respect to both $\phi$ and $\theta$. 
To better understand what this maximization procedure corresponds to, 
consider the expression of the log marginal likelihood of the data in terms of the \gls{ELBO}
\begin{align}\label{eq:log_marginal}
	\log p_{\theta}(\bx) 
		&= \mathbb{E}_{q_{\phi}(\bz)}\left[ \log p_{\theta}(\bx , \bz) - \log q_{\phi}(\bz) \right] + \acrshort{KL}\left(q_{\phi}(\bz) \vert\vert p_{\theta}(\bz \g \bx)\right). 
\end{align}
The log marginal likelihood $\log p_{\theta}(\bx)$ does not depend on the variational parameters $\phi$. 
Therefore performing posterior inference by minimizing the \gls{KL} term $\acrshort{KL}\left(q_{\phi}(\bz) \vert\vert p_{\theta}(\bz \g \bx)\right)$ 
is equivalent to maximizing the \gls{ELBO}, for a fixed $\theta$. However $\log p_{\theta}(\bx)$ depends 
on the parameters $\theta$, which makes the \gls{KL} minimization over 
a ``moving target"---the true posterior $p_{\theta}(\bz \g \bx)$ changes with $\theta$. 
As a result, there is possibility of running into a bad local optimum in which the \gls{KL} 
is rendered small but the neural network parameterizing the model is useless. 

In what follows, we first review \gls{EM} and propose an algorithm that leverages \gls{EM} to fit the model parameters $\theta$. 
Posterior inference can be done, once the model is fit, by minimizing \gls{KL} using amortized \gls{VI}. 

\section{Expectation Maximization} 
\Gls{EM} was first introduced in the statistics literature, where it was used to solve problems involving 
missing data~\citep{dempster1977maximum}. One classic application of \gls{EM} is to fit mixtures of 
Gaussians, where the cluster assignments are considered \textit{unobserved} data~\citep{murphy2012machine}. 
Another use of \gls{EM} is for probabilistic PCA~\citep{tipping1999probabilistic}. 
\gls{EM} is a maximum likelihood iterative optimization technique that directly targets the log marginal likelihood 
and served as the departure point for the development of variational inference methods. 

The \gls{EM} objective is the log marginal likelihood of the data in Eq.\nobreakspace \ref {eq:log_marginal}.
\gls{EM} alternates between an \textit{E-step}, which sets the \gls{KL} term in Eq.\nobreakspace \ref {eq:log_marginal} to 
zero, and an \textit{M-step}, which fits the model parameters $\theta$ by maximizing \gls{ELBO}  
using the proposal learned in the E-step. Note that after the E-step, the objective in Eq.\nobreakspace \ref {eq:log_marginal} 
says the log marginal is exactly equal to the \gls{ELBO} which is a tractable objective for fitting the 
model parameters. \gls{EM} alternates these two steps until convergence to an approximate 
maximum likelihood solution for $p_{\theta}(\bx)$.

Contrast this with \gls{VI}. The true objective for \gls{VI} is the 
\gls{KL} term in Eq.\nobreakspace \ref {eq:log_marginal}, $\gls{KL}\left(q_{\phi}(\bz) \vert\vert p_{\theta}(\bz \g \bx)\right)$, 
which is intractable. The argument in \gls{VI} is that minimizing this \gls{KL} is equivalent to 
maximizing the \gls{ELBO}, the first term in Eq.\nobreakspace \ref {eq:log_marginal}. This argument only 
holds when the log marginal likelihood $\log p_{\theta}(\bx)$ has no free parameters, in 
which case it is called the \textit{model evidence}. 
Importantly, \gls{VI} does not necessarily maximize $\log p_{\theta}(\bx)$ because it 
chooses approximate posteriors $q_{\phi}(\bz)$ that may be far from the exact conditional posterior.  

In contrast \gls{EM} effectively maximizes $\log p_{\theta}(\bx)$ after each iteration. 
Consider given $\theta_t$, the state of the model parameters after the $t^{th}$ iteration of \gls{EM}. 
\gls{EM} learns $\theta_{t+1}$ through two steps, which we briefly review:
\begin{align}
	\text{E-step:} &  \text{ set }  q_{\phi}(\bz) = p_{\theta_t}(\bz \g \bx)\label{eq:e_step}\\
	\text{M-step:} & \text{ define } \theta_{t+1}  = \argmax_{\theta} \mathcal{L}(\theta) \nonumber\\
	&= \argmax_{\theta} \mathbb{E}_{q_{\phi}(\bz)}\left[ \log p_{\theta}(\bx , \bz) - \log q_{\phi}(\bz) \right] \nonumber\\
	&= \argmax_{\theta} \mathbb{E}_{p_{\theta_t}(\bz \g \bx)}\left[ \log p_{\theta}(\bx , \bz) - \log p_{\theta_t}(\bz \g \bx) \right] \nonumber\\
	&=  \argmax_{\theta} \mathbb{E}_{p_{\theta_t}(\bz \g \bx)}\left[ \log p_{\theta}(\bx , \bz)\right]
	\label{eq:m_step}
\end{align}
The value of the log marginal likelihood for $\theta_{t+1}$ is greater than for  $\theta_t$. To see this, write
\begin{align*}
	\log p_{\theta_t}(\bx) &= \mathcal{L}(\theta_t)  + \gls{KL}\left(q_{\phi}(\bz) \vert\vert p_{\theta_t}(\bz \g \bx)\right) = \mathcal{L}(\theta_t) \\
	&\leq \mathcal{L}(\theta_{t+1}) 
	\leq \mathcal{L}(\theta_{t+1}) + \gls{KL}\left(q_{\phi}(\bz) \vert\vert p_{\theta_{t+1}}(\bz \g \bx)\right) \\
	&= \log p_{\theta_{t+1}}(\bx)
\end{align*}
where the second equality is due to the E-step, the first inequality is due to the M-step, and 
the second inequality is due to the nonnegativity of \gls{KL}. 

We next propose an algorithm that leverages \gls{EM} to fit the model parameters $\theta$. 

\section{Reweighted Expectation Maximization} 
\label{sec:method}
We develop \gls{REM}, an algorithm that leverages \gls{EM} to fit the model parameters $\theta$. 
Assume given $\theta_t$ from the previous iteration of \gls{EM}. 
We want to find the next settings of the parameters  $\theta_{t+1}$ that maximize the objective in the M-step in Eq.\nobreakspace \ref {eq:m_step},   
\begin{align}\label{eq:factorization_2}
	\mathcal{L}(\theta) &=\sum_{i=1}^{N} \mathbb{E}_{p_{\theta_t}(\bz_i \g \bx_i)}\left[\log p_{\theta}(\bx_i, \bz_i)\right]\\
	&= \sum_{i=1}^{N} \int_{}^{} \frac{p_{\theta_t}(\bz_i , \bx_i)}{p_{\theta_t}(\bx_i)} \log p_{\theta}(\bx_i, \bz_i) \text{ } d\bz_i
	.
\end{align}
This objective is intractable because it involves the marginal 
$p_{\theta_t}(\bx_i)$\footnote{Although the marginal here does not depend on $\theta$, it cannot be ignored because it depends on the $i^{th}$ datapoint. Therefore it cannot be pulled outside the summation. }. However we can make it tractable using self-normalized importance sampling~\citep{Owen2013},
\begin{align}\label{eq:rem_new}
	\mathcal{L}(\theta) 
	&= \sum_{i=1}^{N} \mathbb{E}_{r_{\eta_t}(\bz_i \g \bx_i)}\left[
		\frac{\bw(\bx_i, \bz_i; \theta_t, \eta_t) \log p_{\theta}(\bx_i, \bz_i)}{\mathbb{E}_{r_{\eta_t}(\bz_i \g \bx_i)}\left(\bw(\bx_i, \bz_i; \theta_t, \eta_t)\right)} 
	\right]
	.
\end{align}
where $\bw(\bx_i, \bz_i; \theta_t, \eta_t) = \frac{p_{\theta_t}(\bz_i , \bx_i)}{r_{\eta_t}(\bz_i \g \bx_i)}$. 
Here $r_{\eta_t}(\bz_i \g \bx_i)$ is a proposal distribution. 
Its parameters $\eta_t$ were fitted in the previous iteration (the $t^{th}$ iteration.) 
We now approximate the expectations in Eq.\nobreakspace \ref {eq:rem_new} using 
Monte Carlo by drawing $K$ samples $\bz_i^{(1)}, \dots, \bz_i^{(K)}$ from the proposal, 
\begin{align}\label{eq:rem_samples}
\balpha_{it}^{k} &= \frac{\bw(\bx_i, \bz_i^{(k)}; \theta_t, \eta_t)}{\sum_{k=1}^{K} \bw(\bx_i, \bz_i^{(k)}; \theta_t, \eta_t)} \nonumber \\ 
	\mathcal{L}(\theta) 
	&= \sum_{i=1}^{N} \sum_{k=1}^{K} \balpha_{it}^{k}\cdot \log p_{\theta}(\bx_i, \bz_i^{(k)})
\end{align} 
Note the approximation in Eq.\nobreakspace \ref {eq:rem_samples} is biased but asymptotically unbiased. 
More specifically, the approximation improves as the number of particles $K$ increases. 

We use gradient-based learning which requires to compute the gradient of $\mathcal{L}(\theta)$ 
with respect to the model parameters $\theta$, this is 
\begin{align}\label{eq:grad_theta}
	\nabla_{\theta}\mathcal{L}(\theta) 
	&= \sum_{i=1}^{N} \sum_{k=1}^{K} \balpha_{it}^{k} \cdot \nabla_{\theta}\log p_{\theta}(\bx_i, \bz_i^{(k)}). 
\end{align}
We now describe how to learn expressive proposals by leveraging moment matching.

\begin{algorithm}[tb]
  \caption{Learning with reweighted expectation maximization (\acrshort{REM} (v1))}
  \label{alg:rem}
  \begin{algorithmic}
     \STATE {\bfseries Input:} Data $\bx$
     \STATE Initialize model and proposal parameters $\theta, \eta$
     \FOR{\emph{iteration} $t=1,2,\ldots$}
      \STATE Draw minibatch of observations $\{\bx_n\}_{n=1}^{B}$
      \FOR{\emph{observation} $n=1,2,\ldots, B$}
        \STATE Draw $\bz^{(1)}_n, \dots, \bz^{(K)}_n \sim r_{\eta_t}(\bz_n^{(k)} \g \bx_n)$
        \STATE Compute importance weights $\bw^{(k)} = \frac{p_{\theta_t}(\bz_n^{(k)} , \bx_n)}{r_{\eta_t}(\bz_n^{(k)} \g \bx_n)}$
        \STATE Compute $\bmu_{nt}$ and $\bSigma_{nt}$  using Eq.\nobreakspace \ref {eq:moment_matching_3} and Eq.\nobreakspace \ref {eq:moment_matching_3_bis} 
        \STATE Set proposal $s(\bz_n^{(t)}) = \mathcal{N}(\bmu_{nt}, \bSigma_{nt})$
      \ENDFOR
      \STATE Compute $\nabla_{\eta}\mathcal{L}(\eta)$ as:
      \STATE $\nabla_{\eta}\mathcal{L}(\eta) = \frac{1}{\vert B\vert}\sum_{n \in B}^{} \sum_{k=1}^{K} \frac{\bv^{(k)}}{\sum_{k=1}^{K} \bv^{(k)}} \nabla_{\eta}\log r_{\eta}(\bz_n^{(k)} \g \bx_n)$ 
      \STATE Update $\eta$ using Adam
      \STATE Compute $\nabla_{\theta}\mathcal{L}(\btheta)$ as 
       \STATE $\nabla_{\theta}\mathcal{L}(\btheta) = \frac{1}{\vert B\vert}\sum_{n \in B}^{} \sum_{k=1}^{K} \frac{\bw^{(k)}}{\sum_{k=1}^{K} \bw^{(k)}} \nabla_{\theta}\log p_{\theta}(\bx_n, \bz^{(k)}_n)$ 
       \STATE Update $\theta$ using Adam
     \ENDFOR
  \end{algorithmic}
\end{algorithm}

\subsection{Learning Expressive Proposals via Moment Matching}

Denote by $\eta_t$ the proposal parameters at the previous iteration. 
We learn $\eta_{t+1}$ by targeting the true posterior $p_{\theta_t}(\bz \g \bx)$, 
\begin{align}\label{eq:inclusive_kl}
	\eta_{t+1} &= \argmin_{\eta} \mathcal{L}_{\gls{REM}}(\eta) = \gls{KL}(p_{\theta_t}(\bz \g \bx) \vert\vert r_{\eta}(\bz \g \bx))
	.
\end{align}
Unlike the \gls{IWAE}, the proposal here targets the true posterior using a well defined 
objective---the inclusive \gls{KL} divergence. The inclusive \gls{KL} induces overdispersed 
proposals which are beneficial in importance sampling~\citep{minka2005divergence}. 

The objective in Eq.\nobreakspace \ref {eq:inclusive_kl} is still intractable as it involves the true posterior $p_{\theta_t}(\bz\g \bx)$, 
\begin{align}\label{eq:proposal_loss}
	\mathcal{L}_{\gls{REM}}(\eta) &= -\sum_{i=1}^{N} \mathbb{E}_{p_{\theta_t}(\bz_i \g \bx_i)}\left[\log r_{\eta}(\bz_i \g \bx_i) \right] + \text{const.},
\end{align}
where $\text{const.}$ is a constant with respect to $\eta$ that we can ignore. 
We use the same approach as for fitting the model parameters $\theta$. That is, we write
\begin{align}\label{eq:rem}
	\mathcal{L}_{\gls{REM}}(\eta) 
	&= \!-\!\sum_{i=1}^{N} \mathbb{E}_{s(\bz_i)}\!\left[
		\frac{\bv(\bx_i, \bz_i; \theta_t, \eta_t) \log r_{\eta}(\bz_i \g \bx_i)}{\mathbb{E}_{s(\bz_i)}\left(\bv\left(\bx_i, \bz_i; \theta_t, \eta_t\right)\right)} \!
	\right]\!.
\end{align}
where $\bv(\bx_i, \bz_i; \theta_t, \eta_t) = \frac{p_{\theta_t}(\bz_i , \bx_i)}{s(\bz_i)}$. 
Here $s(\bz_i)$ is a hyperproposal that has no free parameters. (We will describe it shortly.) 
The hyperobjective in Eq.\nobreakspace \ref {eq:rem} is still intractable due to the expectations. 
We approximate it using Monte Carlo by drawing $K$ samples $\bz_i^{(1)}, \dots, \bz_i^{(K)}$ from $s(\bz_i)$. Then 
\begin{align}\label{eq:rem_2}
	\bbeta_{it}^{k} &= \frac{\bv\left(\bx_i, \bz_i^{(k)}; \theta_t, \eta_t\right)}{\sum_{k'=1}^{K} \bv\left(\bx_i, \bz_i^{(k')}; \theta_t, \eta_t\right)} \nonumber \\
	\mathcal{L}_{\gls{REM}}(\eta) &= -\sum_{i=1}^{N} \sum_{k=1}^{K} \bbeta_{it}^{k}  \cdot \log r_{\eta}(\bz_i^{(k)} \g \bx_i), 
\end{align}
We choose the proposal $s(\bz_i)$ to be a full Gaussian whose parameters are found by 
matching the moments of the true posterior $p_{\theta_t}(\bz_i \g \bx_i)$. 
More specifically, $s(\bz_i) = \mathcal{N}(\bmu_{it}, \Sigma_{it})$ where 
\begin{align}\label{eq:moment_matching}
	\bmu_{it} &= \mathbb{E}_{p_{\theta_t}(\bz_i \g \bx_i)}[\bz_i] \nonumber \\
        \Sigma_{it} &=  \mathbb{E}_{p_{\theta_t}(\bz_i \g \bx_i)}\left[\left(\bz_i - \bmu_i^{(t)}\right)\left(\bz_i - \bmu_i^{(t)}\right)^\top\right]
	.
\end{align}

The expressions for the mean and covariance matrix are still intractable. 
We estimate them using self-normalized importance sampling, with 
proposal $r_{\eta_t}(\bz_i\g \bx_i)$, and Monte Carlo. We first write
\begin{align}\label{eq:moment_matching_2}
	\bmu_{it} &= \mathbb{E}_{r_{\eta_t}(\bz_i \g \bx_i)}\left(
		\frac{\bw(\bx_i, \bz_i; \theta_t, \eta_t)}{\mathbb{E}_{r_{\eta_t}(\bz_i \g \bx_i)}\left(\bw(\bx_i, \bz_i; \theta_t, \eta_t)\right)} \bz_i
	\right),
\end{align}
(the covariance $\bSigma_{it}$ is analogous), and then estimate the expectations using Monte Carlo,
\begin{align}\label{eq:moment_matching_3}
	\bmu_{it} &\approx \sum_{k=1}^{K} \balpha_{it}^{k}\cdot \bz_i^{(k)} \nonumber \\
	\bSigma_{it} &\approx \sum_{k=1}^{K} \balpha_{it}^{k} \left[(\bz_i^{(k)} - \bmu_{it})(\bz_i^{(k)} - \bmu_{it} )^\top\right]
	.
\end{align}
Note Eq.\nobreakspace \ref {eq:moment_matching_3} imposes the implicit constraint that the number of 
particles $K$ be greater than the square of the dimensionality of the latents for the 
covariance matrix $\bSigma_{it}$ to have full rank. We lift this constraint by adding a 
constant $\epsilon$ to the diagonal of $\bSigma_{it}$ and setting 
\begin{align}\label{eq:moment_matching_3_bis}
	\bSigma_{it} \approx \sum_{k=1}^{K} (\bz_i^{(k)} - \bmu_{it})(\bz_i^{(k)} - \bmu_{it} )^\top
	.
\end{align}

Algorithm\nobreakspace \ref {alg:rem} summarizes the procedure for fitting deep generative models 
with \gls{REM} where $\bv^{(k)}$ is computed the same way as $\bv(\bx_i, \bz_i; \theta_t, \eta_t)$. 
We call this algorithm \gls{REM} (v1).

\begin{algorithm}[tb]
  \caption{Learning with reweighted expectation maximization (\acrshort{REM} (v2))}
  \label{alg:rem_2}
  \begin{algorithmic}
     \STATE {\bfseries Input:} Data $\bx$
     \STATE Initialize model and proposal parameters $\theta, \eta$
     \FOR{\emph{iteration} $t=1,2,\ldots$}
      \STATE Draw minibatch of observations $\{\bx_n\}_{n=1}^{B}$ 
      \FOR{\emph{observation} $n=1,2,\ldots, B$}
        \STATE Draw $\bz^{(1)}_n, \dots, \bz^{(K)}_n \sim r_{\eta_t}(\bz_n^{(k)} \g \bx_n)$        \STATE Compute importance weights $\bw^{(k)} = \frac{p_{\theta_t}(\bz_n^{(k)} , \bx_n)}{r_{\eta_t}(\bz_n^{(k)} \g \bx_n)}$
        \STATE Compute $\bmu_{nt} = \sum_{k=1}^{K} \frac{\bw^{(k)}}{\sum_{k=1}^{K} \bw^{(k)}} \bz^{(k)}_n$ and $\bSigma_{nt} = \sum_{k=1}^{K} \frac{\bw^{(k)}}{\sum_{k=1}^{K} \bw^{(k)}}  (\bz^{(k)}_n - \bmu_{nt})(\bz^{(k)}_n  - \bmu_n)^\top$
        \STATE Set proposal $s(\bz_n^{(t)}) = \mathcal{N}(\bmu_{nt}, \bSigma_{nt})$
      \ENDFOR
      \STATE Compute $\nabla_{\eta}\mathcal{L}(\eta) = \frac{1}{\vert B\vert}\sum_{n \in B}^{} \sum_{k=1}^{K} \frac{\bv^{(k)}}{\sum_{k=1}^{K} \bv^{(k)}} \nabla_{\eta}\log r_{\eta}(\bz_n^{(k)} \g \bx_n)$ and update $\eta$ using Adam
      \STATE Compute $\nabla_{\theta}\mathcal{L}(\btheta)  = \frac{1}{\vert B\vert}\sum_{n \in B}^{} \sum_{k=1}^{K} \frac{\bv^{(k)}}{\sum_{k=1}^{K} \bv^{(k)}} \nabla_{\theta}\log p_{\theta}(\bx_n, \bz^{(k)}_n)$ and update $\theta$ using Adam
     \ENDFOR
  \end{algorithmic}
\end{algorithm}

We can also consider using the rich moment matched distribution $s(\bz)$ to directly update the generative model. 
This changes the objective $\mathcal{L}(\theta)$ in Eq.\nobreakspace \ref {eq:rem_samples} to 
\begin{align}\label{eq:rem_samples_new}
	\mathcal{L}(\theta) 
	&= \sum_{i=1}^{N} \sum_{k=1}^{K} \bbeta_{it}^{k}\cdot \log p_{\theta}(\bx_i, \bz_i^{(k)})
\end{align}
where $\bz_i^{(1)}, \dots, \bz_i^{(K)} \sim s(\bz_i)$ and $\bbeta_{it}^{k}$ is as defined in Eq.\nobreakspace \ref {eq:rem_2}. 
We let the recognition network $r_{\eta_t}(\bz_i \g \bx_i)$ be learned the same way as done for \gls{REM} (v1). 
Algorithm\nobreakspace \ref {alg:rem_2} summarizes the procedure for fitting \glspl{DPGM} with \gls{REM} (v2).

\begin{table*}[t]
	\centering
	\small
	\captionof{table}[Methodological differences between \gls{REM}, the \gls{VAE}, the \gls{IWAE}, and \gls{RWS}]{Comparing \gls{REM} against the \gls{VAE}, the \gls{IWAE}, and \gls{RWS}. 
	\gls{REM} uses a rich distribution $s(\bz)$ found by moment matching to learn the 
	generative model and/or the recognition network $r_{\eta}(\bz \g \bx)$.}
	\begin{tabular}{ccccc}
	\toprule
	 Method & Objective & Proposal & Hyperobjective & Hyperproposal\\
	 \hline
	 \gls{VAE} & \acrshort{VI} & $r_{\eta}(\bz \g \bx)$ & $\text{KL}(r_{\eta}(\bz \g \bx) \vert\vert p_{\theta}(\bz \g \bx))$ & $r_{\eta}(\bz \g \bx)$\\
	 \gls{IWAE} & \acrshort{EM} & $r_{\eta}(\bz \g \bx)$ & $\mathcal{L}_{\gls{IWAE}}(\eta)$ & $r_{\eta}(\bz \g \bx)$\\
	 \acrshort{RWS} &  \acrshort{EM} & $r_{\eta}(\bz \g \bx)$ & $\text{KL}(p_{\theta}(\bz \g \bx) \vert\vert r_{\eta}(\bz \g \bx))$ &  $r_{\eta}(\bz \g \bx)$\\
	 \gls{REM}(v1) & \acrshort{EM} & $r_{\eta}(\bz \g \bx)$ & $\text{KL}(p_{\theta}(\bz \g \bx) \vert\vert r_{\eta}(\bz \g \bx))$ & $s(\bz)$\\
	 \gls{REM} (v2) & \acrshort{EM} & $s(\bz)$ & $\text{KL}(p_{\theta}(\bz \g \bx) \vert\vert r_{\eta}(\bz \g \bx))$ &  $r_{\eta}(\bz \g \bx)$\\
	\bottomrule
	\end{tabular}
	\label{tab:approaches}
\end{table*}

\section{Connections} 
\gls{REM} generalizes and connects 
algorithms that rely on importance sampling to optimize a tighter approximation of the log 
marginal likelihood (e.g. \gls{IWAE}~\citep{burda2015importance} and \gls{RWS}~\citep{bornschein2014reweighted}.)
We discuss this next. 

\subsection{Importance-Weighted Auto-Encoders} 
\gls{IWAE} was introduced to learn better generative models~\citep{burda2015importance}. 
It relies on importance sampling to optimize both the model parameters and the recognition network. 
\gls{IWAE} maximizes
\begin{align}
	\mathcal{L}_{\gls{IWAE}}(\theta, \eta) 
	&= \sum_{i=1}^{N} \log \left(\frac{1}{K}\sum_{k=1}^{K} \frac{p_{\theta}(\bx_i, \bz_i^{(k)})}{r_{\eta}(\bz_i^{(k)} \g \bx_i)}\right)
\end{align}
where $r_{\eta}(\bz_i^{(k)} \g \bx_i)$ is an importance sampling proposal 
and $\bz_1^{(k)}, \dots, \bz_i^{(K)} \sim r_{\eta}(\bz_i^{(k)} \g \bx_i)$. 
This objective is simply a biased Monte Carlo approximation of the 
log marginal likelihood using importance sampling. To confirm this, write 
\begin{align}
	\log p_{\theta}(\bx_{1:N})
	&= \sum_{i=1}^{N} \log p_{\theta}(\bx_i)\\
	&= \sum_{i=1}^{N} \log \int_{}^{} \frac{p_{\theta}(\bx_i, \bz_i) \cdot r_{\eta}(\bz_i \g \bx_i)}{r_{\eta}(\bz_i \g \bx_i)} \text{ } d\bz_i \\
	&= \sum_{i=1}^{N} \log\mathbb{E}_{r_{\eta}(\bz_i \g \bx_i)}  \left(\frac{p_{\theta}(\bx_i, \bz_i)}{r_{\eta}(\bz_i \g \bx_i)} \right) \label{eq:iwaeeq}\\
	&\approx \sum_{i=1}^{N} \log \left(\frac{1}{K} \sum_{k=1}^{K} \frac{p_{\theta}(\bx_i, \bz^{(k)}_i)}{r_{\eta}(\bz^{(k)}_i \g \bx_i)} \right) \label{eq:biasedmarginal}
\end{align}
where $\bz_1^{(k)}, \dots, \bz_i^{(K)} \sim r_{\eta}(\bz_i^{(k)} \g \bx_i)$. 
Note the \gls{VAE} lower bounds Eq.\nobreakspace \ref {eq:iwaeeq} using concavity of logarithm, which leads to the \gls{ELBO} objective. 

The \gls{IWAE} objective is shown to be a tighter approximation to the log marginal likelihood of the data than 
the \gls{ELBO}~\citep{burda2015importance}; this tightness is determined by the number of particles $K$ used 
for importance sampling. 

Consider taking gradients of $\mathcal{L}_{\gls{IWAE}}(\theta, \eta)$ with respect to the model parameters $\theta$, 
\begin{align}
	\nabla_{\theta}\mathcal{L}_{\gls{IWAE}}(\theta, \eta) 
	&= \sum_{i=1}^{N} \nabla_{\theta} \log \left(\frac{1}{K}\sum_{k=1}^{K} \frac{p_{\theta}(\bx_i, \bz_i^{(k)})}{r_{\eta}(\bz_i^{(k)} \g \bx_i)}\right)\\
	&=  \sum_{i=1}^{N} \frac{\nabla_{\theta}\left(\frac{1}{K}\sum_{k=1}^{K} \frac{p_{\theta}(\bx_i, \bz_i^{(k)})}{r_{\eta}(\bz_i^{(k)} \g \bx_i)}\right)}{\frac{1}{K}\sum_{k=1}^{K} \frac{p_{\theta}(\bx_i, \bz_i^{(k)})}{r_{\eta}(\bz_i^{(k)} \g \bx_i)}}\\
	&=   \sum_{i=1}^{N} \frac{\frac{1}{K}\sum_{k=1}^{K} \frac{p_{\theta}(\bx_i, \bz_i^{(k)})}{r_{\eta}(\bz_i^{(k)} \g \bx_i)} \nabla_{\theta} \log p_{\theta}(\bx_i, \bz_i^{(k)})}{\frac{1}{K}\sum_{k=1}^{K} \frac{p_{\theta}(\bx_i, \bz_i^{(k)})}{r_{\eta}(\bz_i^{(k)} \g \bx_i)}}\\
	&=  \sum_{i=1}^{N} \sum_{k=1}^{K} \balpha_{it}^{k} \cdot \nabla_{\theta}\log p_{\theta}(\bx_i, \bz_i^{(k)}) \label{eq:last_iwae}
\end{align}
where $\balpha_{it}^{k}$ was previously defined in Eq.\nobreakspace \ref {eq:rem_samples}. 
Note Eq.\nobreakspace \ref {eq:last_iwae} is the expression of the \gls{REM} gradient with respect to the model parameters $\theta$ (Eq.\nobreakspace \ref {eq:grad_theta}.)

\gls{IWAE} updates the proposal by taking gradients of $\mathcal{L}_{\gls{IWAE}}(\theta, \eta)$ with respect to $\eta$. 
As pointed out in \citet{le2017auto} this objective does not correspond to minimizing any divergence between 
the \gls{IWAE}'s proposal $r_{\eta}(\bz_i^{(k)} \g \bx_i)$ and the true posterior. 
However $\mathcal{L}_{\gls{IWAE}}(\theta, \eta)$ can be viewed as a divergence between 
an \textit{importance weighted distribution} and the true posterior. We refer the reader to \citet{cremer2017reinterpreting} 
for a detailed exposition. 

To illustrate how \gls{REM} (v1) improves upon the \gls{IWAE}, consider replacing $s(\bz_i)$ in 
the definition of $\bv(\bx_i, \bz_i; \theta_t, \eta_t)$ with $r_{\eta_t}(\bz_i \g \bx_i)$. 
Then taking gradients of Eq.\nobreakspace \ref {eq:rem_2} with respect to $\eta$ reduces to the \gls{IWAE} gradient for 
updating the recognition network $r_{\eta}(\bz_i \g \bx_i)$. Instead of using $r_{\eta_t}(\bz_i \g \bx_i)$, \gls{REM} (v1) 
uses a more expressive distribution found via moment matching to update the recognition network. 
This further has the advantage of decoupling the generative model and the recognition network as they do not use the same objective for learning. 

\subsection{Reweighted Wake-Sleep} 
The \gls{RWS} algorithm extends the \gls{WS} algorithm of ~\citet{hinton1995wake} to importance sampling the same 
way the \gls{IWAE} algorithm extends the \gls{VAE} to importance sampling. It uses the same importance sampling 
approximation of the log marginal likelihood as \gls{IWAE} (Eq.\nobreakspace \ref {eq:biasedmarginal}.) Therefore \gls{RWS} leads to 
the same gradients with respect to the model parameters than \gls{IWAE}. The two approaches differ in how they learn the proposal. 

The \gls{RWS} proposal minimizes the inclusive \gls{KL}, similarly to \gls{REM},
\begin{align}\label{eq:inclusive_kl}
	\eta_{t+1} &= \argmin_{\eta} \mathcal{L}_{\gls{RWS}}(\eta) = \gls{KL}(p_{\theta_t}(\bz \g \bx) \vert\vert r_{\eta}(\bz \g \bx))
	.
\end{align}
However, unlike \gls{REM} which leverages the fact that the inclusive \gls{KL} admits moment matching 
as a solution, \gls{RWS} minimizes an approximation of the \gls{KL},
\begin{align}
	\mathcal{L}_{\gls{RWS}}(\eta) 
	&= -\mathbb{E}_{p_{\theta}(\bz \g \bz)} \left(\log r_{\eta}(\bz \g \bx) \right)+ \text{cst} \\
	&= -\int_{}^{} \frac{p_{\theta}(\bx, \bz)}{\int_{}^{} p_{\theta}(\bx, \bz) \text{ } d\bz} \log r_{\eta}(\bz \g \bx) d\bz + \text{cst} \\
	&= -\mathbb{E}_{r_{\eta}(\bz \g \bx)} \left( \frac{\frac{p_{\theta}(\bx, \bz)}{ r_{\eta}(\bz \g \bx)}}{\mathbb{E}_{r_{\eta}(\bz \g \bx)} \left(\frac{p_{\theta}(\bx, \bz)}{ r_{\eta}(\bz \g \bx)}\right)}  \log r_{\eta}(\bz \g \bx) \right)+ \text{cst} \\
	&\approx -\sum_{k=1}^{K} \bw_k \log r_{\eta}(\bz^{(k)} \g \bx) + \text{cst}
\end{align}
where $\bz^{(1)}, \dots, \bz^{(K)} \sim r_{\eta}(\bz \g \bx)$ and $\bw_k  = \frac{p_{\theta}(\bz^{(k)} , \bx)}{r_{\eta}(\bz^{(k)} \g \bx)}$. 
\gls{RWS} updates its proposal by taking gradients of $\mathcal{L}_{\gls{RWS}}(\eta)$ with respect to $\eta$,
\begin{align}
	\nabla_{\eta} \mathcal{L}_{\gls{RWS}}(\eta) 
	&= -\sum_{k=1}^{K} \bw_k \nabla_{\eta} \log r_{\eta}(\bz^{(k)} \g \bx) 
\end{align}
\gls{REM} improves upon \gls{REM} by using a richer hyperproposal than $r_{\eta}(\bz^{(k)} \g \bx)$ to update 
its proposal. To see this, replace $s(\bz)$ used to compute the gradients of the \gls{REM} 
objective with respect to $\eta$ with $r_{\eta}(\bz^{(k)} \g \bx)$ to recover the \gls{RWS} gradients. 

Table\nobreakspace \ref {tab:approaches} highlights the differences between the \gls{VAE}, the \gls{IWAE}, \gls{RWS}, \gls{REM}(v1), and \gls{REM}(v2).

\section{Empirical Study} 
In this section, we showcase the benefits of using \gls{REM} over the \gls{VAE} and the \gls{IWAE}.  
We assess generalization using predictive log-likelihood on held-out data. 

Note \gls{RWS} requires specific architectures, e.g. NADE~\citep{uria2016neural} or SBN~\citep{saul1996mean}, to achieve good results. 
In our empirical studies we focus on the controlled setting of \citet{burda2015importance}, which uses simple MLPs for density estimation.  
\gls{RWS} achieves significantly worse results than the \gls{VAE} when using MLPs and we don't report those results.  

We consider several benchmark datasets, which we describe next. 

\begin{figure*}[t]
	\centering
	\centerline{\includegraphics[width=1.2\textwidth, height=11cm]{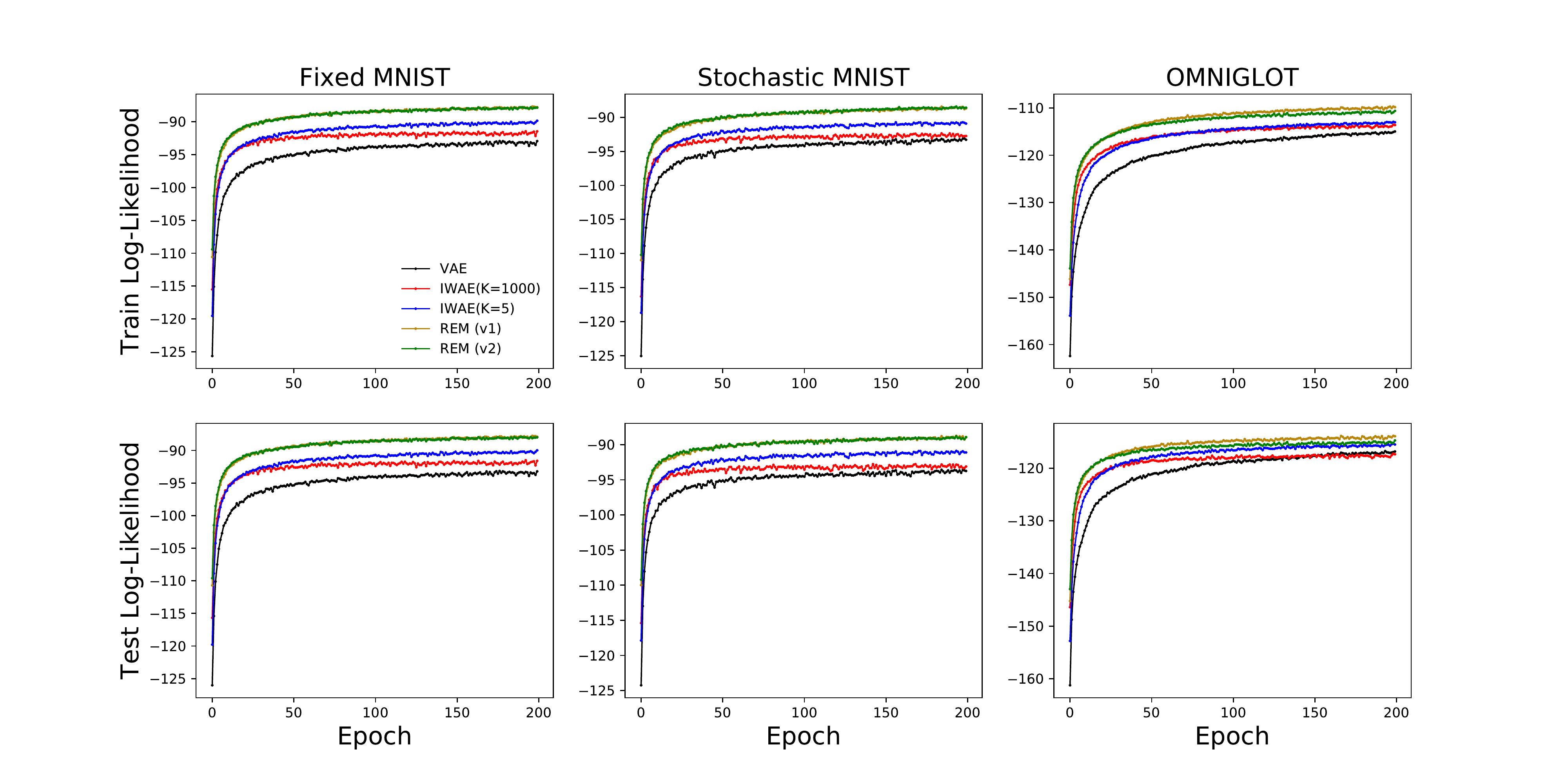}}
	\caption[Generalization performance of \gls{REM}, the \gls{VAE}, and the \gls{IWAE}]{\gls{REM} achieves significantly better performance than the \gls{VAE} and the \gls{IWAE} on three benchmark datasets in terms of log-likelihood (the higher the better).}
	\label{fig:nll}
\end{figure*}

\subsection{Datasets} We evaluated all methods on the \textsc{omniglot} dataset and two versions of \textsc{mnist}. 
The \textsc{omniglot} is a dataset of handwritten characters in a total of $50$ different alphabets~\citep{lake2013one}. 
Each of the characters is a single-channel image with dimension $28 \times 28$. 
There are in total $24{,}345$ images in the training set and $8{,}070$ images in the test set. 
\textsc{mnist} is a dataset of images of handwritten digits introduced by \citet{lecun1998gradient}. 
The first version of \textsc{mnist} we consider is the fixed binarization of the \textsc{mnist} dataset 
used by \citet{larochelle2011neural}. The second version of \textsc{mnist} corresponds to 
random binarization; a random binary sample of digits is newly created during optimization to 
get a minibatch of data. In both cases the images are single-channel and have dimension $28 \times 28$. 
There are $60{,}000$ images in the training set and $10{,}000$ images in the test set. 
All these datasets are available online at \url{https://github.com/yburda/iwae}.

\subsection{Settings} We used the same network architecture for all methods. 
We followed~\citet{burda2015importance} and set the generative model, also called a \textit{decoder}, to 
be a fully connected feed-forward neural network with two layers where each layer has $200$ hidden units. 
We set the recognition network, also called an \textit{encoder}, to be a fully connected feed-forward 
neural network with two layers and $200$ hidden units in each layer. We use two additional linear maps to 
get the mean and the log-variance for the distribution $r_{\eta}(\bz \g \bx)$. 
The actual variance is obtained by exponentiating the log-variance.

We used a minibatch size of $20$ and set the learning rate following the schedule describes 
in \citet{burda2015importance} with an initial learning rate of $10^{-3}$. We use this same learning rate 
schedule for both the learning of the generative model and the recognition network. 
We set the dimension of the latents used as input to the generative model to $20$. 
We set the seed to $2019$ for reproducibility. We set the number of particles $K$ to $1{,}000$ 
for both training and testing. We ran all methods for $200$ epochs. 
We used Amazon EC-2 P3 GPUs for all our experiments. 

\subsection{Results}
We now describe the results in terms of quality of the learned generative model and proposal. 

\parhead{\gls{EM}-based methods learn better generative models.} We assess the 
quality of the fitted generative model for each method using log-likelihood. 
We report log-likelihood on both the training set and the test set. Figure\nobreakspace \ref {fig:nll} illustrates the results. 
The \gls{VAE} performs the worse on all datasets and on both the training and the test set. 
The \gls{IWAE} performs better than the \gls{VAE} as it optimizes a better objective function to train its generative model.  
Finally, both versions of \gls{REM} significantly outperform the \gls{IWAE} on all cases. 
This is evidence of the effectiveness of \gls{EM} as a good alternative for learning deep generative models. 

\parhead{Recognition networks are good proposals.} Here we study the effect of the proposal on the performance of \gls{REM}. 
We report the log-likelihood on both the train and the test set in Table\nobreakspace \ref {tab:rem_vs_rem}.
 As shown in Table\nobreakspace \ref {tab:rem_vs_rem}, using the richer distribution $s(\bz)$ does not always lead 
 to improved performance. These results suggest that recognition networks are good proposals 
 for updating model parameters in deep generative models. 

\begin{table*}[t]
	\centering
	\small
	\captionof{table}[On the effect of the proposal and the hyperproposal in \gls{REM}]{\gls{REM} (v1) outperforms \gls{REM} (v2) on all but one dataset. This suggests that recognition networks are effective proposals for the purpose of learning the generative model.}
	\begin{tabular}{|cc|cc|cc|cc|}
	\cline{1-8}
	 \multicolumn{2}{|c|}{\gls{REM}} & \multicolumn{2}{c|}{Fixed MNIST} & \multicolumn{2}{c|}{Stochastic MNIST} & \multicolumn{2}{c|}{Omniglot}  \\
	 \hline
	 Proposal & Hyperproposal & Train & Test & Train & Test & Train & Test \\
	 \hline
	 $r_{\eta}(\bz \g \bx)$ & $s(\bz)$ & $\textbf{87.77}$ & $\textbf{87.91}$ & $88.68$ & $88.95$ & $\textbf{109.84}$ & $\textbf{113.94}$\\
	 $s(\bz)$ & $r_{\eta}(\bz \g \bx)$ & $87.84$ & $87.99$ & $\textbf{88.58}$ & $\textbf{88.92}$ & $110.63$ & $114.73$ \\
	\hline
	\end{tabular}
	\label{tab:rem_vs_rem}
\end{table*}
\begin{figure*}[t]
	\centering
 	\vspace*{-8pt}
	\centerline{\includegraphics[width=1.2\textwidth, height=10cm]{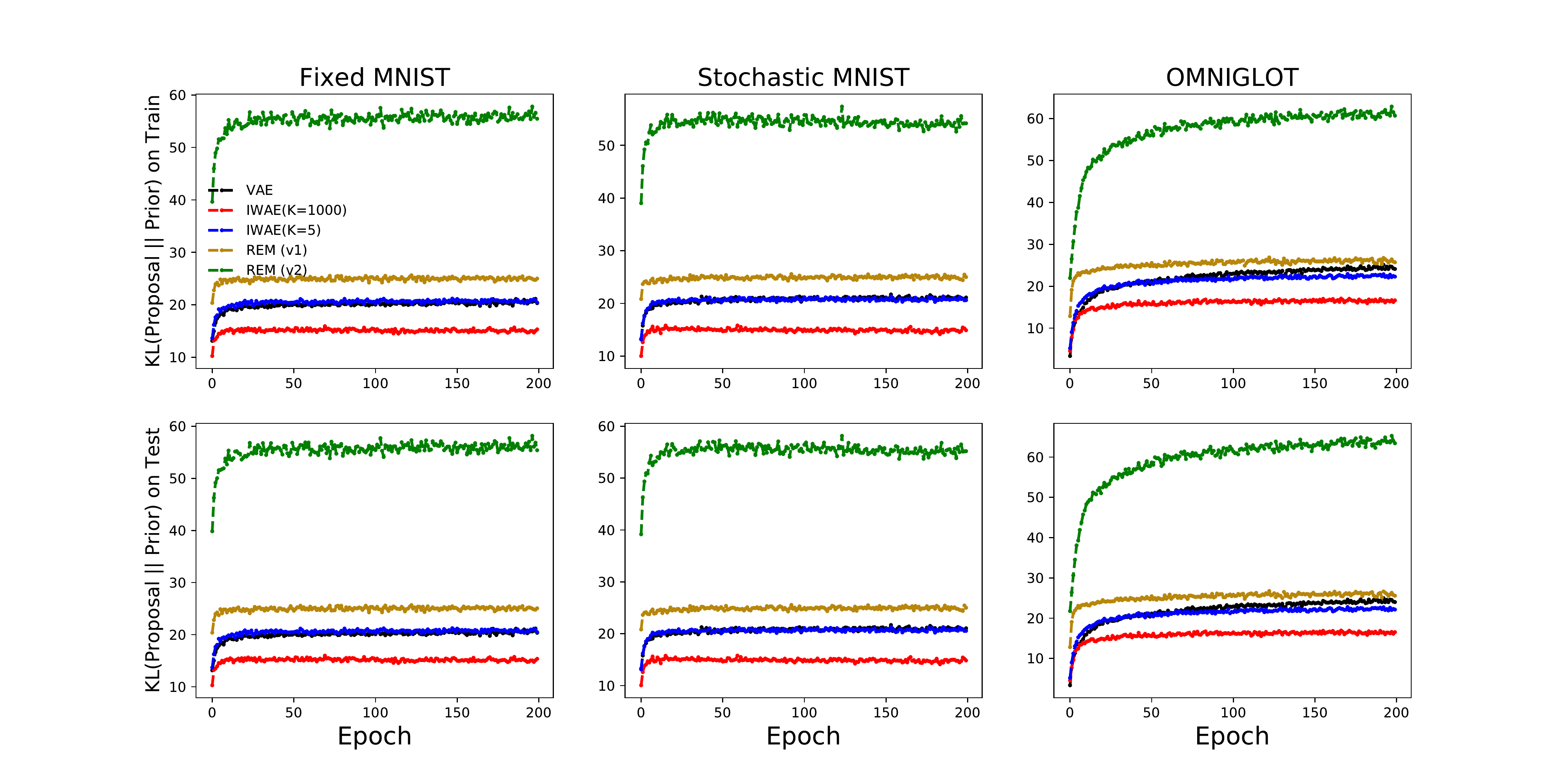}}
 	\vspace*{-4pt}
	\caption{\gls{REM} learns a better proposal than the \gls{VAE} and the \gls{IWAE}. This figure also shows that the quality of the \gls{IWAE}'s fitted posterior deteriorates as $K$ increases.}
	\label{fig:kl_prior}
\end{figure*}

\parhead{The inclusive KL is a better hyperobjective.}  We also assessed the quality of the learned proposal for each method. 
We use the \gls{KL} from the fitted proposal to the prior as a quality measure. 
This form of \gls{KL} is often used to assess latent variable collapse. Figure\nobreakspace \ref {fig:kl_prior} shows \gls{REM} learns 
better proposals than both the \gls{IWAE} and the \gls{VAE}. It also confirms the quality of 
the \gls{IWAE} degrades when the number of particles $K$ increases.

\chapter{Entropy-Regularized Adversarial Learning}\label{chap:presgan}

Maximum likelihood is the de-facto approach for fitting \glspl{PGM} to data. 
The models in Chapter\nobreakspace \ref {chap:dpgm} were fit by maximizing likelihood using \gls{AVI}. 
Because the likelihood is intractable for \glspl{DPGM}, we relied on amortized \gls{VI} and maximized the \gls{ELBO} 
\begin{align}\label{eq:elbo}
	\gls{ELBO}  &= E_{p_d(\bx)} E_{q_{\phi}(\bz \g \bx)}\left[ \log \frac{p_{\theta}(\bx , \bz)}{q_{\phi}(\bz \g \bx)} \right] 
	= -\text{KL}(q_{\phi}(\bz \g \bx) p_d(\bx) \vert\vert p_{\theta}(\bx , \bz)).
\end{align}
Maximizing the \gls{ELBO} is equivalent to minimizing the \gls{KL} between the model joint and the variational joint, 
which leads to issues such as latent variable collapse~\citep{bowman2015generating, dieng2018avoiding}. 
Furthermore, optimizing Eq.\nobreakspace \ref {eq:elbo} may lead to blurriness in the generated samples 
because of a property of the reverse $\gls{KL}$ known as \textit{zero-forcing}~\citep{minka2005divergence}.

In Chapter\nobreakspace \ref {chap:rem}, we proposed \gls{REM}, an algorithm that optimizes a better 
approximation to the log marginal likelihood of the data using \gls{EM} and moment matching.  

In this chapter, we develop \text{entropy-regularized adversarial learning} as an alternative 
to maximum likelihood for fitting \glspl{DPGM}. From the perspective of \gls{PGM}, 
entropy-regularized adversarial learning opens the door for using \gls{PGM} in tasks where 
high simulation quality matters (e.g. image generation, image superresolution, 
data augmentation, and model-based reinforcement learning.) 
From the \gls{DL} perspective, entropy-regularized adversarial learning provides a solution 
to the long-standing mode collapse problem of \glspl{GAN}. 

\section{Generative Adversarial Networks}

The \gls{GAN} of \citet{goodfellow2014generative} is a \gls{DL} technique for simulating high-quality data. 
The \gls{GAN} and its extensions have achieve state-of-the-art performance in the image domain; 
for example in image generation \citep{karras2019style, brock2018large}, 
image super-resolution \citep{ledig2017photo}, and image translation \citep{isola2017image}. 

The algorithmic idea behind \glspl{GAN} is to learn to sample high-quality data from a \textit{generator} 
by following feedback from a \textit{critic} (also called a \textit{discriminator}.) Both the generator and 
the discriminator are deep neural networks. 

A \gls{GAN} samples data by sampling noise $\bdelta$ from a fixed distribution $p(\bdelta)$ and 
then using this noise as input to the generator, the output of which is the sample from the \gls{GAN}. 
Denote by $\theta$ the parameters of the generator. Denote by $\tilde{\bx}(\bdelta; \theta)$ the \gls{GAN} sample.  
The generative process for data defined by the \gls{GAN} implies a density $p_{\theta}(\bx)$. 
However this density is undefined \citep{mohamed2016learning}. 
Although \glspl{GAN} do not define a tractable density over the generated samples, 
they can fit their parameters $\theta$ by leveraging feedback from the discriminator. 
Denote by $D_{\phi}$ the discriminator; it is a deep neural network with parameters $\phi$ 
that takes a sample and outputs the probability that the input sample is from the true data 
generating distribution or from the generator. The parameters $\theta$ and $\phi$ are learned jointly by optimizing the \gls{GAN} objective,
\begin{align}\label{eq:gan_loss}
		\Lcal_{\textrm{GAN}}(\theta, \phi) 
		&= \; \E{\bx\sim p_d(\bx)}{\log D_{\phi}(\bx)} + \E{\bdelta\sim p(\bdelta)}{\log\left(1-D_{\phi}(\tilde{\bx}(\bdelta; \theta))\right)}, 
\end{align}
where $p_d(\bx)$ is the empirical data distribution. 
\glspl{GAN} iteratively maximize the loss in Eq.\nobreakspace \ref {eq:gan_loss} with respect to $\phi$ and minimize it with respect to $\theta$. 
Maximizing the loss $\Lcal_{\textrm{GAN}}(\theta, \phi)$ with respect to $\phi$ forces the discriminator 
to assign high probability to the real data and low probability to samples from the generator. 
On the other hand, minimizing the loss $\Lcal_{\textrm{GAN}}(\theta, \phi)$ with respect to $\theta$ forces the discriminator 
to assign high probability to samples from the generator. These two iterative optimization loops are at odds with each other, 
hence the word ``adversarial" in the name of the approach. 

In practice, the minimax procedure described above is stopped when the generator produces realistic data. 
This is problematic because producing realistic data does not necessarily correlate with achieving goodness of fit to the true 
data generating distribution. For example, memorizing the training data is a trivial solution to producing realistic data. 
Fortunately, \glspl{GAN} do not merely memorize the training data~\citep{zhang2017discrimination, arora2017generalization}. 

However \glspl{GAN} are able to produce samples indistinguishable from real data while still failing 
to fully capture the data generating distribution~\citep{brock2018large, karras2019style}. 
Indeed \glspl{GAN} suffer from an issue known as \textit{mode collapse}. When mode collapse happens, 
the generative distribution $p_{\theta}(\bx)$ implied by the \gls{GAN} sampling procedure is degenerate 
and has low support~\citep{arora2017generalization, arora2018gans}. Mode collapse causes \glspl{GAN}, 
to fail both qualitatively and quantitatively. Qualitatively, mode collapse causes lack of diversity in the generated samples. 
This is problematic for certain applications of \glspl{GAN}, e.g. data augmentation. 
Quantitatively, mode collapse causes poor generalization to new data. This is because when mode collapse happens, 
there is a (support) mismatch between the learned distribution $p_{\theta}(\bx)$ and the data distribution. 
Using annealed importance sampling with a kernel density estimate of the likelihood, \citet{wu2016quantitative} 
report significantly worse log-likelihood scores for \glspl{GAN} when compared to \glspl{VAE}. 
Similarly poor generalization performance was reported by \citet{grover2018flow}.  

\subsection{Why does mode collapse happen?}\label{subsec:collapse}
For simplicity, and only for the rest of this section, let's denote by $t(\bx)$ a target distribution of interest. 
Assume we are using the \gls{GAN} minimax framework to approximate $t(\bx)$ with $f(\bx)$. 
Denote by $D(\bx)$ the discriminator. The loss is,
\begin{align}\label{eq:gan_loss_toy}
		\Lcal_{\textrm{GAN}}
		&= \; \E{\bx\sim t(\bx)}{\log D(\bx)} + \E{\bx \sim f(\bx)}{\log (1 - D(\bx))}.
\end{align}
The loss $\Lcal_{\textrm{GAN}}$ in Eq.\nobreakspace \ref {eq:gan_loss_toy} is a concave function of $D(\bx)$. Taking the gradient of $\Lcal_{\textrm{GAN}}$ in Eq.\nobreakspace \ref {eq:gan_loss_toy} with respect to $D(\bx)$ and setting it to zero yields the optimal discriminator~\citep{goodfellow2014generative}, 
\begin{align}\label{eq:optimal_disc}
		D^*(\bx)
		&= \frac{t(\bx)}{t(\bx) + f(\bx)}
\end{align}
Replacing this optimal discriminator in Eq.\nobreakspace \ref {eq:gan_loss_toy} and rearranging terms leads to the following objective 
for learning $f(\bx)$:
 \begin{align}\label{eq:gen_loss}
		\text{JS}(t(\bx) \vert\vert f(\bx))
		&= \frac{1}{2}\gls{KL}(t(\bx) \vert\vert r(\bx)) + \frac{1}{2}\gls{KL}(f(\bx) \vert\vert r(\bx)),
\end{align}
where $r(\bx) = \frac{t(\bx) + f(\bx)}{2}$. 

Let's look more closely at the objective function $\text{JS}(t(\bx) \vert\vert f(\bx))$, which we minimize to find a good approximation $f(\bx)$ for the target distribution $t(\bx)$. The objective $\text{JS}(t(\bx) \vert\vert f(\bx))$ is the sum of two \gls{KL} divergences. The first \gls{KL}, $\gls{KL}(t(\bx) \vert\vert r(\bx))$ has a \textit{zero-avoiding} behavior~\citep{minka2005divergence, dieng2017variational}, minimizing it yields a distribution $r(\bx)$ that overgeneralizes $t(\bx)$. This can be achieved without requiring $f(\bx)$ to cover all the modes of $t(\bx)$. Furthermore, the second \gls{KL} term, $\gls{KL}(f(\bx) \vert\vert r(\bx))$, has a \textit{zero-forcing} behavior~\citep{minka2005divergence, dieng2017variational}, minimizing it yields a distribution $f(\bx)$ that undergeneralizes $r(\bx)$. As a consequence, minimizing $\text{JS}(t(\bx) \vert\vert f(\bx))$ tends to lead to a distribution $f(\bx)$ that does not cover all the modes of the target distribution $t(\bx)$. 

\subsection{Motivating entropy regularization} 
In light of the analysis in Section\nobreakspace \ref {subsec:collapse}, a natural way to prevent mode collapse in \glspl{GAN} is to maximize entropy~\citep{belghazi2018mine}. Indeed, adding the entropy of $f(\bx)$ to the objective $\text{JS}(t(\bx) \vert\vert f(\bx))$ leads to an entropy-regularized objective,
 \begin{align}\label{eq:optimal_disc}
		\mathcal{L}(f(\bx))
		&= \frac{1}{2}\gls{KL}(t(\bx) \vert\vert r(\bx)) + \frac{1}{2}\gls{KL}(f(\bx) \vert\vert r(\bx)) - \lambda \mathbb{E}_{\bx\sim f(\bx)} [\log f(\bx)]\\
		&= \frac{1}{2}\gls{KL}(t(\bx) \vert\vert r(\bx)) + \left(\frac{1}{2} - \lambda\right) \gls{KL}(f(\bx) \vert\vert r(\bx)) - \lambda \mathbb{E}_{\bx\sim f(\bx)} [\log r(\bx)].\label{eq:two}
\end{align}
Let's now look closely at Eq.\nobreakspace \ref {eq:two}. The first \gls{KL}, $\gls{KL}(t(\bx) \vert\vert r(\bx))$ has the same weight as in Eq.\nobreakspace \ref {eq:gen_loss}, it yields $r(\bx)$ that overgeneralizes $t(\bx)$. The second \gls{KL}, which leads to a distribution $f(\bx)$ that undergeneralizes $r(\bx)$, has reduced effect. There is a new term, $-\mathbb{E}_{\bx\sim f(\bx)} [\log r(\bx)]$, whose minimization enforces high cross-entropy between $f(\bx)$ and $r(\bx)$. This in turn forces $f(\bx)$ to cover the modes of the target distribution $t(\bx)$.

Unfortunately maximizing entropy is impossible for \glspl{GAN}, because their entropy is not well-defined.  

\gls{GAN} researchers have looked at indirect ways to alleviate mode collapse. 
For example, \citet{srivastava2017veegan} use a reconstructor network that 
reverses the action of the generator. \citet{lin2018pacgan} use multiple observations 
(either real or generated) as an input to the discriminator to prevent mode collapse. 
\citet{azadi2018discriminator} and \citet{turner2018metropolis} use sampling mechanisms 
to correct errors of the generative distribution. \citet{xiao2018bourgan} relies on identifying the 
geometric structure of the data embodied under a specific distance metric.
Other works have combined adversarial learning with maximum likelihood \citep{grover2018flow, yin2019semi}; 
however, the low sample quality induced by maximum likelihood still occurs. 
Finally, \citet{cao2018improving} introduce a regularizer for the discriminator to encourage 
diverse activation patterns in the discriminator across different samples.  

\section{Prescribed Generative Adversarial Networks} 
In this section we leverage the minimax procedure used when fitting \glspl{GAN}, called adversarial learning, within the context of \glspl{DPGM}. 
We build on adversarial learning in such a way that our desiderata for \gls{DPGM} are met. 
In particular, we maximize entropy to enforce diversity in the data generating process. 
We call the resulting learning algorithm \textit{entropy-regularized adversarial learning}. 
We call a \gls{DPGM} fit using entropy-regularized adversarial learning a prescribed \gls{GAN} (Pres\gls{GAN}.) 

In this section, we focus on a simple \gls{DPGM} where the generative process 
is to sample from the prior $p(\bz)$ and then condition on the sample to draw data 
from $p_{\theta}(\bx\g \bz)$, an exponential family distribution parameterized by a deep neural network. 
This generative process implies a well-defined density over $\bx$, 
\begin{equation}\label{eq:vaes}
	p_{\theta}(\bx) = \int_{}^{} p_{\theta}(\bx\g \bz) \cdot p(\bz) d\bz.
\end{equation}

For simplicity we define the prior $p(\bz)$ and the likelihood $p_{\theta}(\bx\g \bz)$ to be Gaussians, 
\begin{equation}\label{eq:semi_implicit}
	p(\bz) = \mathcal{N}(\bz\g\bzero, \bI) \quad \text{and} \quad 
	p_{\theta}(\bx\g \bz) = \Ncal\left(\bx\g \bmu_{\theta}(\bz), \bSigma_{\theta}(\bz) \right).
\end{equation}
The mean $\bmu_{\theta}(\bz)$ and covariance $\bSigma_{\theta}(\bz)$ of the 
conditional $p_{\theta}(\bx\g \bz)$ are given by a neural network that takes $\bz$ as input.
In general, both the mean $\bmu_{\theta}(\bz)$ and the covariance $\bSigma_{\theta}(\bz)$ can be functions of $\bz$. 
For simplicity, in order to speed up the learning procedure, we set the covariance matrix to be diagonal 
with elements independent from $\bz$, i.e., $\bSigma_{\theta}(\bz)=\diag{\bsigma^2}$, and we learn the 
vector $\bsigma$ together with $\theta$. From now on, we parameterize the mean with $\eta$, 
write $\bmu_{\eta}(\bz)$, and define $\theta = (\eta, \bsigma)$ as the parameters of the generative distribution.

To fit the model parameters $\theta$, we optimize an adversarial loss similarly to \glspl{GAN}. 
Unlike \glspl{GAN}, the entropy of the generative distribution of a Pres\gls{GAN} is well-defined, 
and therefore we can prevent mode collapse by adding an entropy regularizer to Eq.\nobreakspace \ref {eq:gan_loss}. 
The idea of entropy regularization has been widely applied in 
many problems that involve estimation of unknown probability distributions. Examples include approximate 
Bayesian inference, where the variational objective contains an entropy 
penalty \citep{jordan1998learning,bishop2006pattern,wainwright2008graphical,blei2017variational}; 
reinforcement learning, where the entropy regularization allows to estimate more uncertain and 
explorative policies \citep{schulman2015trust, mnih2016asynchronous}; statistical learning, where entropy 
regularization allows an inferred probability distribution to  avoid collapsing to a deterministic 
solution \citep{freund1997decision, soofi2000principal, jaynes2003probability}; or 
optimal transport \citep{rigollet2018entropic}. More recently, \citet{kumar2019maximum} have 
developed maximum-entropy generators for energy-based models using mutual information as a proxy for entropy. 

Entropy regularized adversarial learning keeps \glspl{GAN}' ability to generate samples with high perceptual quality 
while enforcing diversity in the data generation process. The loss is 
\begin{equation}\label{eq:sigan_loss}
	\Lcal_{\text{Pres}\textrm{GAN}}(\theta, \phi) = \Lcal_{\textrm{GAN}}(\theta, \phi) - \lambda \mathcal{H}\left(p_{\theta}(\bx)\right).
\end{equation}
Here $\mathcal{H}\left(p_{\theta}(\bx)\right)$ denotes the entropy of the generative distribution. It is defined as
\begin{equation}\label{eq:def_entropy}
	\mathcal{H}\left(p_{\theta}(\bx)\right) = -\E{p_{\theta}(\bx)}{\log p_{\theta}(\bx)}.
\end{equation}
The loss $\Lcal_{\textrm{GAN}}(\theta, \phi)$ in Eq.\nobreakspace \ref {eq:sigan_loss} can be that of any of the existing \gls{GAN} variants. 
In our empirical study we explore the standard \gls{DCGAN}~\citep{radford2015unsupervised} and the more 
recent Style\gls{GAN}~\citep{karras2019style}.  

The constant $\lambda$ in Eq.\nobreakspace \ref {eq:sigan_loss} is a hyperparameter that controls the strength of the entropy regularization. 
In the extreme case when $\lambda=0$, the loss function $\Lcal_{\text{Pres}\textrm{GAN}}(\theta, \phi)$ coincides with the 
loss of a \gls{GAN}, where we replaced its ill-defined generative distribution with that in Eq.\nobreakspace \ref {eq:vaes}. 
In the other extreme when $\lambda=\infty$, optimizing $\Lcal_{\text{Pres}\textrm{GAN}}(\theta, \phi)$ 
corresponds to fitting a maximum entropy generator that ignores the data. For any intermediate 
values of $\lambda$, the first term of $\Lcal_{\text{Pres}\textrm{GAN}}(\theta, \phi)$ encourages the 
generator to fit the data distribution, whereas the second term encourages diversity.

The entropy $\mathcal{H}\left(p_{\theta}(\bx)\right)$ is intractable because the integral 
in Eq.\nobreakspace \ref {eq:def_entropy} cannot be computed. However, fitting the parameters $\theta$ of Pres\glspl{GAN} 
only requires the gradients of the entropy. 

We fit Pres\glspl{GAN} following the same adversarial procedure used in \glspl{GAN}. 
That is, we alternate between updating the parameters of the generative distribution $\theta$ 
and the parameters of the discriminator $\phi$. The full procedure is given in Algorithm\nobreakspace \ref {alg:full_algo}.
We now describe each part in detail. 

\subsection{Fitting the discriminator}
Since the entropy term in Eq.\nobreakspace \ref {eq:sigan_loss} does not depend on $\phi$, optimizing the discriminator is 
analogous to optimizing the discriminator of a \gls{GAN},
\begin{align}\label{eq:grad_phi}
\nabla_{\phi}\Lcal_{\text{Pres}\textrm{GAN}}(\theta, \phi) = \nabla_{\phi}\Lcal_{\textrm{GAN}}(\theta, \phi).
\end{align}

To prevent the discriminator from getting stuck in a bad local optimum where it can perfectly 
distinguish between real and generated data by relying on the added noise, we apply the same 
amount of noise to the real data $\bx$ as the noise added to the generated data. 
That is, when we train the discriminator we corrupt the real data according to
\begin{align}\label{eq:noised_data}
	\widehat{\bx} &= \bx + \bsigma \odot \bepsilon,
\end{align}
where $\bsigma$ is the standard deviation of the generative distribution and $\bx$ denotes the real data. 
We then let the discriminator distinguish between $\widehat{\bx}$ and $\bx(\bz,\bepsilon; \theta)$ from Eq.\nobreakspace \ref {eq:sigan_sample}. 

Using the same noise has a theoretical motivation. Let $p_d(\bx)$ denote the data distribution and $p_g(\bx)$ the distribution implied by the sampling procedure: 
\begin{align}\label{eq:gen_mean}
	\bz &\sim p(\bz) \text{ and } \bx = \bmu_{\eta}(\bz)
\end{align}
where $\bmu_{\eta}(\cdot)$ is the output of the generator. Adding noise with the same variance $\bsigma$ to a sample from $p_d(\bx)$ and to a sample from $p_g(\bx)$ is equivalent to convolving both distributions with the same Gaussian $\mathcal{N}(\tilde\bx \vert \bx, \bsigma^2)$:
\begin{align}
	p_{d, \sigma}(\tilde \bx) &= \int_{}^{} p_d(\bx)\mathcal{N}(\tilde\bx \vert \bx, \bsigma^2) d\bx \\
	p_{g, \sigma}(\tilde \bx) &= \int_{}^{} p_g(\bx)\mathcal{N}(\tilde\bx \vert \bx, \bsigma^2) d\bx
\end{align}

Now observe that if $p_d = p_g$ then $p_{d,\sigma} = p_{g, \sigma}$ for any value of $\bsigma$. This property holds only when using the same noise variance $\sigma$.  

The data noising procedure described above is a form of \textit{instance noise} \citep{sonderby2016amortised}. 
However, instead of using a fixed annealing schedule for the noise variance as \citet{sonderby2016amortised}, 
we let $\bsigma$ be part of the parameters of the generative distribution and fit it using gradient descent according to Eq.\nobreakspace \ref {eq:grad_sigma}.

\subsection{Fitting the generator}
We fit the generator using stochastic gradient descent. This requires computing the 
gradients of $\Lcal_{\text{Pres}\textrm{GAN}}(\theta, \phi)$ with respect to $\theta$,
\begin{equation}\label{eq:grad_theta}
	\nabla_{\theta}\Lcal_{\text{Pres}\textrm{GAN}}(\theta, \phi) = \nabla_{\theta}\Lcal_{\textrm{GAN}}(\theta, \phi) - \lambda \nabla_{\theta} \mathcal{H}\left(p_{\theta}(\bx)\right).
\end{equation}
We form stochastic estimates of $\nabla_{\theta}\Lcal_{\textrm{GAN}}(\theta, \phi)$ based on 
reparameterization \citep{kingma2013auto,rezende2014stochastic,titsias2014doubly}; this requires 
differentiating Eq.\nobreakspace \ref {eq:gan_loss}. Specifically, we introduce a noise variable $\bepsilon$ to 
reparameterize the conditional from Eq.\nobreakspace \ref {eq:semi_implicit},\footnote{With this reparameterization we use the notation $\bx(\bz, \beps; \theta)$ instead of $\tilde{\bx}(\bz; \theta)$ 
to denote a sample from the generative distribution.}
\begin{align}\label{eq:sigan_sample}
	\bx(\bz, \beps; \theta) &= \bmu_{\eta}(\bz) + \bsigma \odot \bepsilon,
\end{align}
where $\theta = (\eta, \bsigma)$ and $\bepsilon \sim \mathcal{N}(\bzero, \bI)$. 
Here $\bmu_{\eta}(\bz)$ and $\bsigma$ denote the mean and standard deviation of the 
conditional $p_{\theta}(\bx\g\bz)$, respectively. 
We now write the first term of Eq.\nobreakspace \ref {eq:grad_theta} as an expectation with respect to the 
latent variable $\bz$ and the noise variable $\beps$ and push the gradient into the expectation,
\begin{align}\label{eq:grad_gan}
	\nabla_{\theta}\Lcal_{\textrm{GAN}}(\theta, \phi) &= \mathbb{E}_{p(\bz) p(\bepsilon)} \left[\nabla_{\theta} \log\left(1-D_{\phi}(\bx(\bz, \beps; \theta))\right)\right]
	.
\end{align}
In practice we use an estimate of Eq.\nobreakspace \ref {eq:grad_gan} using one sample from $p(\bz)$ and one sample from $p(\bepsilon)$,
\begin{align}\label{eq:estimate_grad_gan}
		\widehat{\nabla}_{\theta} \Lcal_{\textrm{GAN}}(\theta, \phi) &= \nabla_{\theta} \log\left(1-D_{\phi}(\bx(\bz, \beps; \theta))\right)
\end{align}
The second term in Eq.\nobreakspace \ref {eq:grad_theta}, corresponding to the gradient of the entropy, is intractable.  
We estimate it using the same approach as \citet{titsias2018unbiased}. 
We first use the reparameterization in Eq.\nobreakspace \ref {eq:sigan_sample} to express the gradient of the entropy as an expectation,
\begin{align*}
    \nabla_{\theta} \mathcal{H}\left(p_{\theta}(\bx)\right) 
    &= - \nabla_{\theta} \E{p_{\theta}(\bx)}{\log p_{\theta}(\bx)}\\
    &= - \nabla_{\theta} \E{p(\bepsilon)p(\bz)}{\log p_{\theta}(\bx)\big|_{\bx = \bx(\bz, \beps; \theta)}} \\
    & = - \E{p(\bepsilon)p(\bz)}{\nabla_{\theta} \log p_{\theta}(\bx)\big|_{\bx=\bx(\bz, \beps; \theta)}}\\
    &= - \E{p(\bepsilon)p(\bz)}{\nabla_{\bx} \log p_{\theta}(\bx)\big|_{\bx=\bx(\bz, \beps; \theta)} \nabla_{\theta}\bx(\bz,\bepsilon; \theta)},
\end{align*}
where we have used the score function identity $\E{p_{\theta}(\bx)}{\nabla_{\theta}\log p_{\theta}(\bx)}=0$ on the second line.
We form a one-sample estimator of the gradient of the entropy as
\begin{align}\label{eq:estimate_grad_entropy}
		\widehat{\nabla}_{\theta} \mathcal{H}\left(p_{\theta}(\bx)\right) 
		&=  -\nabla_{\bx} \log p_{\theta}(\bx)\big|_{\bx = \bx(\bz,\bepsilon; \theta)} \times \nabla_{\theta} \bx(\bz,\bepsilon; \theta).
\end{align}
In Eq.\nobreakspace \ref {eq:estimate_grad_entropy}, the gradient with respect to the reparameterization transformation $\nabla_{\theta} \bx(\bz,\bepsilon;\theta)$ is tractable and can be obtained via back-propagation. We now derive $\nabla_{\bx} \log p_{\theta}(\bx)$,
\begin{align*}
	\nabla_{\bx} \log p_{\theta}(\bx) 
	& = \frac{\nabla_{\bx}p_{\theta}(\bx)}{p_{\theta}(\bx)} \\
	&= \frac{\int_{}^{}\nabla_{\bx}p_{\theta}(\bx, \bz) d\bz}{p_{\theta}(\bx)}\\
	&= \int_{}^{}\frac{\frac{\nabla_{\bx}p_{\theta}(\bx \g \bz)}{p_{\theta}(\bx \g \bz)} p_{\theta}(\bx , \bz)}{p_{\theta}(\bx)}d\bz \\
	&= \int_{}^{} \nabla_{\bx}\log p_{\theta}(\bx \g \bz) p_{\theta}(\bz \g \bx) d\bz \\
	&= \E{p_{\theta}(\bz \g \bx)}{\nabla_{\bx}\log p_{\theta}(\bx \g \bz)}.
\end{align*}
While this expression is still intractable, we can estimate it. One way is to use self-normalized importance 
sampling with a proposal learned using moment matching with an encoder as we did in Chapter\nobreakspace \ref {chap:rem}~\citep{dieng2019reweighted}. 
However, this would lead to a biased (albeit asymptotically unbiased) estimate of the entropy. 
In this paper, we form an unbiased estimate of $\nabla_{\bx} \log p_{\theta}(\bx)$ using samples $\bz^{(1)}, \dots, \bz^{(M)}$ from the posterior,
\begin{equation}\label{eq:estimate_logdensity}
	\widehat{\nabla}_{\bx} \log p_{\theta}(\bx)
	= \frac{1}{M}\sum_{m=1}^{M}\nabla_{\bx} \log p_{\theta}(\bx \g \bz^{(m)}), \quad \text{where} \quad \bz^{(m)}\sim p_{\theta}(\bz \g \bx). 
\end{equation}
We obtain these samples using \gls{HMC} \citep{neal2011mcmc}.
Crucially, in order to speed up the algorithm, we initialize the \gls{HMC} sampler at stationarity. 
That is, we initialize the \gls{HMC} sampler with the sample $\bz$ that was used to produce 
the generated sample $\bx(\bz,\bepsilon; \theta)$ in Eq.\nobreakspace \ref {eq:sigan_sample}, which by construction 
is an exact sample from $p_{\theta}(\bz \g \bx)$. This implies that only a few \gls{HMC} iterations suffice 
to get good estimates of the gradient \citep{titsias2018unbiased}. We also found this holds empirically; 
for example in the empirical study, we use $2$ burn-in iterations and $M=2$ \gls{HMC} samples to 
form the Monte Carlo estimate in Eq.\nobreakspace \ref {eq:estimate_logdensity}.

Finally, using 
Eqs.\nobreakspace \ref {eq:grad_theta} and\nobreakspace   \ref {eq:estimate_grad_gan} to\nobreakspace  \ref {eq:estimate_logdensity} 
we can approximate the gradient of the entropy-regularized adversarial loss with respect to the model parameters $\theta$, 
\begin{align}\label{eq:grad_theta2}
  \widehat{\nabla}_{\theta}\Lcal_{\text{Pres}\textrm{GAN}}(\theta, \phi) 
  &= \nabla_{\theta} \log\left(1-D_{\phi}(\bx(\bz, \beps; \theta))\right) \nonumber\\
  &+ \frac{\lambda}{M}\sum_{m=1}^{M} \nabla_{\bx} \log p_{\theta}(\bx \g \bz^{(m)})\big|_{\bx = \bx(\bz^{(m)},\bepsilon;\theta)} \times \nabla_{\theta} \bx\left(\bz^{(m)},\bepsilon;\theta\right).
\end{align}
In particular, the gradient with respect to the generator's parameters $\eta$ is unbiasedly approximated by
\begin{align}\label{eq:grad_eta}
  \widehat{\nabla}_{\eta}\Lcal_{\text{Pres}\textrm{GAN}}(\theta, \phi) 
  &= \nabla_{\eta} \log\left(1-D_{\phi}(\bx(\bz, \beps; \theta))\right)
  - \frac{\lambda}{M}\sum_{m=1}^{M} \frac{\bx(\bz^{(m)},\bepsilon;\theta) - \bmu_{\eta}\left(\bz^{(m)}\right)}{\bsigma^2} \nabla_{\eta} \bmu_{\eta}(\bz^{(m)}),
\end{align}
and the gradient estimator with respect to the standard deviation $\bsigma$ is 
\begin{align}\label{eq:grad_sigma}
  \widehat{\nabla}_{\bsigma}\Lcal_{\text{Pres}\textrm{GAN}}(\theta, \phi) &= \nabla_{\bsigma} \log\left(1-D_{\phi}(\bx(\bz, \beps; \theta))\right) 
  - \frac{\lambda}{M}\sum_{m=1}^{M} \frac{\bx(\bz^{(m)},\bepsilon;\theta) - \bmu_{\eta}\left(\bz^{(m)}\right)}{\bsigma^2} \cdot \bepsilon.
\end{align}
These gradients are used in a stochastic optimization algorithm to fit the generative distribution of Pres\gls{GAN}.

\begin{algorithm}[t]
  \caption{Entropy-Regularized Adversarial Learning}\label{alg:full_algo}
  \begin{algorithmic}
    \STATE Initialize parameters $\eta, \bsigma, \phi$
    \FOR{iteration $t=1,2,\ldots$}
    \STATE Draw minibatch of observations $\bx_1, \dots, \bx_b, \dots, \bx_B$
    \FOR{$b=1,2,\ldots, B$}
    \STATE Get noised data: $ \bepsilon_b \sim \mathcal{N}(\bzero, \bI)$ and $\widehat{\bx}_b = \bx_b + \bsigma \odot \bepsilon_b$
    \STATE Draw latent variable $\bz_b \sim \mathcal{N}(\bzero, \bI)$
    \STATE Generate data: $ \bs_b \sim \mathcal{N}(\bzero, \bI)$ and $\tilde{\bx}_b=\tilde{\bx}_b(\bz_b,\bs_b; \theta) = \bmu_{\eta}(\bz_b) + \bsigma \odot \bs_b$
    \ENDFOR
    \STATE Compute $\nabla_{\phi}\Lcal_{\text{Pres}\textrm{GAN}}(\theta, \phi)$ (Eq.\nobreakspace \ref {eq:grad_phi}) and take a gradient step for $\phi$
    \STATE Initialize an \acrshort{HMC} sampler using $\bz_b$
    \STATE Draw $\tilde{\bz}_b^{(m)} \sim p_{\theta}(\bz\g \tilde{\bx}_b)$ for $m = 1, \dots, M$ and $b = 1, \dots, B$ using that sampler
    \STATE Compute $\widehat{\nabla}_{\eta}\Lcal_{\text{Pres}\textrm{GAN}}((\eta, \bsigma), \phi)$ (Eq.\nobreakspace \ref {eq:grad_eta}) and take a gradient step for $\eta$
    \STATE Compute $\widehat{\nabla}_{\bsigma}\Lcal_{\text{Pres}\textrm{GAN}}((\eta, \bsigma), \phi)$ (Eq.\nobreakspace \ref {eq:grad_sigma}) and take a gradient step for $\bsigma$
    \STATE Truncate $\bsigma$ in the range $[\bsigma_{\text{low}}, \bsigma_{\text{high}}]$
    \ENDFOR
  \end{algorithmic} 
\end{algorithm}

Note there are two failure cases brought in by learning the variance $\bsigma^2$ 
using gradient descent. 

The first failure mode is when $\bsigma^2$ gets very small, which makes the 
gradient of the entropy in Eq.\nobreakspace \ref {eq:grad_eta} dominate the overall gradient of the generator. 
This is problematic because the learning signal from the discriminator is lost. 

The second failure mode is when the variance gets very large. Consider the adversarial loss with data noising,
\begin{align}
	\mathcal{L}(\eta, \bsigma, \phi) 
	&= \mathbb{E}_{p_d(\bx)p(\bepsilon)}\left[\log D_{\phi}(\bx + \bsigma \odot \bepsilon)\right] 
	+ \mathbb{E}_{p(\bz)p(\bepsilon)}\left[ \log\left(1 - D_{\phi}\left(\bmu_{\eta}(\bz) + \bsigma \odot \bepsilon\right)\right) \right]
\end{align}
When $p_d = p_g$, then the adversarial loss function $\mathcal{L}(\eta, \bsigma, \phi)$ is constant with respect to $\bsigma$ and 
as a result, the gradient of $\mathcal{L}(\eta, \bsigma, \phi)$ with respect to $\bsigma$ is zero. However, during training $p_d \ne p_g$. 
This can lead to large values for $\bsigma$ because the generator can completely fool the discriminator 
so that $D_{\phi}(\tilde \bx) = \frac{1}{2}$, its optimal value, by letting $\bsigma \rightarrow \infty$.
However setting $\bsigma$ very large is undesirable since it corresponds to the bad equilibrium point where 
the samples from the data distribution and from the generative distribution are indistinguishable from one another 
simply because they are both buried in noise.    

\subsection{Variance Regularization}
We propose to alleviate the two failure modes discussed above by regularizing the variance to prevent it from reaching very low or very large values. 

\parhead{Truncation.} One way to regularize the variance $\bsigma$ is to simply bound it during optimization, 
$\bsigma_{\text{low}} \leq \bsigma \leq \bsigma_{\text{high}}$. Note this is applied element-wise. 
The limits $\bsigma_{\text{low}}$ and $\bsigma_{\text{high}}$ are hyperparameters. 

\parhead{Entropy minimization.} To avoid large values of $\bsigma$, we can minimize the entropy of the noise process 
$\mathcal{N}(\tilde \bx \vert \bx, \bsigma^2)$. The regularized objective for Pres\gls{GAN} becomes
\begin{align}\label{eq:full_presgan}
	\mathcal{L}_{\text{PresGAN}}(\eta, \bsigma, \phi) 
	&= \mathcal{L}_{\text{GAN}}(\eta, \bsigma, \phi) - \lambda \mathcal{H}(p_{\eta, \sigma}(\bx))
	+ \tilde\lambda \sum_{d=1}^{D} \log \bsigma_d^2 
\end{align}
where $\tilde\lambda > 0$ is a hyperparameter that determines the strength of the regularization of the entropy of the noise process. The hyperparameter $\lambda$ controls the entropy regularization of the generative distribution, as described earlier. 

Note making $\bsigma$ arbitrarily large increases the entropy of the generative distribution $p_{\eta, \sigma}(\bx)$. However, the term $\tilde\lambda \sum_{d=1}^{D} \log \bsigma_d^2$ in Eq.\nobreakspace \ref {eq:full_presgan} will prevent that behavior and ensures the entropy of the generative distribution is maximized by means of the latent variables $\bz$ and not the noise variance $\bsigma$. 

Regularizing the variance of the noise process as described above yields an interesting result we summarize in the following proposition. 

\textit{\parhead{Proposition.}} \textit{Consider the generative distribution of Pres\gls{GAN} 
under a Gaussian likelihood}
\begin{align}
	p_{\eta, \sigma}(\bx) &= \int_{}^{} \mathcal{N}(\bx \vert \bmu_{\eta}(\bz), \bsigma^2) p(\bz) d\bz. \nonumber
\end{align}
\textit{Then when $\tilde\lambda = \lambda$ in Eq.\nobreakspace \ref {eq:full_presgan}},
\begin{align}
	 \mathcal{I}(\bx, \bz) &= \mathcal{H}(p_{\eta, \sigma}(\bx)) - \sum_{d=1}^{D} \log \bsigma_d^2 \nonumber
\end{align}
\textit{where $\mathcal{I}(\bx, \bz)$ denotes the mutual information between $\bx$ and $\bz$ under the generative model.}

The proposition above means that under Gaussian likelihood and Gaussian noise process, optimizing Eq.\nobreakspace \ref {eq:full_presgan} is equivalent to adversarial learning with a mutual information regularizer. 

\textit{\parhead{Proof.}} Denote by $\mathcal{I}(\bx, \bz)$ the mutual information between $\bx$ and $\bz$ under the Pres\gls{GAN} generative distribution. Then,
\begin{align}
	\mathcal{I}(\bx, \bz) 
	&= \int_{}^{} p_{\eta, \sigma}(\bx, \bz) \log \frac{p_{\eta, \sigma}(\bx, \bz)}{p_{\eta, \sigma}(\bx)p(\bz)} d\bx d\bz\\
	&= \int_{}^{} p_{\eta, \sigma}(\bx, \bz) \log \frac{p_{\eta, \sigma}(\bx \vert \bz)}{p_{\eta, \sigma}(\bx)} d\bx d\bz\\
	&= -\int_{}^{} p_{\eta, \sigma}(\bx) \log p_{\eta, \sigma}(\bx) d\bx + \int_{}^{} p(\bz) \left(\int_{}^{} p_{\eta, \sigma}(\bx \vert \bz) \log p_{\eta, \sigma}(\bx \vert \bz) d\bx \right) d\bz\\
	&= \mathcal{H}(p_{\eta, \sigma}(\bx)) - \sum_{d=1}^{D} \log \bsigma_d^2 + \text{cst}
\end{align}
where we used the Gaussian assumption on the likelihood to replace $\int_{}^{} p_{\eta, \sigma}(\bx \vert \bz) \log p_{\eta, \sigma}(\bx \vert \bz) d\bx$, the negative entropy of a Gaussian, with $-\sum_{d=1}^{D} \log \bsigma_d^2 + \text{cst.}$

\section{Empirical Study}

Here we demonstrate Pres\glspl{GAN}' ability to prevent mode collapse and generate high-quality samples. We also evaluate its predictive performance as measured by log-likelihood.

\subsection{Simulation Study}
\label{subsec:illustrative}

In this section, we fit a \gls{GAN} to a toy synthetic dataset of $10$ modes. We choose the hyperparameters such that the \gls{GAN} collapses. We then apply these same hyperparameters to fit a Pres\gls{GAN} on the same synthetic dataset. This experiment demonstrates the Pres\gls{GAN}'s ability to correct the mode collapse problem of a \gls{GAN}.

We form the target distribution by organizing a uniform mixture of $K=10$ two-dimensional Gaussians on a ring. The radius of the ring is $r = 3$ and each Gaussian has standard deviation $0.05$. We then slice the circle 
into $K$ parts. The location of the centers of the mixture components are determined as follows. Consider the $k^{\textrm{th}}$ mixture component. Its coordinates in the $2$D space are
\begin{align*}
	\text{center}_x &=  r \cdot \text{cos}\Big(k \cdot \frac{2\pi}{K}\Big) \quad \text{and} \quad
	\text{center}_y = r \cdot \text{sin}\Big(k \cdot \frac{2\pi}{K}\Big)
	.
\end{align*}
We draw $5{,}000$ samples from the target distribution and fit a \gls{GAN} and a Pres\gls{GAN}.

\begin{figure*}[t]
	\centerline{\includegraphics[width=1\textwidth, height=4.5cm]{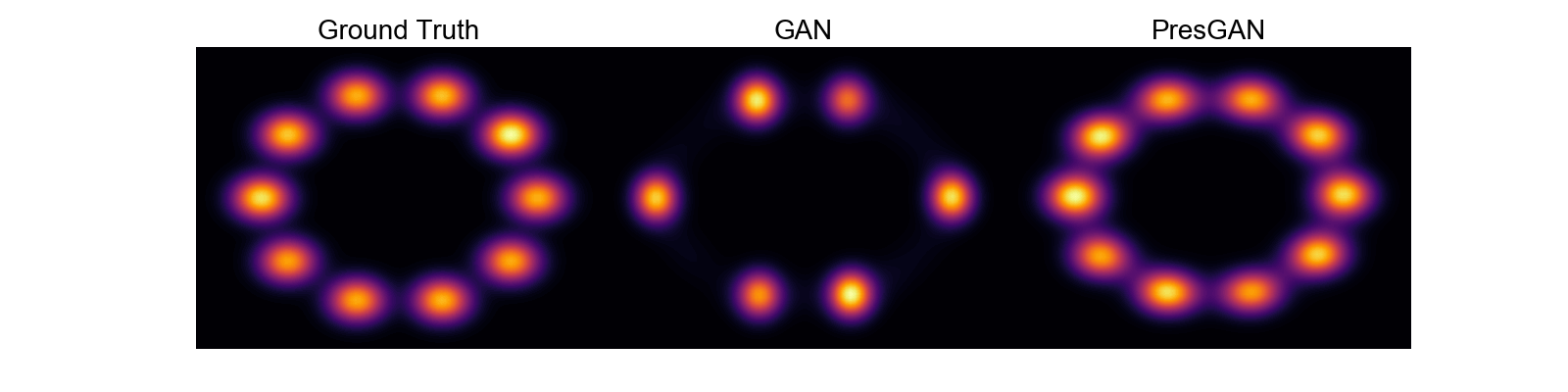}}
	\caption[Simulation study illustrating how entropy regularization prevents a \acrshort{GAN} from collapsing]{Density estimation with \acrshort{GAN} and Pres\acrshort{GAN} on a toy two-dimensional experiment. The ground truth is a uniform mixture of $10$ Gaussians organized on a ring. Given the right set of hyperparameters, a \acrshort{GAN} could perfectly fit this target distribution. In this example we chose the \acrshort{GAN} hyperparameters such that it collapses---here $4$ out of $10$ modes are missing. We then fit the Pres\acrshort{GAN} using the same hyperparameters as the collapsing \acrshort{GAN}. The Pres\acrshort{GAN} is able to correct the collapsing behavior of the \acrshort{GAN} and learns a good fit for the target distribution.}
	\label{fig:toy_example_fig1}
\end{figure*}

We set the dimension of the latent variables $\bz$ used as the input to the generators to $10$. We let both the generators and the discriminators have three fully connected layers with tanh activations and $128$ hidden units in each layer. We set the minibatch size to $100$ and use Adam for optimization \citep{kingma2014adam}, with a learning rate of $10^{-3}$ and $10^{-4}$ for the discriminator and the generator respectively. The Adam hyperparameters are $\beta_1 = 0.5$ and $\beta_2 = 0.999$. We take one step to optimize the generator for each step of the discriminator. We pick a random minibatch at each iteration and run both the \gls{GAN} and the Pres\gls{GAN} for $500$ epochs. 

For Pres\gls{GAN} we set the burn-in and the number of \gls{HMC} samples to $2$. We choose a standard number of $5$ leapfrog steps and set the \gls{HMC} learning rate to $0.02$. The acceptance rate is fixed at $0.67$. The log-variance of the noise of the generative distribution of Pres\gls{GAN} is initialized at $0.0$. We put a threshold on the variance to a minimum value of $\bsigma_{\text{low}} = 10^{-2}$ and a maximum value of $\bsigma_{\text{high}} = 0.3$. The regularization parameter $\lambda$ is $0.1$. We fit the log-variance using Adam with a learning rate of $10^{-4}$. 

Figure\nobreakspace \ref {fig:toy_example_fig1} demonstrates how the Pres\gls{GAN} alleviates mode collapse. The distribution learned by the regular \gls{GAN} misses $4$ modes of the target distribution. The Pres\gls{GAN} is able to recover all the modes of the target distribution.

\subsection{Assessing mode collapse}
\label{subsec:assessing_mode_collapse}

In this section we evaluate Pres\glspl{GAN}' ability to mitigate mode collapse on real datasets. We run two sets of experiments. In the first set of experiments we adopt the current experimental protocol for assessing mode collapse in the \gls{GAN} literature. That is, we use the \textsc{mnist} and \textsc{stackedmnist} datasets, for which we know the true number of modes, and report two metrics: the number of modes recovered by the Pres\gls{GAN} and the \gls{KL} divergence between the label distribution induced by the Pres\gls{GAN} and the true label distribution. In the second set of experiments we demonstrate that mode collapse can happen in \glspl{GAN} even when the number of modes is as low as $10$ but the data is imbalanced. 

\parhead{Increased number of modes.} We consider the \textsc{mnist} and \textsc{stackedmnist} datasets. \textsc{mnist} is a dataset of hand-written digits,\footnote{See \url{http://yann.lecun.com/exdb/mnist}.} in which each $28\times 28 \times 1$ image corresponds to a digit. There are $60{,}000$ training digits and $10{,}000$ digits in the test set. \textsc{mnist} has $10$ modes, one for each digit. \textsc{stackedmnist} is formed by concatenating triplets of randomly chosen \textsc{mnist} digits along the color channel to form images of size $28\times 28\times 3$ \citep{Metz2017}. We keep the same size as the original \textsc{mnist}, $60{,}000$ training digits for $10{,}000$ test digits. The total number of modes in \textsc{stackedmnist} is $1{,}000$, corresponding to the number of possible triplets.

We consider \gls{DCGAN} as the base architecture and, following \citet{radford2015unsupervised}, we resize the spatial resolution of images to $64\times 64$ pixels. 

\begin{table*}[t]
	\centering
	\small
	\captionof{table}[Assessing mode collapse on \textsc{mnist}]{Assessing mode collapse on \textsc{mnist}. The true total number of modes is $10$. All methods capture all the $10$ modes. The \acrshort{KL} captures a notion of discrepancy between the labels of real versus generated images. Pres\acrshort{GAN} generates images whose distribution of labels is closer to the data distribution, as evidenced by lower \acrshort{KL} scores.}
	\begin{tabular}{ccc}
	\toprule
	 Method & Modes & KL \\
	 \hline
	 \acrshort{DCGAN} \citep{radford2015unsupervised} &  $10 \pm 0.0$ & $0.902 \pm 0.036$ \\
	 \acrshort{VEEGAN} \citep{srivastava2017veegan} & $10 \pm 0.0$ & $0.523\pm0.008$  \\
	  \acrshort{PACGAN} \citep{lin2018pacgan} & $10 \pm 0.0$ & $0.441\pm0.009$ \\
	  Pres\acrshort{GAN} (this paper) & $\textbf{10} \pm \textbf{0.0}$ & $\mathbf{0.003 \pm 0.001}$ \\
	\bottomrule
	\end{tabular}
	\label{tab:collapse_dimensionality_mnist}
\end{table*}

\begin{table*}[t]
	\centering
	\small
	\captionof{table}[Assessing mode collapse on \textsc{stackedmnist}]{Assessing mode collapse on \textsc{stackedmnist}. The true total number of modes is $1{,}000$. All methods suffer from collapse except Pres\gls{GAN}, which captures nearly all the modes of the data distribution. Furthermore, Pres\acrshort{GAN} generates images whose distribution of labels is closer to the data distribution, as evidenced by lower \acrshort{KL} scores.}
	\begin{tabular}{ccc}
	\toprule
	 Method & Modes & KL \\
	 \hline
	 \acrshort{DCGAN} \citep{radford2015unsupervised} &  $392.0 \pm 7.376$ & $8.012 \pm 0.056$ \\
	 \acrshort{VEEGAN} \citep{srivastava2017veegan} & $761.8\pm5.741$ & $2.173\pm0.045$ \\
	  \acrshort{PACGAN} \citep{lin2018pacgan} &  $992.0\pm1.673$ & $0.277\pm0.005$ \\
	  Pres\acrshort{GAN} (this paper) & $\mathbf{999.6\pm0.489}$ & $\mathbf{0.115}\pm\mathbf{0.007}$ \\
	\bottomrule
	\end{tabular}
	\label{tab:collapse_dimensionality_smnist}
\end{table*}

To measure the degree of mode collapse we form two diversity metrics, following \citet{srivastava2017veegan}. Both of these metrics require to fit a classifier to the training data. Once the classifier has been fit, we sample $S$ images from the generator. The first diversity metric is the \textit{number of modes captured}, measured by the number of classes that are captured by the classifier. We say that a class $k$ has been captured if there is at least one generated sample for which the probability of being assigned to class $k$ is the largest. The second diversity metric is the \textit{\gls{KL} divergence} between two discrete distributions: the empirical average of the (soft) output of the classifier on generated images, and the empirical average of the (soft) output of the classifier on real images from the test set. We choose the number of generated images $S$ to match the number of test samples on each dataset. That is, $S=10{,}000$ for both \textsc{mnist} and \textsc{stackedmnist}. We expect the \gls{KL} divergence to be zero if the distribution of the generated samples is indistinguishable from that of the test samples. 

We measure the two mode collapse metrics described above against \gls{DCGAN} \citep{radford2015unsupervised} (the base architecture of Pres\gls{GAN} for this experiment). 
We also compare against other methods that aim at alleviating mode collapse in \glspl{GAN}, namely, \acrshort{VEEGAN} \citep{srivastava2017veegan} and \acrshort{PACGAN} \citep{lin2018pacgan}. 
For Pres\gls{GAN} we set the entropy regularization parameter $\lambda$ to $0.01$. 
We chose the variance thresholds to be $\bsigma_{\text{low}} = 0.001$ and $\bsigma_{\text{high}} = 0.3$.

\begin{table*}[t]
	\centering
	\small
	\captionof{table}[Assessing the impact of entropy regularization on mode collapse on \textsc{mnist} and \textsc{stackedmnist}]{Assessing the impact of the entropy regularization parameter $\lambda$ on mode collapse on \textsc{mnist} and \textsc{stackedmnist}. When $\lambda = 0$ (i.e., no entropy regularization is applied to the generator), then mode collapse occurs as expected. When entropy regularization is applied but the value of $\lambda$ is very small ($\lambda = 10^{-6}$) then mode collapse can still occur as the level of regularization is not enough. When the value of $\lambda$ is appropriate for the data then mode collapse does not occur. Finally, when $\lambda$ is too high then mode collapse can occur because the entropy maximization term dominates and the data is poorly fit.}
	\begin{tabular}{ccccc}
	\toprule
	   &  \multicolumn{2}{c}{\textsc{mnist}} &  \multicolumn{2}{c}{\textsc{stackedmnist}} \\
	 \midrule
	 $\lambda$ & Modes & KL & Modes & KL \\
	 \hline
	$0$ & $10 \pm 0.0$ & $0.050 \pm 0.0035$ & $418.2 \pm 7.68$ & $4.151 \pm 0.0296$ \\
	$10^{-6}$ & $10\pm0.0$ & $0.005\pm0.0008$ & $989.8\pm1.72$& $0.239\pm0.0059$ \\
	$10^{-2}$ & $\textbf{10} \pm \textbf{0.0}$ & $\textbf{0.003}\pm\textbf{0.0006}$ & $\textbf{999.6}\pm\textbf{0.49}$& $0.115\pm0.0074$ \\
	$5\times 10^{-2}$ & $10 \pm 0.0$ & $0.004\pm0.0008$ & $999.4\pm0.49$ & $\textbf{0.099}\pm\textbf{0.0047}$ \\
	$10^{-1}$ & $10 \pm 0.0$ & $0.005\pm0.0004$ & $999.4\pm0.80$ & $0.102\pm0.0032$\\
	$5\times 10^{-1}$ & $10\pm0.0$ & $0.006\pm0.0011$ & $907.0\pm9.27$& $0.831\pm0.0209$ \\
	\bottomrule
	\end{tabular}
	\label{tab:collapse_dim_bis}
\end{table*}

\begin{figure*}[t]
	\centering
	\vspace*{-10pt}
	\centerline{\includegraphics[width=1.2\textwidth]{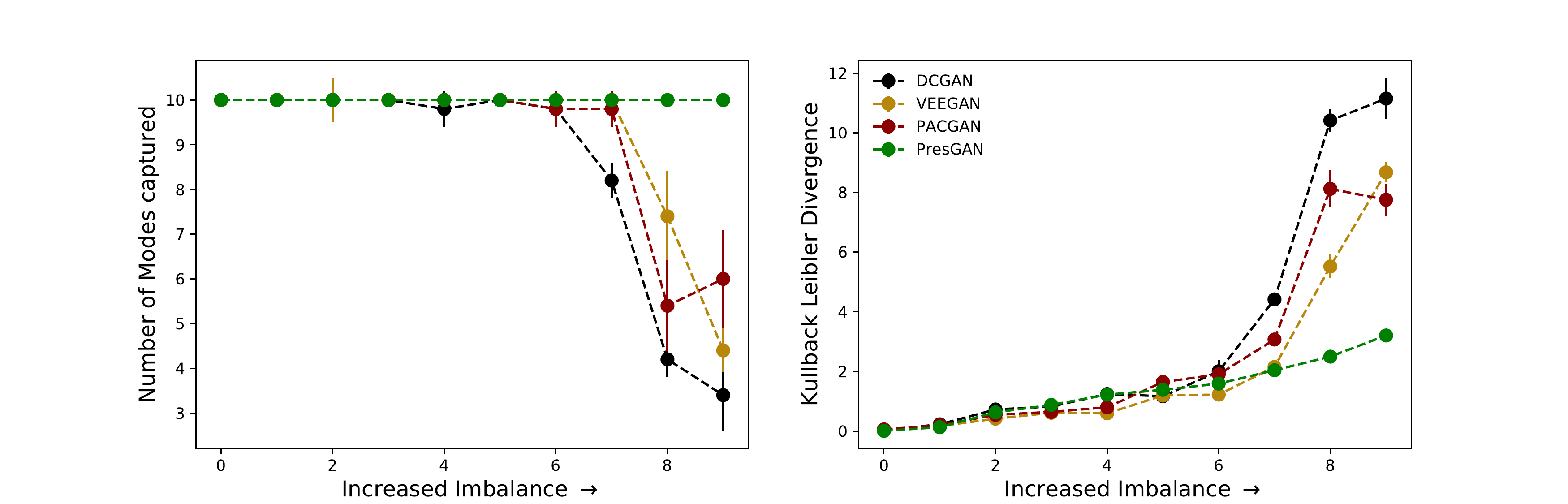}}
	\caption[Assessing mode collapse under increased data imbalance]{Assessing mode collapse under increased data imbalance on \textsc{mnist}. The figures show the number of modes captured (higher is better) and the \gls{KL} divergence (lower is better) under increasingly imbalanced settings. The maximum number of modes in each case is $10$. All methods suffer from mode collapse as the level of imbalance increases except for the Pres\gls{GAN} which is robust to data imbalance.}
	\label{fig:imbalance}
	\vspace*{-8pt}
\end{figure*}

Tables\nobreakspace \ref {tab:collapse_dimensionality_mnist} and\nobreakspace  \ref {tab:collapse_dimensionality_smnist} show the number of captured modes and the \gls{KL} for each method. The results are averaged across $5$ runs. All methods capture all the modes of \textsc{mnist}. This is not the case on \textsc{stackedmnist}, where the Pres\gls{GAN} is the only method that can capture all the modes. 
Finally, the proportion of observations in each mode of Pres\gls{GAN} is closer to the true proportion in the data, as evidenced by lower \acrshort{KL} divergence scores.

We also study the impact of the entropy regularization by varying the hyperparameter $\lambda$ from $0$ to $0.5$. Table\nobreakspace \ref {tab:collapse_dim_bis} illustrates the results. Unsurprisingly, when there is no entropy regularization, i.e., when $\lambda = 0$, then mode collapse occurs. This is also the case when the level of regularization is not enough ($\lambda = 10^{-6}$). There is a whole range of values for $\lambda$ such that mode collapse does not occur ($\lambda \in \{0.01, 0.05, 0.1\}$). Finally, when $\lambda$ is too high for the data and architecture under study, mode collapse can still occur. This is because when $\lambda$ is too high, the entropy regularization term dominates the loss in Eq.\nobreakspace \ref {eq:sigan_loss} and in turn the generator does not fit the data as well. This is also evidenced by the higher \gls{KL} divergence score when $\lambda = 0.5$ vs.\ when $0 < \lambda < 0.5$.

\parhead{Increased data imbalance.} We now show that mode collapse can occur in \glspl{GAN} when the data is imbalanced, even when the number of modes of the data distribution is small. We follow \citet{dieng2018learning} and consider a perfectly balanced version of \textsc{mnist} as well as nine imbalanced versions. To construct the balanced dataset we used $5{,}000$ training examples per class, totaling $50{,}000$ training examples. We refer to this original balanced dataset as ${D_0}$. Each additional training set ${D_k}$ leaves only $5$ training examples for each class $j \leq k$, and $5{,}000$ for the rest. (See the Appendix for all the class distributions.)

We used the same classifier trained on the unmodified \textsc{mnist} but fit each method on each of the $9$ new \textsc{mnist} distributions. We chose $\lambda = 0.1$ for Pres\gls{GAN}. Figure\nobreakspace \ref {fig:imbalance} illustrates the results in terms of both metrics---number of modes and \gls{KL} divergence. \gls{DCGAN}, \acrshort{VEEGAN}, and \acrshort{PACGAN} face mode collapse as the level of imbalance increases. This is not the case for Pres\gls{GAN}, which is robust to imbalance and captures all the $10$ modes. 

\subsection{Assessing sample quality}
\label{subsec:assessing_sample_quality}

\begin{figure}[!hbpt]
	\includegraphics[width=0.45\textwidth]{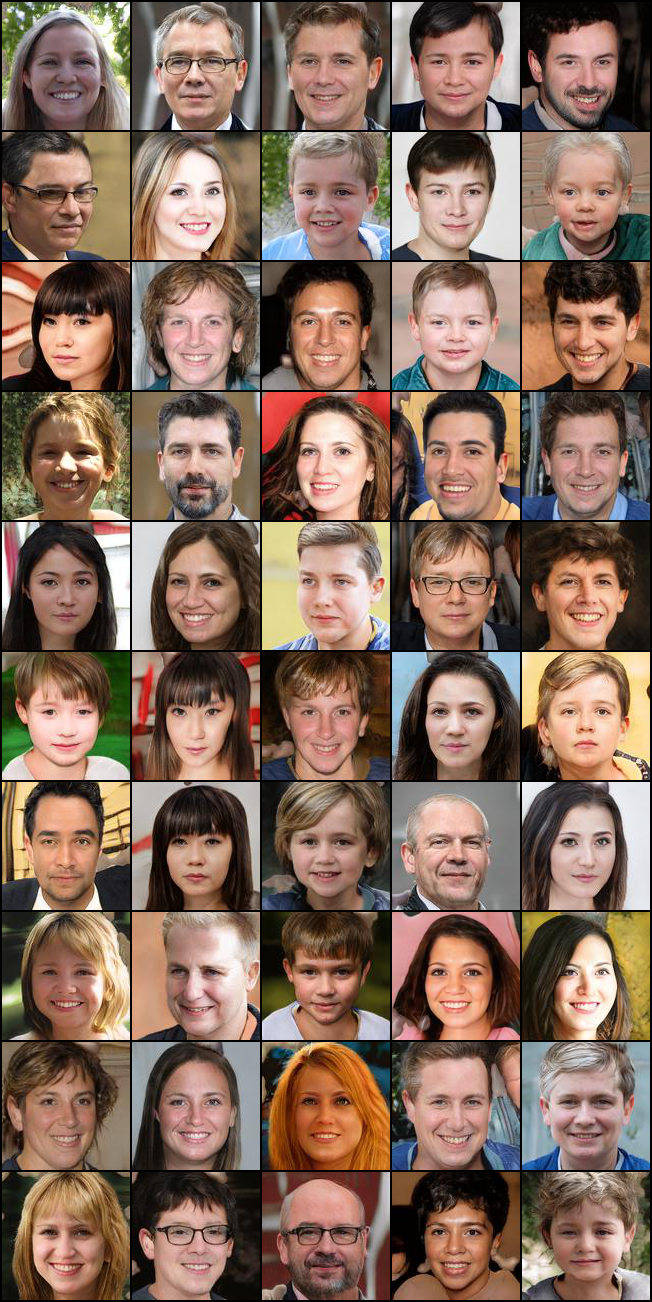}\quad\quad\quad
	\includegraphics[width=0.45\textwidth]{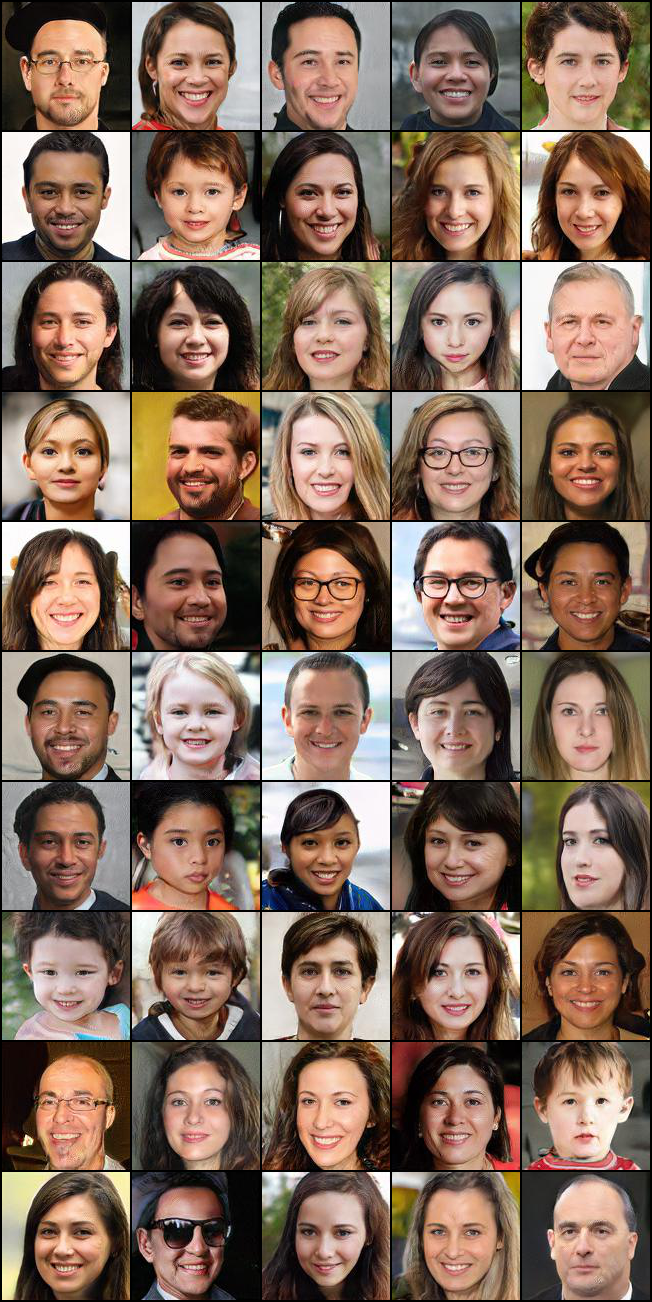}
	\caption[Pres\gls{GAN} enables diverse image generation. Illustration on the \textsc{ffhq} dataset]{Generated images on \textsc{ffhq} for Style\gls{GAN} (left) and Pres\gls{GAN} (right). The Pres\gls{GAN} maintains the high perceptual quality of the Style\gls{GAN}.}
	\label{fig:ffhq}
\end{figure}

In this section we assess Pres\glspl{GAN}' ability to generate samples of high perceptual quality. We rely on perceptual quality of generated samples and on \gls{FID} scores \citep{heusel2017gans}. We also consider two different \gls{GAN} architectures, the standard \gls{DCGAN} and the more recent Style\gls{GAN}, to show robustness of Pres\glspl{GAN} vis-a-vis the underlying \gls{GAN} architecture.

\parhead{\gls{DCGAN}.} We use \gls{DCGAN} \citep{radford2015unsupervised} as the base architecture and build Pres\gls{GAN} on top of it. We consider four datasets: \textsc{mnist}, \textsc{stackedmnist}, \textsc{cifar}-10, and CelebA. \textsc{cifar}-10 \citep{krizhevsky2009learning} is a well-studied dataset of $32 \times 32$ images that are classified into one of the following categories: airplane, automobile, bird, cat, deer, dog, frog, horse, ship, and truck. CelebA \citep{liu2015deep} is a large-scale face attributes dataset. Following \citet{radford2015unsupervised}, we resize all images to $64 \times 64$ pixels.
We use the default \gls{DCGAN} settings. We refer the reader to the code we used for \gls{DCGAN}, which was taken from \url{https://github.com/pytorch/examples/tree/master/dcgan}. We set the seed to $2019$ for reproducibility.

\begin{table*}[t]
	\centering
	\small
	\captionof{table}[Assessing sample quality of different generative models]{\acrfull{FID} (lower is better). Pres\acrshort{GAN} has lower \acrshort{FID} scores than \acrshort{DCGAN}, \acrshort{VEEGAN}, and \acrshort{PACGAN}. This is because Pres\acrshort{GAN} mitigates mode collapse while preserving sample quality.}
	\begin{tabular}{ccc}
	\toprule
	 Method & Dataset &  \acrshort{FID} \\
	 \hline
	 \acrshort{DCGAN} \citep{radford2015unsupervised} & \textsc{mnist} & $113.129\pm0.490$ \\
	 \acrshort{VEEGAN} \citep{srivastava2017veegan} & \textsc{mnist} & $68.749\pm0.428$ \\
	  \acrshort{PACGAN}   \citep{lin2018pacgan}  & \textsc{mnist} & $58.535\pm0.135$ \\
	  Pres\acrshort{GAN} (this paper) & \textsc{mnist} & $\textbf{42.019}\pm\textbf{0.244}$ \\
	  \hline
	  \acrshort{DCGAN}  & \textsc{stackedmnist} & $97.788\pm0.199$ \\
	 \acrshort{VEEGAN} & \textsc{stackedmnist} &  $86.689\pm0.194$ \\
	  \acrshort{PACGAN} & \textsc{stackedmnist} & $117.128\pm0.172$ \\
	  Pres\acrshort{GAN} & \textsc{stackedmnist} & $\textbf{23.965}\pm\textbf{0.134}$ \\
	  \hline
	  \acrshort{DCGAN}  & \textsc{cifar}-10 & $103.049\pm0.195$  \\
	 \acrshort{VEEGAN} & \textsc{cifar}-10 & $95.181\pm0.416$  \\
	  \acrshort{PACGAN} & \textsc{cifar}-10 &  $54.498\pm0.337$ \\
	  Pres\acrshort{GAN} & \textsc{cifar}-10 & $\textbf{52.202}\pm\textbf{0.124}$ \\
	  \hline
	  \acrshort{DCGAN}  & \textsc{celeba} & $39.001\pm0.243$ \\
	 \acrshort{VEEGAN} & \textsc{celeba} & $46.188\pm0.229$  \\
	  \acrshort{PACGAN} & \textsc{celeba} & $36.058\pm0.212$ \\
	  Pres\acrshort{GAN} & \textsc{celeba} & $\textbf{29.115}\pm\textbf{0.218}$ \\
	\bottomrule
	\end{tabular}
	\label{tab:fid}
\end{table*}

There are hyperparameters specific to Pres\gls{GAN}. These are the noise and \gls{HMC} hyperparameters. We set the learning rate for the noise parameters $\bsigma$ to $10^{-3}$ and constrain its values to be between $10^{-3}$ and $0.3$ for all datasets. We initialize $\log \bsigma$ to $-0.5$. We set the burn-in and the number of \gls{HMC} samples to $2$. We choose a standard number of $5$ leapfrog steps and set the \gls{HMC} learning rate to $0.02$. The acceptance rate is fixed at $0.67$. We found that different $\lambda$ values worked better for different datasets. We used $\lambda = 5\times 10^{-4}$ for \textsc{cifar}-10 and \textsc{celeba} $\lambda = 0.01$ for \textsc{mnist} and \textsc{stackedmnist}.

We found the Pres\gls{GAN}'s performance to be robust to the default settings for most of these hyperparameters. However we found the initialization for $\bsigma$ and its learning rate to play a role in the quality of the generated samples. The hyperparameters mentioned above for $\bsigma$ worked well for all datasets. 

Table\nobreakspace \ref {tab:fid} shows the \gls{FID} scores for \gls{DCGAN} and Pres\gls{GAN} across the four datasets.  
We can conclude that Pres\gls{GAN} generates images of high visual quality. In addition, the \gls{FID} scores are lower because Pres\gls{GAN} explores more modes than \gls{DCGAN}. Indeed, when the generated images account for more modes, the \gls{FID} sufficient statistics (the mean and covariance of the Inception-v3 pool3 layer) of the generated data get closer to the sufficient statistics of the empirical data distribution.

We also report the \gls{FID} for \acrshort{VEEGAN} and  \acrshort{PACGAN} in Table\nobreakspace \ref {tab:fid}.
\acrshort{VEEGAN} achieves better \gls{FID} scores than \acrshort{DCGAN} on all datasets but \textsc{celeba}. This is because \acrshort{VEEGAN} collapses less than \acrshort{DCGAN} as evidenced by Table\nobreakspace \ref {tab:collapse_dimensionality_mnist} and Table\nobreakspace \ref {tab:collapse_dimensionality_smnist}.  \acrshort{PACGAN} achieves better \gls{FID} scores than both \acrshort{DCGAN} and \acrshort{VEEGAN} on all datasets but on \textsc{stackedmnist} where it achieves a significantly worse \gls{FID} score. Finally, Pres\gls{GAN} outperforms all of these methods on the \gls{FID} metric on all datasets signaling its ability to mitigate mode collapse while preserving sample quality.

Besides the \gls{FID} scores, we also assess the visual quality of the generated images. In  of the appendix, we show randomly generated (not cherry-picked) images from \gls{DCGAN},  \acrshort{VEEGAN}, \acrshort{PACGAN}, and Pres\gls{GAN}.  
For Pres\gls{GAN}, we show the mean of the conditional distribution of $\bx$ given $\bz$. The samples generated by Pres\gls{GAN} have high visual quality; in fact their quality is comparable to or better than the \gls{DCGAN} samples.

\parhead{Style\gls{GAN}.} We now consider a more recent \gls{GAN} architecture (Style\gls{GAN})~\citep{karras2019style} and a higher resolution image dataset (\textsc{ffhq}). 
\textsc{ffhq} is a diverse dataset of faces from Flickr\footnote{See \url{https://github.com/NVlabs/ffhq-dataset}.} introduced by \citet{karras2019style}. The dataset contains $70{,}000$ high-quality \textsc{png} images with considerable variation in terms of age, ethnicity, and image background. We use a resolution of $128 \times 128$ pixels. 

Style\gls{GAN} feeds multiple sources of noise $\bz$ to the generator. In particular, it adds Gaussian noise after each convolutional layer before evaluating the nonlinearity. Building Pres\gls{GAN} on top of Style\gls{GAN} therefore requires to sample all noise variables $\bz$ through \gls{HMC} at each training step. To speed up the training procedure, we only sample the noise variables corresponding to the input latent code and condition on all the other Gaussian noise variables. In addition, we do not follow the progressive growing of the networks of \citet{karras2019style} for simplicity.

For this experiment, we choose the same \gls{HMC} hyperparameters as for the previous experiments but restrict the variance of the generative distribution to be $\bsigma_{\text{high}} = 0.2$. We set $\lambda=0.001$ for this experiment. 

Figure\nobreakspace \ref {fig:ffhq} shows cherry-picked images generated from Style\gls{GAN} and Pres\gls{GAN}. We can observe that the Pres\gls{GAN} maintains as good perceptual quality as the base architecture. In addition, we also observed that the Style\gls{GAN} tends to produce some redundant images (these are not shown in Figure\nobreakspace \ref {fig:ffhq}), something that we did not observe with the Pres\gls{GAN}. This lack of diversity was also reflected in the \gls{FID} scores which were $14.72 \pm 0.09$ for Style\gls{GAN} and $12.15 \pm 0.09$ for Pres\gls{GAN}. These results suggest that entropy regularization effectively reduces mode collapse while preserving sample quality.

\subsection{Assessing held-out predictive log-likelihood}
\label{subsec:assessing_log_lik}

In this section we evaluate Pres\glspl{GAN} for generalization using predictive log-likelihood. We use the \gls{DCGAN} architecture to build Pres\gls{GAN} and evaluate the log-likelihood on two benchmark datasets, \textsc{mnist} and \textsc{cifar}-10. We use images of size $32 \times 32$.

We compare the generalization performance of the Pres\gls{GAN} against the \gls{VAE} \citep{kingma2013auto, rezende2014stochastic} by controlling for the architecture and the evaluation procedure. In particular, we fit a \gls{VAE} that has the same decoder architecture as the Pres\gls{GAN}. We form the \gls{VAE} encoder by using the same architecture as the \gls{DCGAN} discriminator and getting rid of the output layer. We used linear maps to get the mean and the log-variance of the approximate posterior. 

To measure how Pres\glspl{GAN} compare to traditional \glspl{GAN} in terms of log-likelihood, we also fit a Pres\gls{GAN} with $\lambda = 0$. 

Consider an unseen datapoint $\bx^*$. We estimate its log marginal likelihood $\log p_{\theta}(\bx^*)$ using importance sampling,
\begin{equation}\label{eq:loglik}
  \log p_{\theta}(\bx^*) \approx \log \left( \frac{1}{S} \sum_{s=1}^{S} \frac{p_{\theta}\left(\bx^*\g \bz^{(s)}\right)\cdot p\left(\bz^{(s)}\right)}{r\left(\bz^{(s)}\g \bx^*\right)}\right),
\end{equation}
where we draw $S$ samples $\bz^{(1)}, \dots, \bz^{(S)}$ from a proposal distribution $r(\bz\g \bx^*)$.

There are different ways to form a good proposal $r(\bz\g \bx^*)$, and we discuss several alternatives 
in Section\nobreakspace \ref {app:proposals} of the appendix. In this paper, we take the following approach. We define the proposal as a Gaussian distribution,
\begin{align}
  r(\bz\g \bx^*) &= \mathcal{N}(\bmu_r, \bSigma_r) 
  .
\end{align}
We set the mean parameter $\bmu_r$ to the \textit{maximum a posteriori} solution, i.e., 
\begin{align*}
	\bmu_r &= \argmax_z \left( \log p_{\theta}\left(\bx^*\g \bz \right) + \log p\left(\bz\right) \right)
	.
\end{align*}
We initialize this maximization algorithm using the mean of a pre-fitted encoder, $q_{\gamma}(\bz\g \bx^*)$. 
The encoder is fitted by minimizing the reverse \gls{KL} divergence between $q_{\gamma}(\bz\g \bx)$ and the 
true posterior $p_{\theta}(\bz\g \bx)$ using the training data. This \gls{KL} is 
\begin{align}\label{eq:kl}
  &\gls{KL}\left( q_{\gamma}(\bz\g \bx) \vert\vert p_{\theta}(\bz\g \bx) \right) 
  = \log p_{\theta}(\bx) - \mathbb{E}_{q_{\gamma}(\bz\g \bx)}\left[ \log p_{\theta}(\bx \g \bz)p(\bz) - \log q_{\gamma}(\bz\g \bx) \right]
  .
\end{align}
Because the generative distribution is fixed at test time, minimizing the \gls{KL} here is equivalent to 
maximizing the second term in Eq.\nobreakspace \ref {eq:kl}, which is the \gls{ELBO} objective of \glspl{VAE}.

We set the proposal covariance $\bSigma_r$ as an overdispersed version\footnote{In general, overdispersed proposals lead to better importance sampling estimates.} of the encoder's covariance matrix, which is diagonal. In particular, to obtain $\bSigma_r$ we multiply the elements of the encoder's covariance by a factor $\gamma$. In our experiments we set $\gamma$ to $1.2$.

We use $S=2{,}000$ samples to form the importance sampling estimator. Since the pixel values are normalized in $[-1, +1]$, we use a truncated Gaussian likelihood for evaluation. Specifically, for each pixel of the test image, we divide the Gaussian likelihood by the probability (under the generative model) that the pixel is within the interval $[-1, +1]$. We use the truncated Gaussian likelihood at test time only.

\parhead{Settings.}
For the Pres\gls{GAN}, we use the same \gls{HMC} hyperparameters as for the previous experiments. 
We constrain the variance of the generative distribution using $\bsigma_{\text{low}} = 0.001$ and $\bsigma_{\text{high}} = 0.2$. 
We use the default \gls{DCGAN} values for the remaining hyperparameters, including the optimization settings. 
For the \textsc{cifar}-10 experiment, we choose $\lambda = 0.001$. We set all learning rates to $0.0002$. 
We set the dimension of the latent variables to $100$. We ran both the \gls{VAE} and the Pres\gls{GAN} 
for a maximum of $200$ epochs. For \textsc{mnist}, we use the same settings as for \textsc{cifar}-10 
but use $\lambda = 0.0001$ and ran all methods for a maximum of $50$ epochs.

\begin{table}[t]
	\centering
	\small
	\captionof{table}[Pres\gls{GAN} reduces the gap in generalization performance, as measured by held-out log-likelihood, between a \gls{GAN} and a \gls{VAE}]{Generalization performance as measured by negative log-likelihood (lower is better) on \textsc{mnist} and \textsc{cifar}-10. Here the \gls{GAN} denotes a Pres\gls{GAN} fitted without entropy regularization ($\lambda = 0$). The Pres\gls{GAN} reduces the gap in performance between the \gls{GAN} and the \gls{VAE} on both datasets.}
	\begin{tabular}{ccccc}
		\toprule
		& \multicolumn{2}{c}{\textsc{mnist}} & \multicolumn{2}{c}{\textsc{cifar}-10} \\
		& Train & Test & Train & Test \\ \midrule
		\gls{VAE}     & $-3483.94$ & $-3408.16$ & $-1978.91$ & $-1665.84$   \\
		\gls{GAN} & $-1410.78$ & $-1423.39$ & $-572.25$ & $-569.17$  \\
		Pres\gls{GAN} & $-1418.91$ & $-1432.50$ & $-1050.16$ & $-1031.70$  \\
		\bottomrule
	\end{tabular}
	\label{tab:loglik}
\end{table}

\parhead{Results.}
Table\nobreakspace \ref {tab:loglik} summarizes the results. Here \gls{GAN} denotes the Pres\gls{GAN} fitted using $\lambda = 0$. The \gls{VAE} outperforms both the \gls{GAN} and the Pres\gls{GAN} on both \textsc{mnist} and \textsc{cifar}-10. This is unsurprising given \glspl{VAE} are fitted to maximize log-likelihood. The \gls{GAN}'s performance on \textsc{cifar}-10 is particularly bad, suggesting it suffered from mode collapse. The Pres\gls{GAN}, which mitigates mode collapse achieves significantly better performance than the \gls{GAN} on \textsc{cifar}-10. To further analyze the generalization performance, we also report the log-likelihood on the training set in Table\nobreakspace \ref {tab:loglik}. We can observe that the difference between the training log-likelihood and the test log-likelihood is very small for all methods.

\section{Appendix}
\stepcounter{section}

\subsection{Other ways to compute predictive log-likelihood}\label{app:proposals}

Here we discuss different ways to obtain a proposal in order to approximate the predictive log-likelihood.
For a test instance $\bx^*$, we estimate the marginal log-likelihood $\log p_{\theta}(\bx^*)$ using importance sampling,
\begin{equation}\label{supp_eq:loglik}
	\log p_{\theta}(\bx^*) \approx \log \left( \frac{1}{S} \sum_{s=1}^{S} \frac{p_{\theta}\left(\bx^*\g \bz^{(s)}\right)\; p\left(\bz^{(s)}\right)}{r\left(\bz^{(s)}\g \bx^*\right)}\right),
\end{equation}
where we draw the $S$ samples $\bz^{(1)}, \dots, \bz^{(S)}$ from a proposal distribution $r(\bz\g \bx^*)$. We next discuss different ways to form the proposal $r(\bz\g \bx^*)$. 

One way to obtain the proposal is to set $r(\bz\g \bx^*)$ as a Gaussian distribution whose mean and variance are computed using samples from an \acrshort{HMC} algorithm with stationary distribution $p_{\theta}(\bz\g \bx^*)\propto p_{\theta}(\bx^*\g \bz)p(\bz)$. That is, the mean and variance of $r(\bz\g \bx^*)$ are set to the empirical mean and variance of the \acrshort{HMC} samples.

The procedure above requires to run an \acrshort{HMC} sampler, and thus it may be slow. We can accelerate the procedure with a better initialization of the \acrshort{HMC} chain.
Indeed, the second way to evaluate the log-likelihood also requires the \acrshort{HMC} sampler, but
it is initialized using a mapping $\bz = g_{\eta}(\bx^\star)$. 
The mapping $g_{\eta}(\bx^\star)$ is a network that maps from observed space $\bx$ to latent space $\bz$. The parameters $\eta$ of the network can be learned at test time using generated data. In particular, $\eta$ can be obtained by generating data from the fitted generator of Pres\acrshort{GAN} and then fitting $g_{\eta}(\bx^\star)$ to map $\bx$ to $\bz$ by maximum likelihood. This is, we first sample $M$ pairs $(\bz_m, \bx_m)_{m=1}^{M}$ from the learned generative distribution and then we obtain $\eta$ by minimizing
$\sum_{m=1}^{M} || \bz_m - g_{\eta}(\bx_m) ||_2^2$.
Once the mapping is fitted, we use it to initialize the \acrshort{HMC} chain.

A third way to obtain the proposal is to learn an encoder network $q_{\eta}(\bz \g \bx)$ jointly with the rest of the Pres\acrshort{GAN} parameters. This is effectively done by letting the discriminator distinguish between pairs $(\bx, \bz) \sim p_d(\bx)\cdot q_{\eta}(\bz \g \bx)$ and $(\bx, \bz) \sim p_{\theta}(\bx, \bz)$ rather than discriminate $\bx$ against samples from the generative distribution. These types of discriminator networks have been used to learn a richer latent space for \acrshort{GAN}~\citep{donahue2016adversarial, dumoulin2016adversarially}. 
In such cases, we can use the encoder network $q_{\eta}(\bz\g \bx)$ to define the proposal, either by setting $r(\bz\g \bx^*)=q_{\eta}(\bz\g \bx^*)$ or by initializing the \acrshort{HMC} sampler at the encoder mean.

The use of an encoder network is appealing but it requires a discriminator that takes pairs $(\bx,\bz)$. The approach that we follow in the paper also uses an encoder network but keeps the discriminator the same as for the base \acrshort{DCGAN}. We found this approach to work better in practice. More in detail, we use an encoder network $q_{\eta}(\bz \g \bx)$; however the encoder is fitted at test time by maximizing the variational \acrshort{ELBO}, given by $\sum_n \E{q_{\eta}(\bz_n\g \bx_n)}{\log p_{\theta}(\bx_n,\bz_n) - \log q_{\eta}(\bz_n\g \bx_n)}$. We set the proposal $r(\bz\g \bx^*)=q_{\eta}(\bz\g \bx^*)$. (Alternatively, the encoder can be used to initialize a sampler.)

\subsection{Assessing mode collapse under increased data imbalance}
In the main paper we show that mode collapse can happen not only when there are increasing number of modes, as done in the \gls{GAN} literature, but also when the data is imbalanced. We consider a perfectly balanced version of \textsc{mnist} by using 5,000 training examples per class, totalling 50,000 training examples. We refer to this original balanced dataset as {\bf D$1$}. We build nine additional training sets from this balanced dataset. Each additional training set {\bf D$k$} leaves only $5$ training examples for each class $j < k$. See Table~\ref{supp_tab:class_dist} for all the class distributions. 

\begin{table*}[!hbpt]
\centering
\caption[Class distributions used to create $9$ different imbalanced datasets from \textsc{mnist}]{Class distributions using the \textsc{mnist} dataset. There are $10$ class---one class for each of 
the $10$ digits in \textsc{mnist}. The distribution D$1$ is uniform and the other distributions correspond 
to different imbalance settings as given by the proportions in the table. Note these proportions might not sum to one exactly because of rounding.}
\begin{tabular}[\textwidth]{ccccccccccc}
\toprule
Dist & $0$ & $1$ & $2$  & $3$ & $4$ &  $5$  &  $6$  &  $7$  & $8$  &  $9$ \\
\midrule
D$1$ &  $0.1$ & $0.1$ & $0.1$ & $0.1$ & $0.1$ & $0.1$ & $0.1$ & $0.1$ & $0.1$ & $0.1$ \\
D$2$ &  $10^{-3}$ & $0.11$ & $0.11$ & $0.11$ & $0.11$ & $0.11$ & $0.11$ & $0.11$ & $0.11$ & $0.11$   \\
D$3$ & $10^{-3}$ & $10^{-3}$ & $0.12$  & $0.12$  & $0.12$  & $0.12$  & $0.12$  & $0.12$  & $0.12$ & $0.12$     \\
D$4$ &  $10^{-3}$ & $10^{-3}$ & $10^{-3}$ & $0.14$  & $0.14$   & $0.14$  & $0.14$   & $0.14$   & $0.14$  & $0.14$  \\
D$5$ &  $10^{-3}$ & $10^{-3}$ & $10^{-3}$ & $10^{-3}$  & $0.17$ & $0.17$ & $0.17$ & $0.17$ & $0.17$ & $0.17$   \\
D$6$ &  $10^{-3}$ &$10^{-3}$ &  $10^{-3}$ & $10^{-3}$ & $10^{-3}$  & $0.20$ & $0.20$ & $0.20$ & $0.20$ & $0.20$ \\
D$7$ &  $10^{-3}$& $10^{-3}$ & $10^{-3}$ & $10^{-3}$  & $10^{-3}$ &  $10^{-3}$ & $0.25$ & $0.25$ & $0.25$ & $0.25$   \\
D$8$ &  $10^{-3}$ & $10^{-3}$ &  $10^{-3}$ & $10^{-3}$  & $10^{-3}$ &  $10^{-3}$ & $10^{-3}$ & $0.33$  & $0.33$ & $0.33$  \\
D$9$ &  $10^{-3}$ & $10^{-3}$ & $10^{-3}$ & $10^{-3}$  & $10^{-3}$ &  $10^{-3}$ & $10^{-3}$ &  $10^{-3}$ & $0.49$   & $0.49$  \\
D$10$ & $10^{-3}$& $10^{-3}$ &  $10^{-3}$ & $10^{-3}$  & $10^{-3}$ & $10^{-3}$ &$10^{-3}$ &  $10^{-3}$ & $10^{-3}$   & $0.99$   \\
\bottomrule
\end{tabular}
\label{supp_tab:class_dist}
\end{table*}

\clearpage

\begin{center}
\pagebreak
\vspace*{5\baselineskip}
\textbf{\large Conclusion}\label{chap:conclusion}
\end{center}

\hspace{10mm}

Probabilistic graphical modeling with latent variables provides a useful framework for learning from data. 
It enables accounting for uncertainty, learning the latent structure underlying data in an interpretable way, 
and incorporating prior knowledge. However probabilistic graphical modeling might lack flexibility 
for the purpose of learning from the types of high-dimensional complex data we currently encounter in practice. 
This thesis developed deep probabilistic graphical modeling, which leverages deep learning 
to bring flexibility to probabilistic graphical modeling. 
We used neural networks to extend the canonical \gls{EF-PCA} to model and learn interpretable quantities from image and text data. 
We leveraged recurrent neural networks to build a class of models for sequential data where long-term 
dependencies are accounted for using latent variables. We solved several problems that  
probabilistic topic models suffer from using distributed representations of words for model specification 
and neural networks for inference. This thesis also made contributions on the algorithmic front. 
We developed reweighted expectation maximization (\gls{REM}), an algorithm that unifies several existing 
maximum likelihood-based algorithms for learning models parameterized by 
deep neural networks. This unifying view is made possible using expectation maximization, 
a canonical inference algorithm for probabilistic graphical models. 
\gls{REM} leads to better generalization to unseen data.  
Finally, we showed how to leverage the learning procedure behind generative adversarial networks 
to fit probabilistic latent-variable models. This new algorithm, called entropy-regularized adversarial learning,  
constitutes a solution to the mode collapse problem that is pervasive in generative adversarial networks.  

There are several choice points for deep probabilistic graphical modeling, 
each of which can be explored for future work. 
\begin{enumerate}
	\item \textbf{Prior.} Choosing a prior pertains to specifying our a priori knowledge of the latent structure. 
	Several of the model classes we developed above used simple priors. Future work will explore how to devise 
	richer priors for deep probabilistic graphical modeling. We will also explore how to translate domain knowledge 
	into prior specification to apply deep probabilistic graphical modeling to new domains (e.g. science.)
	\item \textbf{Likelihood.} We leveraged neural networks or word embeddings to define the conditional distribution 
	of the data given the latent variables as an exponential family. The exponential family provides an umbrella distribution 
	for the types of data we encounter in practice (e.g. real-valued, categorical, and binary.) Future work can explore other 
	distributional forms for the likelihood (e.g. distributions specified via a sampling procedure) or use constrained neural networks 
	to parameterize the likelihood (e.g. invertible neural networks.)
	\item \textbf{Posterior.} We used variational inference as a framework for inferring the posterior distribution of the 
	latent variables. In particular, we used distributions that are amenable to reparameterization such as the Gaussian, 
	and parameterized them using neural networks. Future work can explore other choices of approximate posterior distributions, 
	especially for discrete latent variables often used in probabilistic graphical modeling. 
	\item \textbf{Algorithm.} We explored both maximum likelihood and adversarial learning for model fitting. These two 
	paradigms are complementary. Adversarial learning favors high quality of simulation of new data while maximum likelihood 
	favors high held-out likelihood on unseen data. In future work we will explore how we can combine the strength of these two 
	approaches to achieve all aspects of generalization as described in our desiderata. We will also explore how to make 
	entropy-regularized adversarial learning amenable to discrete data and discrete latent variables. 
\end{enumerate}
 \addcontentsline{toc}{chapter}{Conclusion}

\titleformat{\chapter}[display]
{\normalfont\bfseries\filcenter}{}{0pt}{\large\bfseries\filcenter{#1}}  \titlespacing*{\chapter}
  {0pt}{0pt}{30pt}

\addcontentsline{toc}{chapter}{References}  
\begin{singlespace}
\bibliographystyle{icml} 
\bibliography{thesis.bib} 
\end{singlespace}

\titleformat{\chapter}[display]
{\normalfont\bfseries\filcenter}{}{0pt}{\large\chaptertitlename\ \large\thechapter : \large\bfseries\filcenter{#1}}  
\titlespacing*{\chapter}
  {0pt}{0pt}{30pt}	  
\titleformat{\section}{\normalfont\bfseries}{\thesection}{1em}{#1}

\titleformat{\subsection}{\normalfont}{\thesubsection}{0em}{\hspace{1em}#1}

\end{document}